\documentclass[]{elsarticle}
\usepackage{lineno,hyperref}
\usepackage[utf8]{inputenc}
\usepackage{booktabs,tabularx}
\usepackage{times}
\usepackage{epsfig}
\usepackage{graphicx}
\usepackage{amsmath}
\usepackage{amssymb}
\usepackage{pifont}
\usepackage[table,usenames,dvipsnames,svgnames]{xcolor} %for use in color links
\usepackage{colortbl}
\usepackage{wrapfig}
\usepackage{tabu}
\usepackage{bm}
\usepackage{algorithm}
\usepackage[noend]{algpseudocode}
\usepackage{upgreek}
\usepackage{relsize}
\graphicspath{{figures/}}

\makeatother

\usepackage{blindtext}

\newcommand{\dashrule}[1][black]{%
	\color{#1}\rule[\dimexpr.5ex-.2pt]{4pt}{.4pt}\xleaders\hbox{\rule{4pt}{0pt}\rule[\dimexpr.5ex-.2pt]{4pt}{.4pt}}\hfill\kern0pt%
}
\newcommand{\rulecolor}[1]{%
	\def\CT@arc@{\color{#1}}%
}

\usepackage{booktabs}
\usepackage{multirow}

\usepackage{stackengine}
\usepackage{scalerel}
\newlength\lthk
\definecolor{verylightgray}{gray}{0.95}
\newlength{\tabwidth}

\newcommand{\proc}[1]{\textcolor[rgb]{0.64,0,0}{#1}}
\newcommand{\abso}[1]{\textcolor[rgb]{0,0.6875,0.93}{#1}}
\newcommand{\gt}[1]{\hat{#1}}

\newcommand{\ITSCalib}{ITS15}
\newcommand{\EdgeLetsCalib}{Edgelets}
\newcommand{\ManualCalib}{ManualCalib}

\newcommand{\BMVCScale}{BMVC14}

\newcommand{\BBScaleReg}{BBScale + reg}
\newcommand{\ManualScale}{ManualScale}
\newcommand{\SpeedScale}{SpeedScale}

\hypersetup{
	colorlinks=true,
	linkcolor=BrickRed,
	citecolor=OliveGreen,
	filecolor=magenta,
	urlcolor=cyan
}

\journal{Computer Vision and Image Understanding.}
\usepackage{numcompress}\biboptions{authoryear}

\begin{document}
\begin{frontmatter}

\title{Traffic Surveillance Camera Calibration\\by 3D Model Bounding Box Alignment\\for Accurate Vehicle Speed Measurement}

\cortext[fn:BrnoPhdTalent]{Accepted to CVIU. DOI: 10.1016/j.cviu.2017.05.015. Corresponding author at BUT FIT, Božetěchova 2, 612 66 Brno, Czech Republic. Jakub Sochor is a Brno Ph.D. Talent Scholarship Holder --- Funded by the Brno City Municipality.}
\author[AddressIT4I]{Jakub Sochor\corref{fn:BrnoPhdTalent}}
\ead{isochor@fit.vutbr.cz}
\author[AddressIT4I]{Roman Juránek}
\ead{ijuranek@fit.vutbr.cz}
\author[AddressIT4I]{Adam Herout}
\ead{herout@fit.vutbr.cz}
\address[AddressIT4I]{Brno University of Technology,
	Faculty of Information Technology,\\
	Centre of Excellence IT4Innovations,
	Božetěchova 2,
	612 66 Brno,
	Czech Republic
}

\begin{abstract}
In this paper, we focus on fully automatic traffic surveillance camera calibration, which we use for speed measurement of passing vehicles. We improve over a recent state-of-the-art camera calibration method for  traffic surveillance based on two detected vanishing points. More importantly, we propose a novel automatic scene scale inference method. The method is based on matching bounding boxes of rendered 3D models of vehicles with detected bounding boxes in the image. The proposed method can be used from arbitrary viewpoints, since it has no constraints on camera placement. 
We evaluate our method on the recent comprehensive dataset for speed measurement BrnoCompSpeed. Experiments show that our automatic camera calibration method by detection of two vanishing points reduces error by 50\,\% (mean distance ratio error reduced from 0.18 to 0.09) compared to the previous state-of-the-art method. We also show that our scene scale inference method is more precise, outperforming both state-of-the-art automatic calibration method for speed measurement (error reduction by 86\,\% -- 7.98\,km/h to 1.10\,km/h) and manual calibration (error reduction by 19\,\% -- 1.35\,km/h to 1.10\,km/h). 
We also present qualitative results of the proposed automatic camera calibration method on video sequences obtained from real surveillance cameras in various places, and under different lighting conditions (night, dawn, day).
\end{abstract}


\end{frontmatter}
%\linenumbers

\section{Introduction}

Surveillance systems pose specific requirements on camera calibration.  Their cameras are typically placed in hardly accessible locations and optics are focused at longer distances, making the common pattern-based calibration approaches unusable (such as classical \cite{Zhang2000}). That is why many solutions place markers to the observed scene and/or measure existing geometric features \citep{Sina2013,Do2015,You2016,Luvizon2016}.  These approaches are laborious and inconvenient both in terms of camera setup (manually clicking on the measured features in the image) and in terms of physically visiting the scene and measuring the distances.

In our paper, we focus on \emph{precise} and at the same time \emph{fully automatic} traffic surveillance camera calibration including scene scale for speed measurement. The proposed speed measurement method needs to be able to deal with significant viewpoint variation, different zoom factors, various roads and densities of traffic. If the method should be applicable for large-scale deployment, it needs to run fully automatically without the necessity to stop traffic for installation or for performing calibration measurements.

\begin{figure}[h]
	\centering
	\fbox{\includegraphics[width=0.497\linewidth]{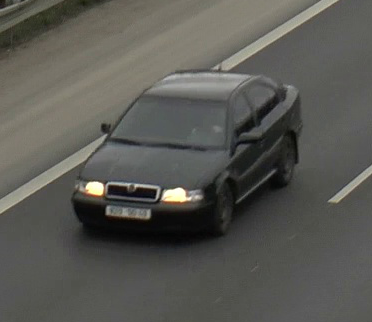}%
		\includegraphics[width=0.497\linewidth]{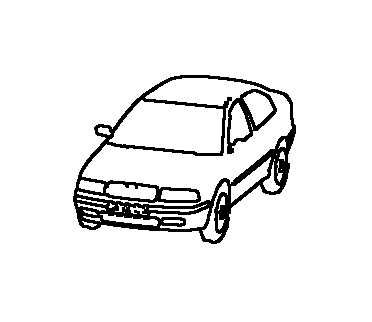}}\\[2pt]
	\includegraphics[height=0.287\linewidth]{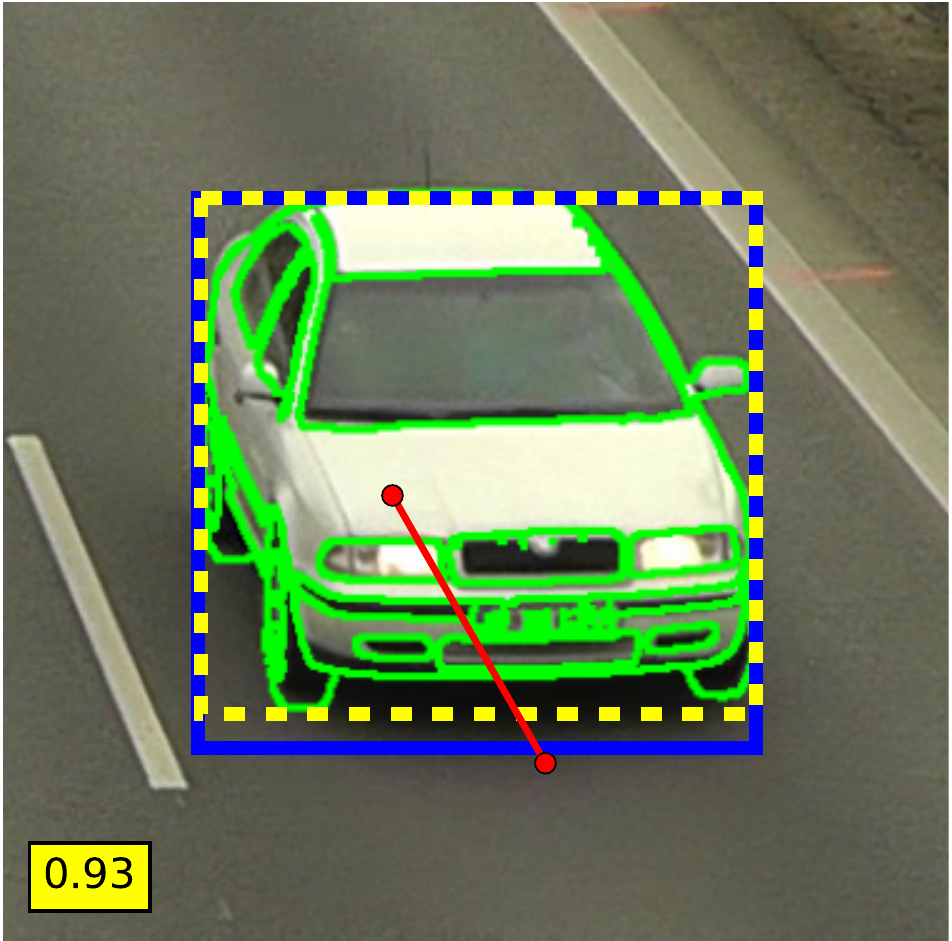}\hfill
	\includegraphics[height=0.287\linewidth]{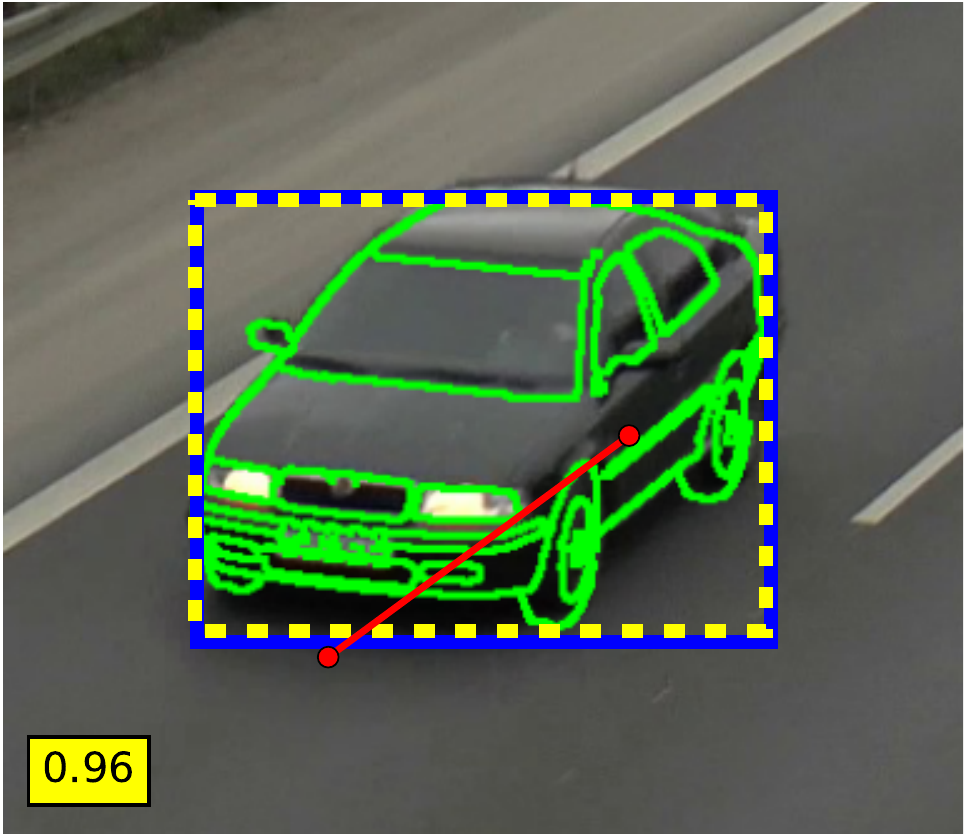}\hfill
	\includegraphics[height=0.287\linewidth]{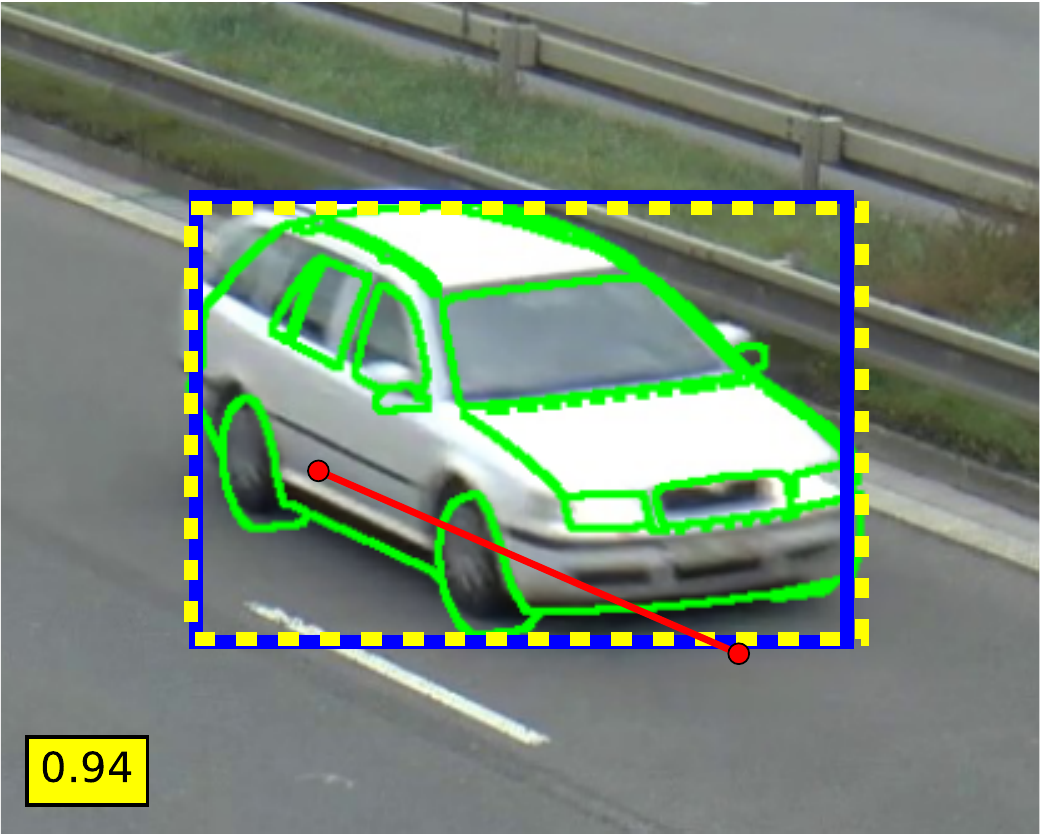}%
	\caption{Examples of detected vehicles and 3D model bounding box aligned to the vehicle detection bounding box. \textbf{top:} detected vehicle and corresponding 3D model (edges only), \textbf{bottom:} examples of aligned bounding boxes with shown 3D model edges (green), its bounding box (yellow) and vehicle detection (blue).}
	\label{fig:Teaser}
\end{figure}
Our solution uses camera calibration obtained from two detected vanishing points and it is built on our previous work \citep{Dubska2014,Dubska2015ITS}. However, this calibration procedure only allows  reconstruction of the rotation matrix and the intrinsic parameters from vanishing points, and it is still necessary to obtain the scene scale.  We propose to detect vehicles on the road by Faster-RCNN \citep{Girshick2015}, classify them into a few common fine-grained types by a CNN \citep{Krizhevsky2012} and use bounding boxes of 3D models for the known classes to align the detected vehicles.  The vanishing point-based calibration allows for full reconstruction of the viewpoint on the vehicle and the only free parameter in the alignment is therefore the scene scale. Figure~\ref{fig:Teaser} shows an example of the 3D model and the aligned images. Our experiments show that our method (mean speed measurement error 1.10\,km/h) significantly outperforms existing automatic camera calibration method by \citet{Dubska2014} (error reduction by 86\,\% -- mean error 7.98\,km/h) and also calibration obtained from manual measurements on the road (error reduction by 19\,\% -- mean error 1.35\,km/h). 
This is important because in previous approaches, automation always compromised accuracy, forcing a trade off by the system developer. Our work shows that fully automatic calibration methods may produce better results than manual calibration (which was performed thoroughly and according to state-of-the-art approaches).

%Our work shows that manual calibration (though laborious, thorough, and carried out according to state-of-the-art approaches) is inferior to the fully automatic approach based on computer vision methods.

%\textbf{Existing Solutions} \blind
Existing solutions for traffic surveillance camera calibration \citep{Dailey2000,Schoepflin2003,Cathey2005,Grammatikopoulos2005,He2007,Maduro2008,Sina2013,Nurhadiyatna2013,Dubska2014,Lan2014,Luvizon2014,Dubska2015ITS,Do2015,Luvizon2016,You2016} (see Section~\ref{sec:SOTASpeedMeasurement} for detailed analysis) usually have limitations for real world applications. They are either limited to some viewpoints (zero pan, second vanishing point at infinity), or they require some per-installed-camera manual work. To our knowledge, there is only one work \citep{Dubska2014} which does not have these limitations, and therefore we compare our results with this solution. For a brief description of the method, see Section~\ref{sec:SOTASpeedMeasurement}; a more comprehensive review can be found in a recent dataset paper BrnoCompSpeed by~\cite{BrnoCompSpeed}.

%\textbf{Contributions} \blind
The key contributions of this paper are:
\begin{itemize}
	\item An improved camera calibration method by detection of two vanishing points. The camera calibration error is reduced by 50\,\% -- 0.18 to 0.09 mean distance ratio error.
	\item A novel method for scene scale inference, which significantly outperforms  automatic traffic camera calibration methods (error reduced by 86\,\% -- 7.98\,km/h to 1.10\,km/h) and also manual calibration (error reduced by~19\,\% -- 1.35\,km/h to 1.10\,km/h) in automatic speed measurement from a monocular camera.
	\item Results show that when used for the speed measurement task, the automatic (zero human input) method can perform better than the laborious manual calibration, which is generally considered accurate and treated as the ground truth. This finding can be important also in other fields beyond traffic surveillance.
\end{itemize}

\section{Related Work} \label{sec:SOTASpeedMeasurement}
% Camera calibration
The camera calibration algorithm (obtaining intrinsic and extrinsic parameters of the surveillance camera) is critical for the accuracy of vehicle speed measurement by a single monocular camera, as it directly influences the speed measurement accuracy. There is a very recent comprehensive review of the traffic surveillance calibration methods \citep{BrnoCompSpeed}, so for detailed information we refer to this review and we include only a brief description of the methods. 

Several methods \citep{He2007,Cathey2005,Grammatikopoulos2005} are based on the detection of vanishing points as an intersection of road markings (lane dividing lines). Other methods \citep{Dubska2014,Dubska2015ITS,Schoepflin2003,Dailey2000} use vehicle motion to calibrate the camera. There is also a set of methods which use some form of manually measured dimensions on the road plane \citep{Maduro2008,Nurhadiyatna2013,Sina2013,Luvizon2014,Luvizon2016,Do2015,Lan2014}.

An important attribute of calibration methods is whether they are able to work automatically without any manual per-camera calibration input. Only two methods \citep{Dailey2000,Dubska2014} are fully automatic and both of them use mean vehicle dimensions for camera calibration. Another important requirement for real-world deployment is whether the camera can be placed in an arbitrary position above the road, which is not true for some methods as they assume to have zero pan or other constraints. 
%\cite{Dubska2014} described one algorithm (to our knowledge the only one so far) that features both these attributes; we will be using this work for comparison.

%\todo{related work -- vehicle classification?} \todo{related work -- detection?} \todo{je nutné o tom psát sem když pouze požíváme jeden systém? -- (Adam) asi stručně jo, ať máme rozumný Related work a ať to dostane kontext == stručně a výstižně.} \blind

Regarding fine-grained vehicle classification, there are several approaches. The first one is based on detected parts of vehicles \citep{Krause2015,Simon2015,Fang2016}, another approach is based on bilinear pooling \citep{Lin2015Bilinear,Gao2016}. There is also an approach based on Convolutional Neural Networks (CNN) and input modification \citep{Sochor2016}. 
For object detection, it is possible to use boosted cascades \citep{Dollar2014}, HOG detectors \citep{Dalal2005}, or Deformable Parts Models (DPMs) \citep{Felzenszwalb2010}. There are also recent advances in object detection based on CNNs \citep{Girshick2014,Girshick2015,Liu2016SSD}.

% 3D model alignment
Several authors deal with alignment of 3D models and vehicles and use this technique for gathering data in the context of traffic surveillance. \cite{Lin2014} propose to jointly optimize 3D model fitting and fine-grained classification, and \cite{Hsiao2014} align edges formulated as an Active Shape Model \citep{Cootes1995,Li2009}. \cite{Krause2013} and propose the use of synthetic data to train geometry and viewpoint classifiers for 3D model and 2D image alignment.
\cite{Prokaj2009} use detected SIFT features \citep{Lowe1999} to align 3D vehicle models and the vehicle's observation.  They use the alignment mainly to overcome vehicle appearance variation under different viewpoints. However, in our case, as the precise viewpoint on the vehicle is known (Section~\ref{sec:ScaleInference}), such alignment does not have to be performed.  Hence, we adopt a simpler and more efficient method based on 2D bounding boxes -- simplifying the procedure considerably without sacrificing the accuracy. 

When it comes to camera calibration in general, various approaches exist. The widely used method by  \cite{Zhang2000} uses a calibration checkerboard to obtain intrinsic and extrinsic camera parameters (relative to the checkerboard). \cite{Liu2012} use controlled panning or tilting with stereo matching to calibrate the camera. Correspondences of lines and points are used by \cite{Chaperon2011}. \cite{Yu2009} focus on automatic camera calibration for tennis videos from detected tennis court lines.

%\todo{ještě něco používáme a mělo by se to objevit?}

\section{Traffic Camera Model} \label{sec:CameraModel}

The main goal of camera calibration in the application of speed measurement is to be able to measure distances on the road plane between two arbitrary points in meters (or other distance units), therefore we only focus on a camera model which enables the measurement of distance between two points on the road plane.

For convenience and better comparison of the methods, we adopt the traffic camera model and notation proposed in previous papers \citep{Dubska2014,Dubska2015ITS}; however, to make the paper self-contained, we briefly describe the model and notation. For intrinsic parameters of our camera model, we assume to have zero pixel skew, and the principal point~$\mathbf{c}$ in the center of the image.
The method also assumes the road section to be flat and straight; the experiments reported in the previous work and our experiments as well show that this requirement is not very strict, because most roads that are not sharply curved locally meet this assumption for practical purposes.

Homogeneous 2D image coordinates are referenced by bold small letters $\mathbf{p} = [p_x, p_y, 1]^T$, points on the image plane $\mathbf{\overline{p}} = [p_x, p_y, f]^T$ in 3D, where $f$ is the focal length, are denoted by small bold letters with overline. Finally, other 3D points (on the road plane) are denoted by bold capital letters $\mathbf{P} = [P_x, P_y, P_z]^T$. 
%Dot and matrix product is denoted by $\cdot$ and cross product by $\times$.

\begin{figure}[t]
	\centering
	\includegraphics[width=0.6\linewidth]{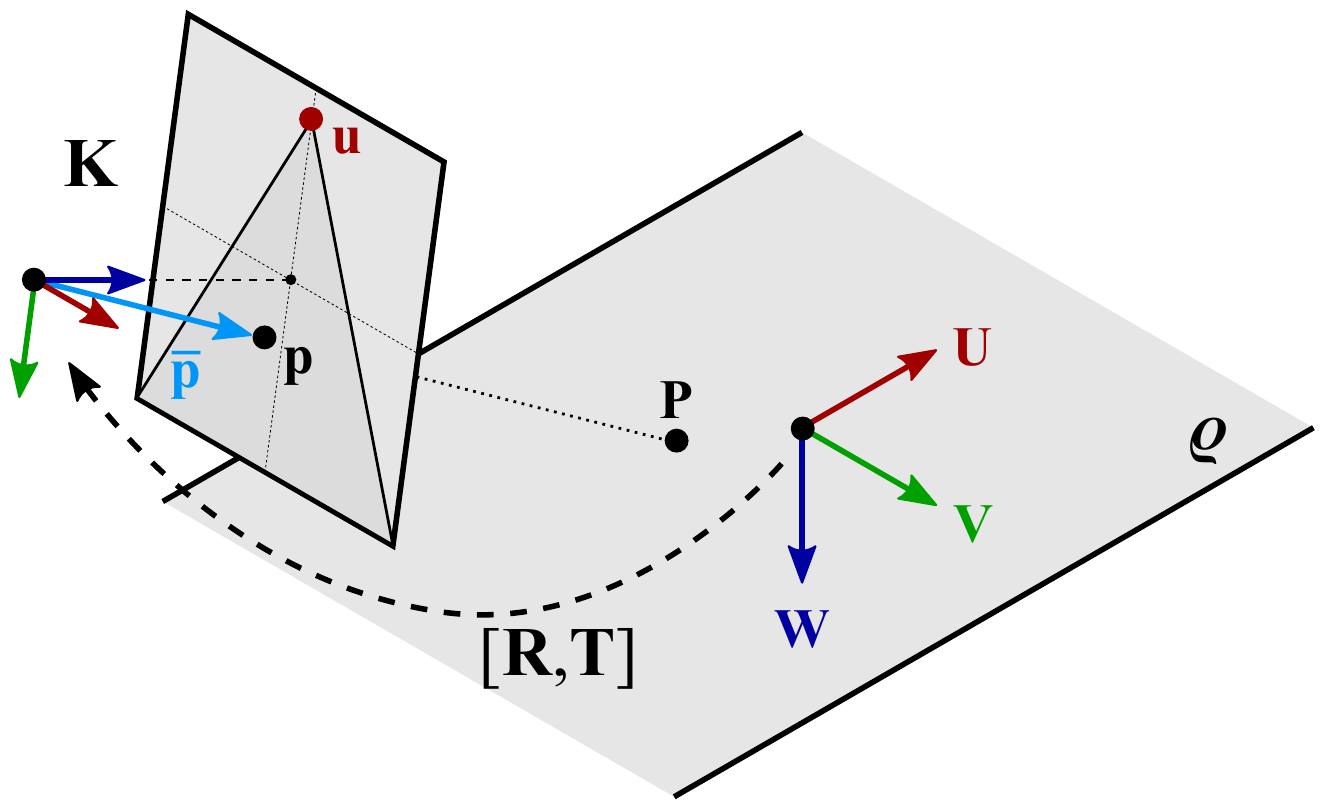}
	\caption{Camera model and coordinates. Points denoted by small letters represent points in image space while points in the world space on the road plane  $\rho$ are represented by capital letters. The representation stays the same for both finite and ideal points.}
	\label{fig:coords}
\end{figure}

Figure \ref{fig:coords} shows the camera model and its notation. For convenience, we assume that the origin of the image coordinate system is at the center of the image; therefore, the principal point $\mathbf{c}$ has 2D homogeneous coordinates $[0,0,1]^T$ (3D coordinates of the center of camera projection are $[0,0,0]^T$). As it is shown, the road plane is denoted by $\boldsymbol{\rho}$. We encode vanishing points in the following way.  The first one (in the direction of vehicle flow) is referenced as $\mathbf{u}$;
the second vanishing point (whose direction is perpendicular to the first one and which is parallel to the road plane) is denoted by $\mathbf{v}$; and the third one (direction perpendicular to the road plane) is $\mathbf{w}$.

Using the first two vanishing points $\mathbf{u}$, $\mathbf{v}$ and the principal point $\mathbf{c}$, it is possible to compute the focal length $f$, the third vanishing point $\mathbf{w}$, the road plane normalized normal vector $\mathbf{n}$, and the road plane $\boldsymbol{\rho}$. However, the road plane is computed only up to scale (as it is not possible to recover the distance to the road plane only from the vanishing points) and therefore, we add an arbitrary value $\delta = 1$ as the constant term in Equation~\eqref{eq:RoadPlane}.
\begin{eqnarray}
	\label{eq:focal}
	f &=& \sqrt{-\mathbf{u}^T \cdot \mathbf{v}}\\
	\mathbf{\overline{u}} &=& [u_x, u_y, f]^T\\
	\mathbf{\overline{v}} &=& [v_x, v_y, f]^T\\
	\mathbf{\overline{w}} &=& \mathbf{\overline{u}} \times \mathbf{\overline{v}} \\
	\mathbf{n} &=& \frac{\mathbf{\overline{w}}}{\|\mathbf{\overline{w}}\|}\\
	\boldsymbol{\rho} &=& \left[\mathbf{n}^T,\delta\right]^T \label{eq:RoadPlane}
\end{eqnarray}

%\todo{Transformace bodu na obrazovce do 3D na road plane}
With known road plane $\boldsymbol{\rho}$, it is possible to compute 3D coordinates $\mathbf{P} = [P_x, P_y, P_z]^T$  of an arbitrary point $\mathbf{p}~=~[p_x,p_y,1]^T$ by projecting it onto the road plane using the following equations:
\begin{eqnarray}
\mathbf{\overline{p}} &=& [p_x, p_y, f]^T\\
\mathbf{P} &=&  - \frac{\delta}{\left[\mathbf{\overline{p}}^T, 0\right] \cdot \boldsymbol{\rho}} \mathbf{\overline{p}}  \label{eq:3DCoords}
\end{eqnarray}

It is possible to measure distances on the road plane directly with 3D coordinates $\mathbf{P}$; however, as the road plane is shifted to a predefined distance by a constant term, the distance $\| \mathbf{P}_1 - \mathbf{P}_2 \|$ between points $\mathbf{P_1}$ and $\mathbf{P_2}$ is not directly expressed in meters (or other real-world units of distance). Therefore, it is necessary to introduce another calibration parameter, referred to as the scene scale $\lambda$, which converts the distance $\| \mathbf{P}_1 - \mathbf{P}_2 \|$ from pseudo-units on the road plane to meters by scaling the distance to $\lambda \| \mathbf{P}_1 - \mathbf{P}_2 \|$.

Under the assumptions that the principal point is in the center of the image and zero pixel skew, it is necessary for the calibration method to compute two vanishing points ($\mathbf{u}$ and $\mathbf{v}$ in our case) together with the scene scale $\lambda$, yielding 5 degrees of freedom.
Methods to convert these camera parameters to the standard intrinsic and extrinsic camera model $\mathbf{K~[R~T]}$ have been discussed before in several papers \citep{Zhang2013Calib,Fung2003,Zheng2014}, therefore we refer to them.

\section{Camera Calibration and Vehicle Tracking} \label{sec:Methodology}

We adopted the calibration method of \cite{Dubska2014}, which gives the image coordinates of the vanishing points and scene scale information. We improved the method with more precise detection of the vanishing points, and we infer the scene scale by using 3D models of frequently passing cars.

Our method measures the speed of passing cars detected by Faster-RCNN \citep{Girshick2015} and tracked by a combination of background subtraction and Kalman filter \citep{Kalman1960} assisted by the detector. This method, more sophisticated than the previous method \citep{Dubska2014}, gives fewer false positives and a comparable recall rate. In the case of very dense flow when vehicles overlap each other in the camera image (which does rarely occur even in real conditions), our method would miss some of the cars as we target free-flow conditions.
In the following text, we describe the components of the method in detail, and evaluate it in Section~\ref{sec:Experiments}.

\subsection{Vanishing Point Estimation from Edgelets} \label{sec:VPEstimation}
We adopted the algorithm proposed by \cite{Dubska2015ITS} (based on the detection of two orthogonal vanishing points) for the detection of the first vanishing point and propose to use a similar algorithm for detecting the second vanishing point. However, we improved the detection of the second vanishing point by using edgelets instead of image gradients used in the previous paper \citep{Dubska2015ITS}.  This change, although subtle, improves the calibration and speed measurement considerably, as the results in Section~\ref{sec:SpeedMeasurementEval} show.

We start with the detection of vanishing points from which the camera rotation with respect to the road can be estimated. The first vanishing point $\mathbf{u}$ is estimated from the movement of the vehicles by a form of cascaded Hough Transform \citep{Dubska2015ITS} of lines formed by tracking points of interest on the moving vehicles. This is a more stable approach than finding the closest point to the lines in an algebraic way, because it is more robust to tracking noise and it is not influenced by vehicles that change lane (and therefore, the vanishing point of their movement is different from the rest of the vehicles).  Similarly to \cite{Dubska2015ITS}, we use the Min-eigenvalue point detector \citep{Shi1994} and the KLT tracker \citep{Tomasi1991}.

\begin{figure}[t]
    \centering
    \includegraphics[height=2.9cm]{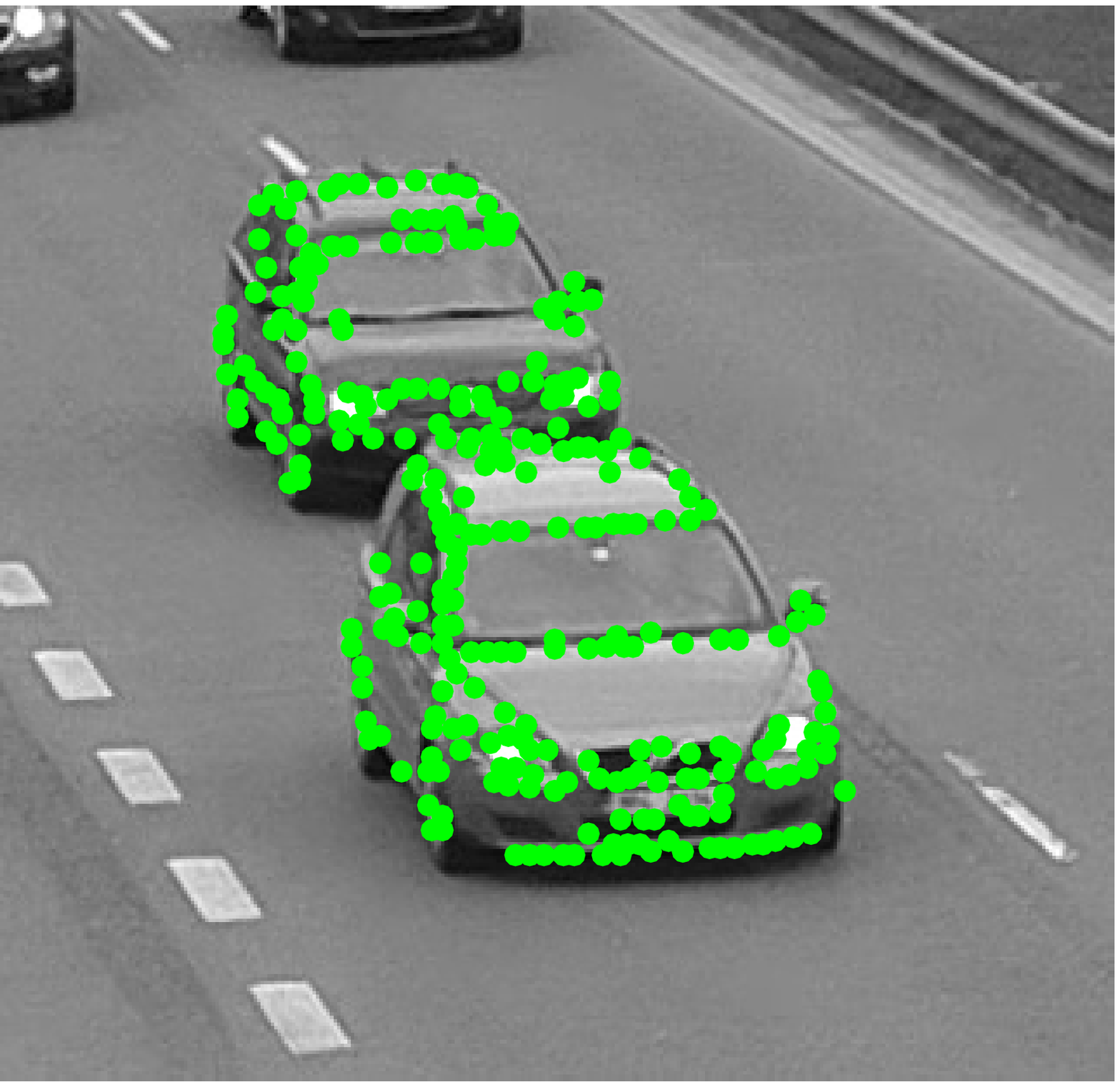}
    \includegraphics[height=2.9cm]{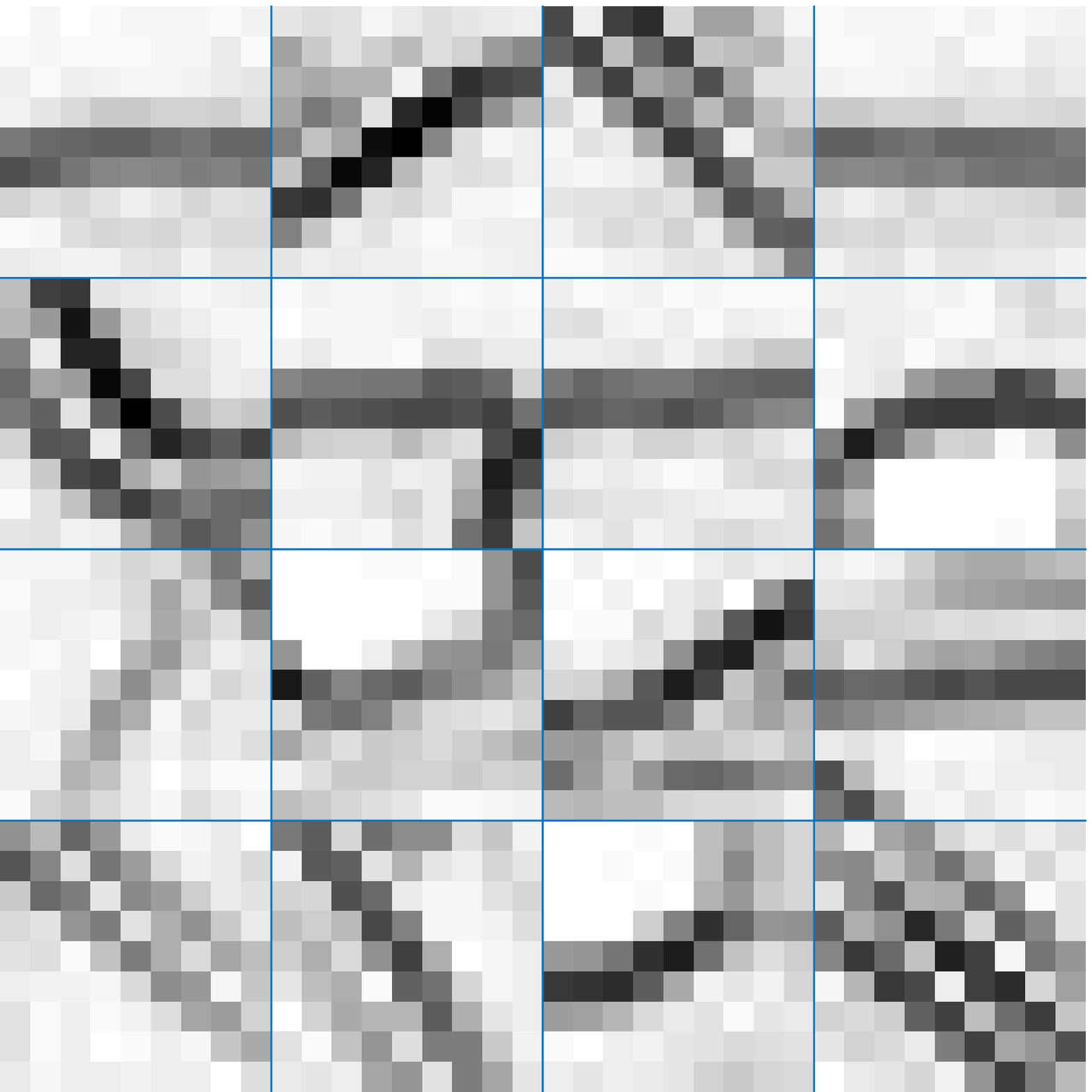}
    \includegraphics[height=2.9cm]{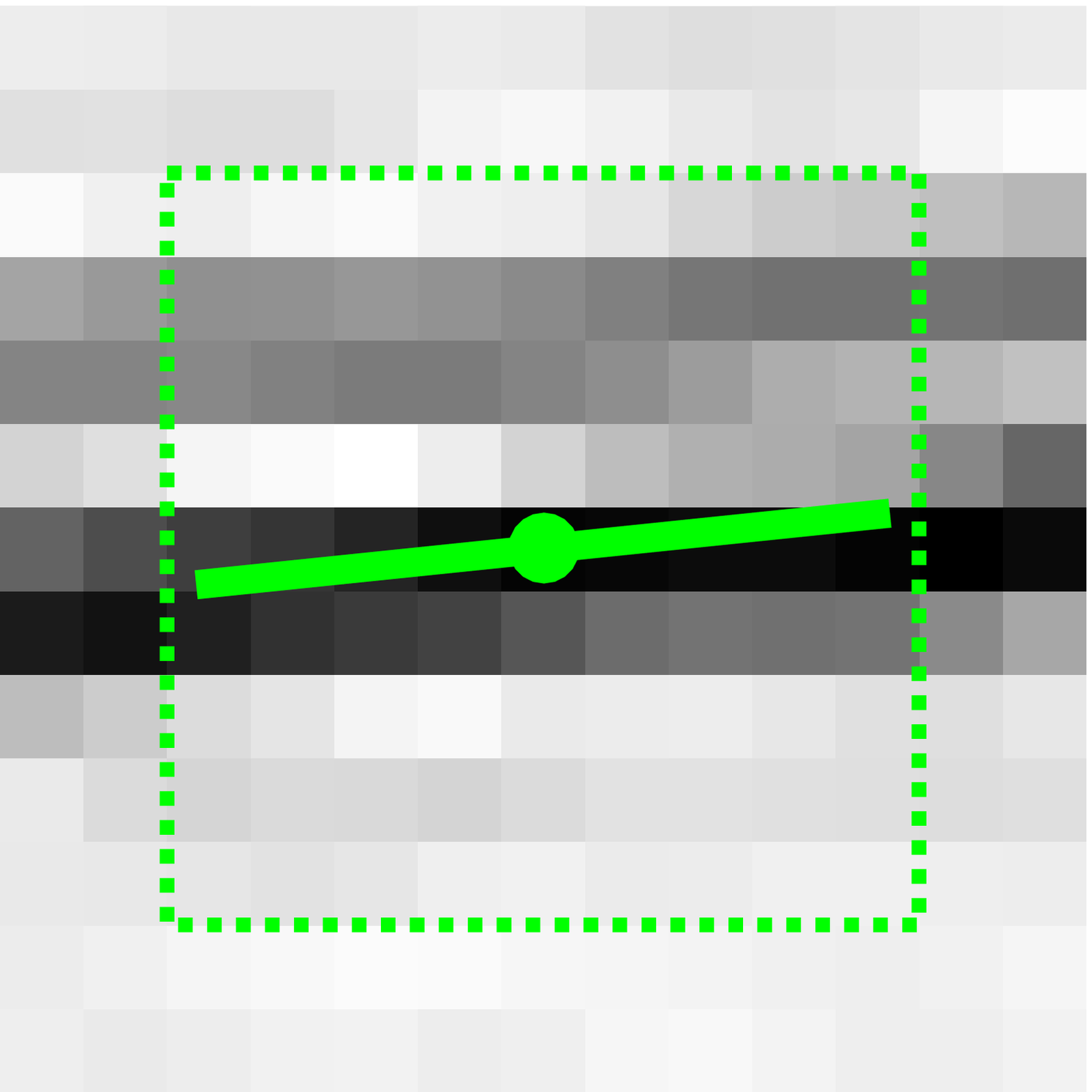}
    \includegraphics[height=2.9cm]{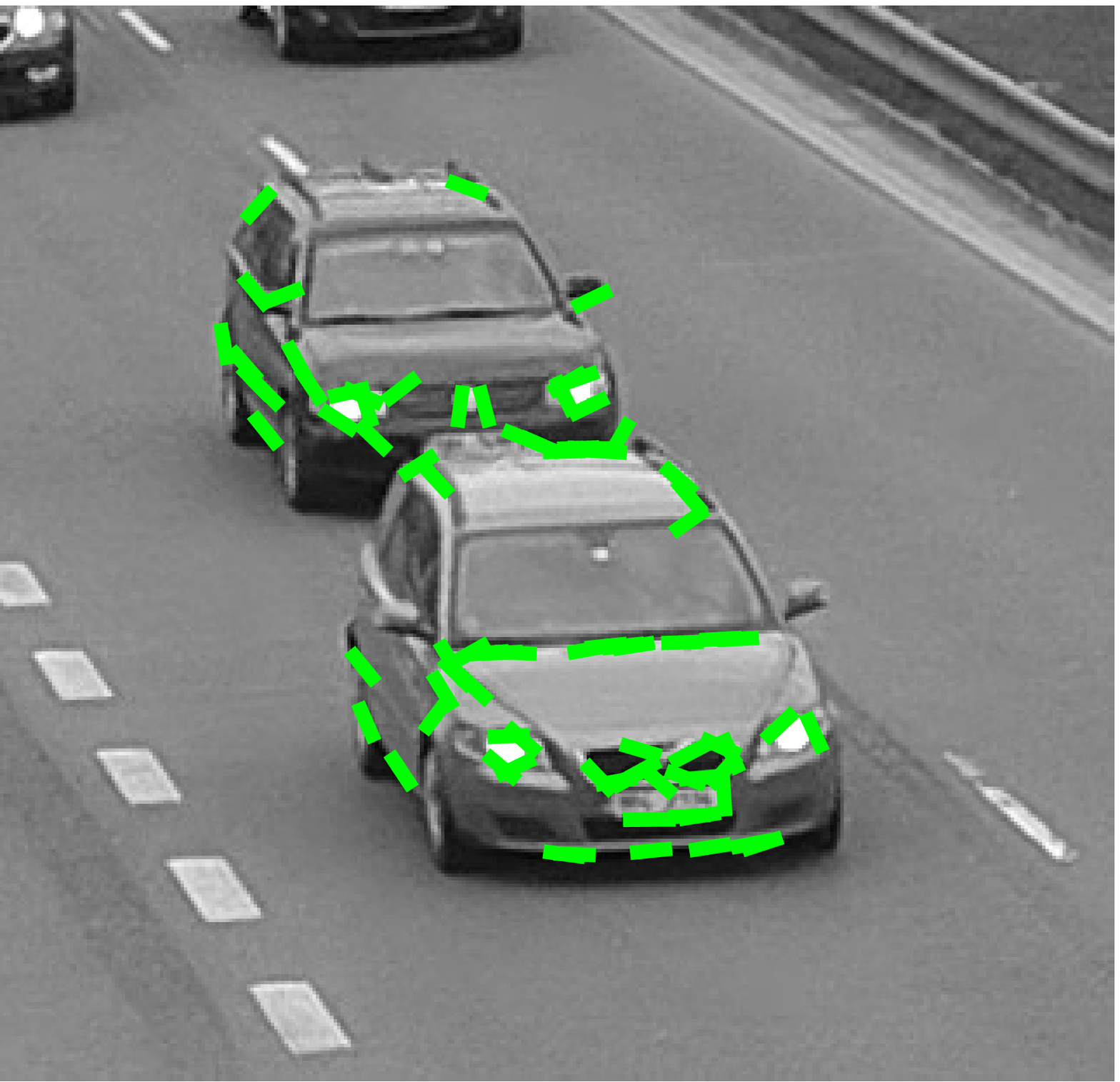}
    \caption{
        Visualization of edgelet detection. From left to right -- Seed points $\mathbf{s}_i$ as local maxima of image gradient (foreground mask was used to filter interesting areas); Patches gatherded around the seed points from which the edge orientation is computed; Detail of an edgelet and its orientation superimposed on the gradient image; Top 25\,\% of edgelets detected in the image.
    }
    \label{fig:Edgelets}
\end{figure}

%Detekce Druheho VP - Edgelets
To detect the second vanishing point $\mathbf{v}$ we use edges on passing vehicles as many lines formed by the edges coincide with~$\mathbf{v}$.
This step heavily relies on the correct estimation of the orientation of the edges. The angle can be easily computed from gradients, but angles close to $k \pi/2$ are almost impossible to accurately recover on small neighborhoods. We estimate edge orientation from a larger neighborhood by analysis of the shape of image gradient magnitude (edgelets). The detection process is shown in Figure~\ref{fig:Edgelets}.

Edgelets are detected by the following algorithm. Given an image $\mathbf{I}$, first, we find seed points $\mathbf{s}_i$ as local maxima of gradient magnitude of the image $\mathbf{E} = \|\nabla{\mathbf{I}}\|$, keeping only the strong ones with magnitudes above a threshold. From the $9\times{}9$ neighborhood of each seed point $\mathbf{s}_i = [x_i, y_i, 1]^T$, matrix $\mathbf{X}_i$ is formed:
\begin{eqnarray}
	&\mathbf{X}_i =
	\left[
	\begin{array}{cc}
		w_1 (m_1 - x_i) & w_1 (n_1 - y_i) \\
		w_2 (m_2 - x_i) & w_2 (n_2 - y_i) \\
		\vdots & \vdots \\
		w_k (m_k - x_i) & w_k (n_k - y_i) \\
	\end{array}
	\right]&
\end{eqnarray}
where $[m_k, n_k,1]^T$ are coordinates of the neighboring pixels ($k=1\dots81$) and $w_k$ is their gradient magnitude from $\mathbf{E}$, i.e. for a $9\times9$ neighborhood, the size of $\mathbf{X}_i$ is $81\times2$. Then, singular vectors and values of $\mathbf{X}_i$ can be computed as:
\begin{eqnarray}
	\label{eq:svd}
	\mathbf{W}_i \mathbf{\Sigma}_i^{2} \mathbf{W}_i^{T}  &=& \mathrm{SVD}\left(\mathbf{X}_i^{T}\mathbf{X}_i\right), \\
    \mathrm{where}&& \nonumber \\
	\label{eq:eigenvectors}
	\mathbf{W}_i &=& \left[ \mathbf{a}_1, \mathbf{a}_2 \right] \\
	\label{eq:eigenvalues}
	\mathbf{\Sigma}_i &=&
	\left(
	\begin{array}{cc}
		\lambda_1 & 0 \\
		0 & \lambda_2 \\
	\end{array}
	\right).
\end{eqnarray}
Vectors $\mathbf{a}_1$ and $\mathbf{a}_2$ represent the eigenvectors of $\mathbf{X}_i$, while $\lambda_1$ and $\lambda_2$ denote the corresponding eigenvalues. Edge orientation is then the first singular column vector $\mathbf{d}_i = \mathbf{a}_1$ from \eqref{eq:eigenvectors} and the edge quality is the ratio of singular values $q_i = \frac{\lambda_1}{\lambda_2}$ from~\eqref{eq:eigenvalues}. Each edgelet is then represented as a triplet $\mathcal{E}_i = \left( \mathbf{s}_i, \mathbf{d}_i, q_i \right)$.

\begin{figure*}[t]
	\centering
	\includegraphics[width=\linewidth]{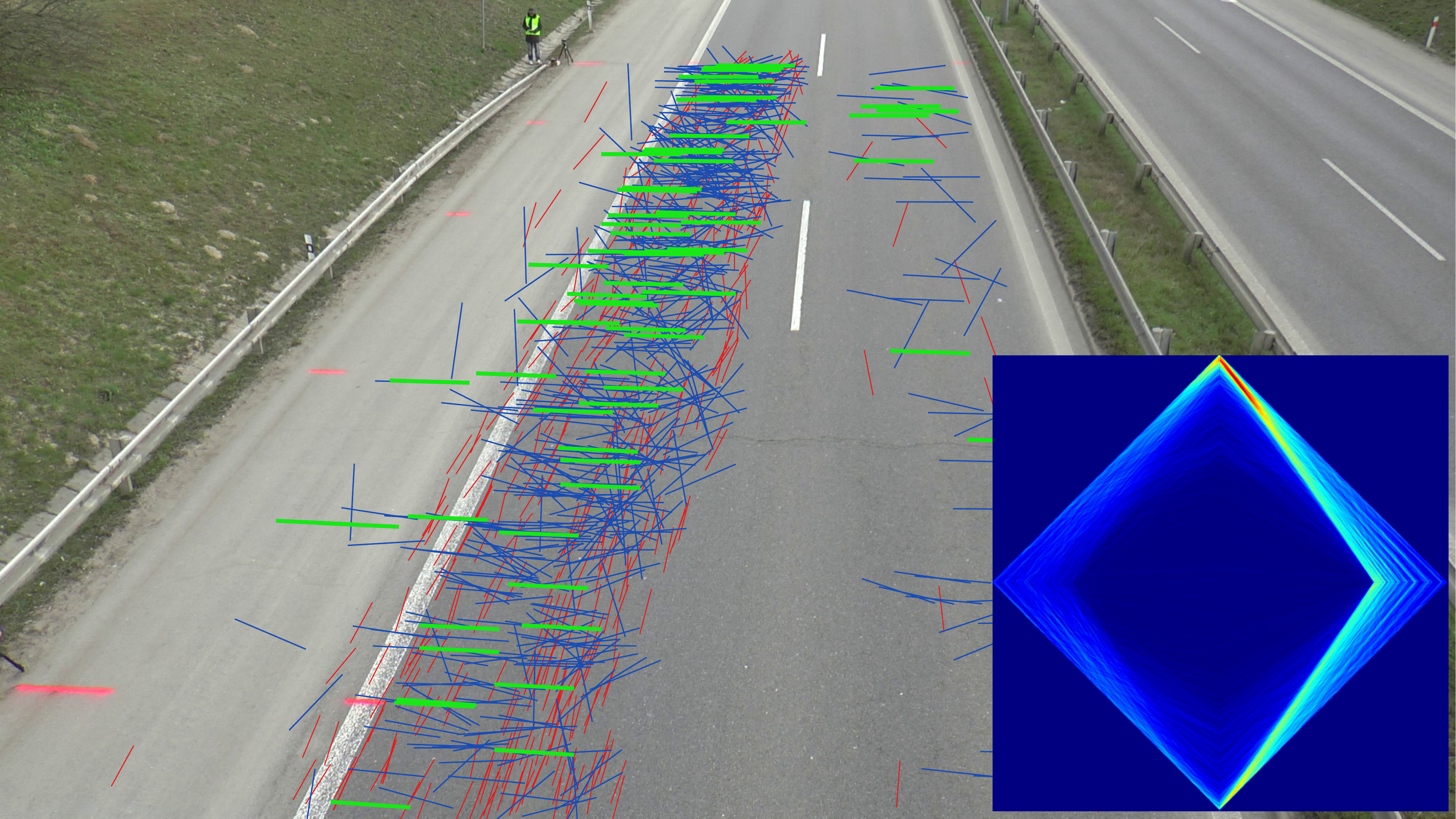}
	\caption{Visualization of edges gathered from a video -- \textbf{(red)} edges that pass close to the first vanishing point, \textbf{(blue and green)} edges accumulated to the Diamond Space, and \textbf{(green)} edges supporting the detected second vanishing point. The corresponding Diamond Space is shown in bottom-right corner.
	} \label{fig:edgeletsvideo}
\end{figure*}

We gather the edgelets from the input video (see Figure \ref{fig:edgeletsvideo}), keeping only the strong ones which do not coincide with the already estimated $\mathbf{u}$, and accumulate them to the Diamond Space accumulator \citep{Dubska2013}. The position of the global maximum in the accumulator is taken as the second vanishing point $\mathbf{v}$. It should be noted that in this step, additional filtering can be applied -- e.g. masking the Diamond Space to find only plausible solutions (i.e. avoid imaginary focal length from Equation~\eqref{eq:focal}), or to find solutions within a certain range of focal lengths or horizon inclinations (when known in advance). This may improve the robustness of the second vanishing point estimation.

\subsection{Vehicle Detection and Tracking} \label{sec:DetectionTracking}
\begin{figure}[t]
	\centering
	\includegraphics[width=0.325\linewidth]{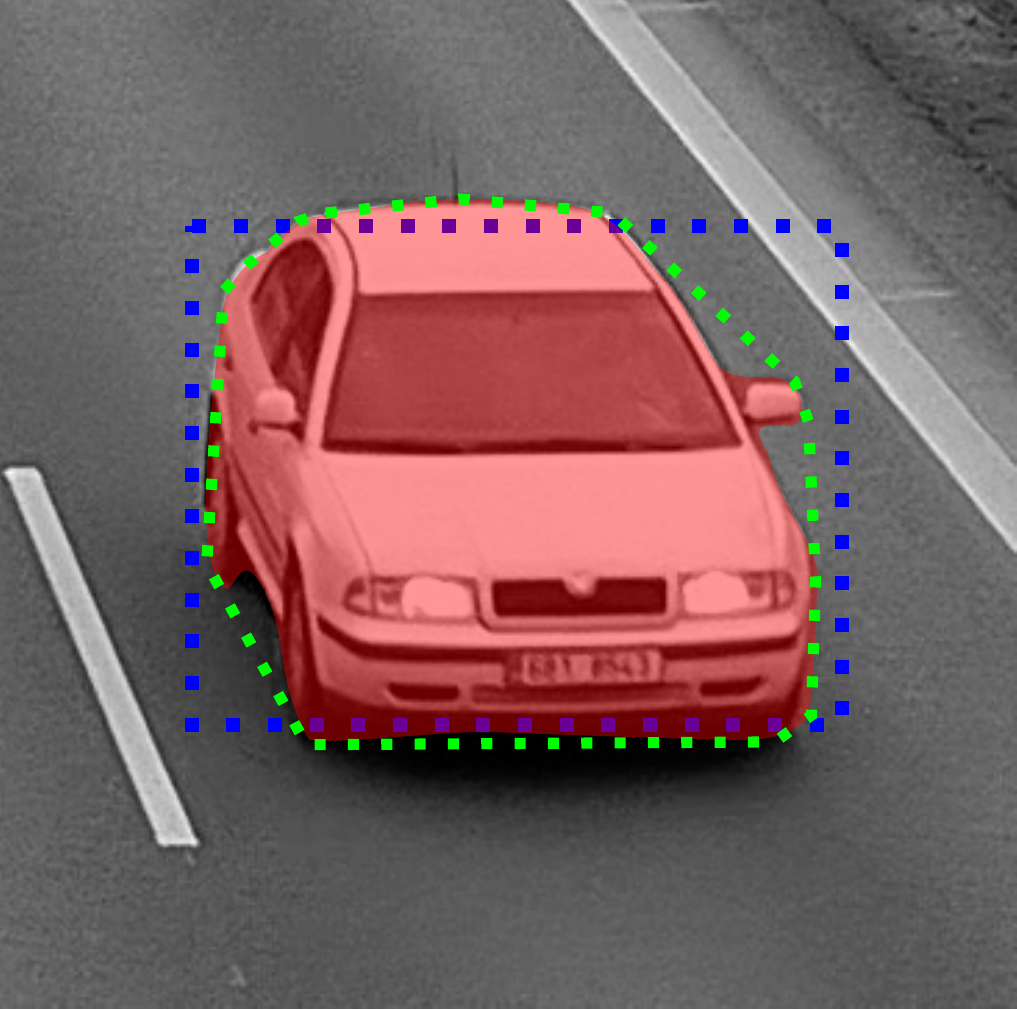}\hfill
	\includegraphics[width=0.325\linewidth]{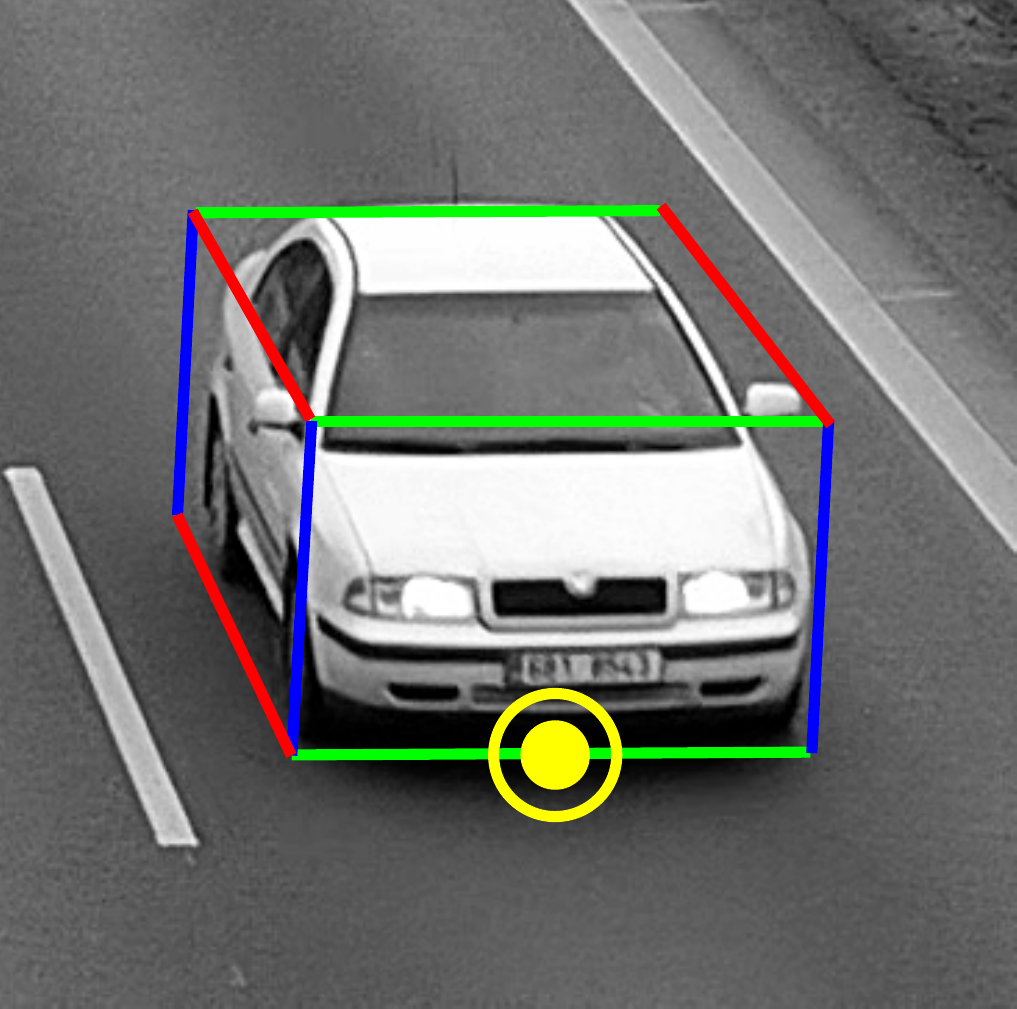}\hfill
	\includegraphics[width=0.325\linewidth]{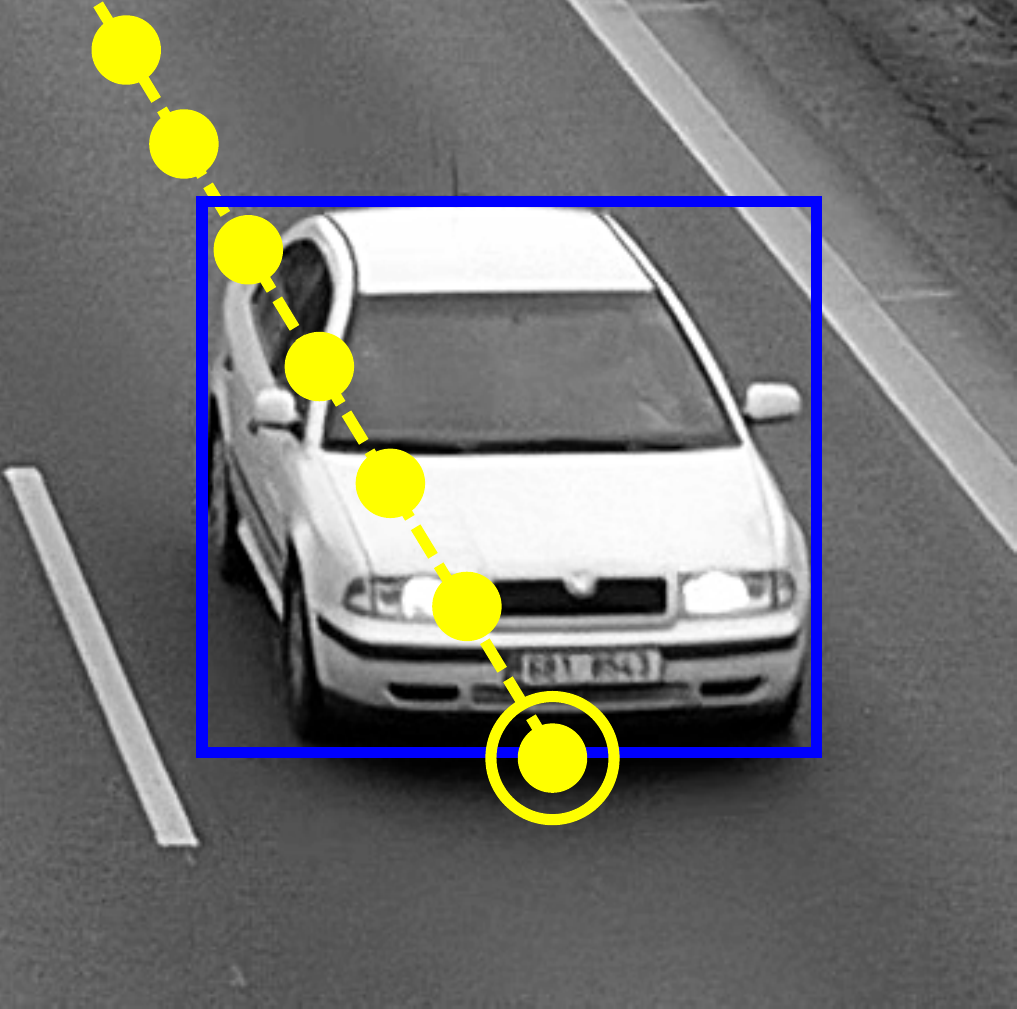}
	\caption{Car detection and tracking. From left to right: Car detected by FRCN (blue), its foreground mask and convex hull (green); 3D bounding box constructed around the convex hull and tracking point on the bottom front edge; Car bounding box (from the convex hull) tracked by Kalman filter.}
	\label{fig:DetectionTracking}
\end{figure}
% Detekce - FRCN, odkud mame data, neni potreba moc rozpitvavat
During speed measurement, passing cars are detected in each frame by the Faster-RCNN (FRCN) detector \citep{Girshick2015} but any detector can be used as well (e.g. ACF, LDCF \citep{Dollar2014}). We trained the detector on the COD20K dataset \citep{Juranek2015}, which contains approximately 20\,k car instances for training from views of surveillance nature. The detection rate of the detector is 96\,\% with 0.02 false positive detections per image on the test part of the COD20K dataset. The detector yields a coarse information about locations of cars in the image (bounding boxes are not precisely aligned).
%Vyrazeni detekci, ktere se mohou prekryvat - bereme tu blizsi. mame nizsi recall, ale take detekce u kterych jsme schoni spolehlive urcit trajektorie
We use a simple heuristic to remove detections that would lead to imprecise tracking and ultimately to wrong speed estimation -- those that are slightly occluded by other detections and that are farther from the camera. Therefore we track only cars that are fully visible.

%Tracking - bloby, spojovani pomoci detekce, convex hull, kalman filtering - tracky pro jednotlive instance
For the tracking itself, we use a simple background model that builds a background reference image by moving average. In the foreground image, compact blobs are detected and the FRCN detections are used to group those blobs that correspond to one car. From each group of blobs, the convex hull and its 2D bounding box are extracted. Finally, we track the 2D bounding box of the convex hull using a Kalman filter to get the movement of the car. For an example, see Figure~\ref{fig:DetectionTracking}.

%Vyber bodu pro mereni - 3D bbox, stred predni hrany.
For each tracked car, we extract a reference point for speed measurement. The convex hull is used to construct the 3D bounding box \citep{Dubska2014} and we take the center of the bottom-front edge -- the reference point located in the ground/road plane. Each track is represented by a sequence of bounding boxes and reference points both constructed from the convex hull.
Our method inherits all the advantages and limitations of similar approaches based on the extraction of the vehicle's foreground mask.  We rely on the extractor to do its job properly, and we can take advantage of works dealing with different issues related to for example lighting and weather (for example contour extractors such as \cite{Yang_2016_CVPR}, or semantic segmentation methods such as \cite{Long_2015_CVPR}). In Section~\ref{sec:RealCameras}, we show a number of examples of real-world surveillance cameras under bad conditions, where the calibration algorithm nonetheless works well.

\subsection{Scale Inference using 3D Model Bounding Box Alignment} \label{sec:ScaleInference}
The previous state-of-the-art automatic method for scale inference in traffic surveillance by \cite{Dubska2014} used three-dimensional bounding boxes built around the vehicle and mean dimensions of vehicles to compute the scale. However, this approach has two main drawbacks. The obvious one is in the usage of mean dimensions of vehicles. 
However, the more important one is less obvious: the constructed bounding box is too tight around the vehicle and the tightness is largely influenced by the particular viewpoint direction.  This causes systematic errors in the calibration depending on the camera location with respect to the road, leading to high sensitivity to viewpoint change.

We propose to use a different approach for scale inference, overcoming the mentioned imprecisions.  We use fine-grained types of vehicles (i.e. make, model, variant, model year) and for a few (two in our experiments) common types we obtained 3D models which are rendered to the image and we align them to the real observed vehicles in order to obtain the proper scale.

%training tracks: 22881, images: 92019
As it is necessary to know the precise vehicle classes (up to model year) for our scale inference method, we used the BoxCars dataset \citep{Sochor2016} and we also collected some other training data from videos related to papers by \cite{Dubska2014,Dubska2015ITS}.
The classification of vehicles is done only into a few most common fine-grained vehicle types on roads in the area plus one class for all the others vehicles. The full training dataset contained $\sim$23\,k  tracks and $\sim$92\,k images of vehicles. We used a CNN \citep{Krizhevsky2012} for the classification itself. The classification accuracy on the validation set ($\sim$7\,k of images) was $0.97$. As only single instances of vehicles are classified by the CNN, we use mean probability over all of the detections belonging to one vehicle track to improve the recognition rates.

For each vehicle, we also build a 3D bounding box around it \citep{Dubska2014} to obtain the center $\mathbf{b}$ of the vehicle's base in image coordinates. To obtain the viewpoint vector $\boldsymbol{\phi}$, we first compute the rotation matrix $\mathbf{R}$, which has columns equal to normalized $\mathbf{\overline{u}}$, $\mathbf{\overline{v}}$, and $\mathbf{\overline{w}}$. It is then possible to compute the 3D viewpoint vector as  $\boldsymbol{\phi}=-\mathbf{R}^T\mathbf{\overline{b}}$. The minus sign is necessary as we need the viewpoint vector going from the vehicle to the camera, not the opposite one. 

\begin{figure*}[t]
    \centering
    %('session0', 'left', 69),  ('session4', 'center', 512), ('session4', 'right', 1640)
    \fbox{\includegraphics[height=0.14\linewidth]{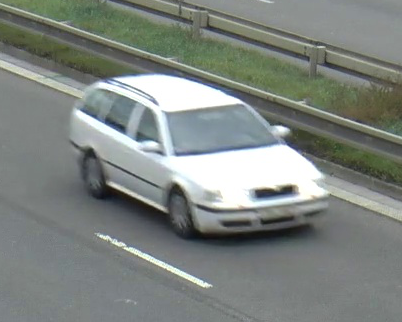}%
        \includegraphics[height=0.14\linewidth]{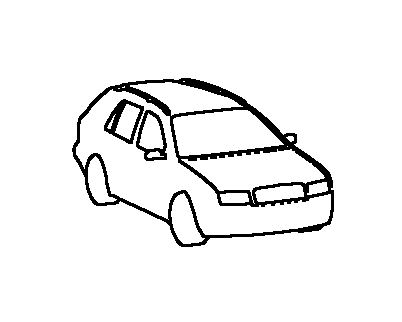}}\hfill
    \fbox{\includegraphics[height=0.14\linewidth]{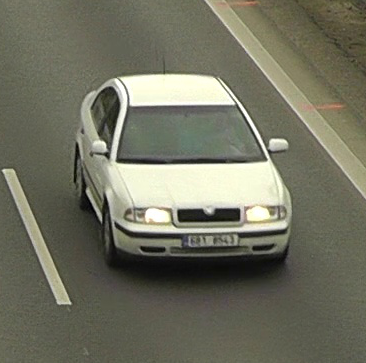}%
        \includegraphics[height=0.14\linewidth]{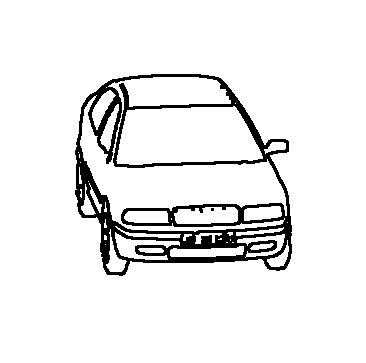}}\hfill
    \fbox{\includegraphics[height=0.14\linewidth]{images_render_real_real_1640_185288.png}%
        \includegraphics[height=0.14\linewidth]{images_render_real_render_1640_185288.png}}
    \caption{Examples of used 3D models (showing only edges) rendered under the same viewpoint as the corresponding real vehicle on the road. The left image shows the model which we will refer as Combi and the other two images show the 3D model Sedan. Both models are for Skoda Octavia mk1 which is common on the observed streets.}
    \label{fig:RenderVsReal}
\end{figure*}
\begin{figure*}[t]
    \centering
    \includegraphics[width=0.2\linewidth]{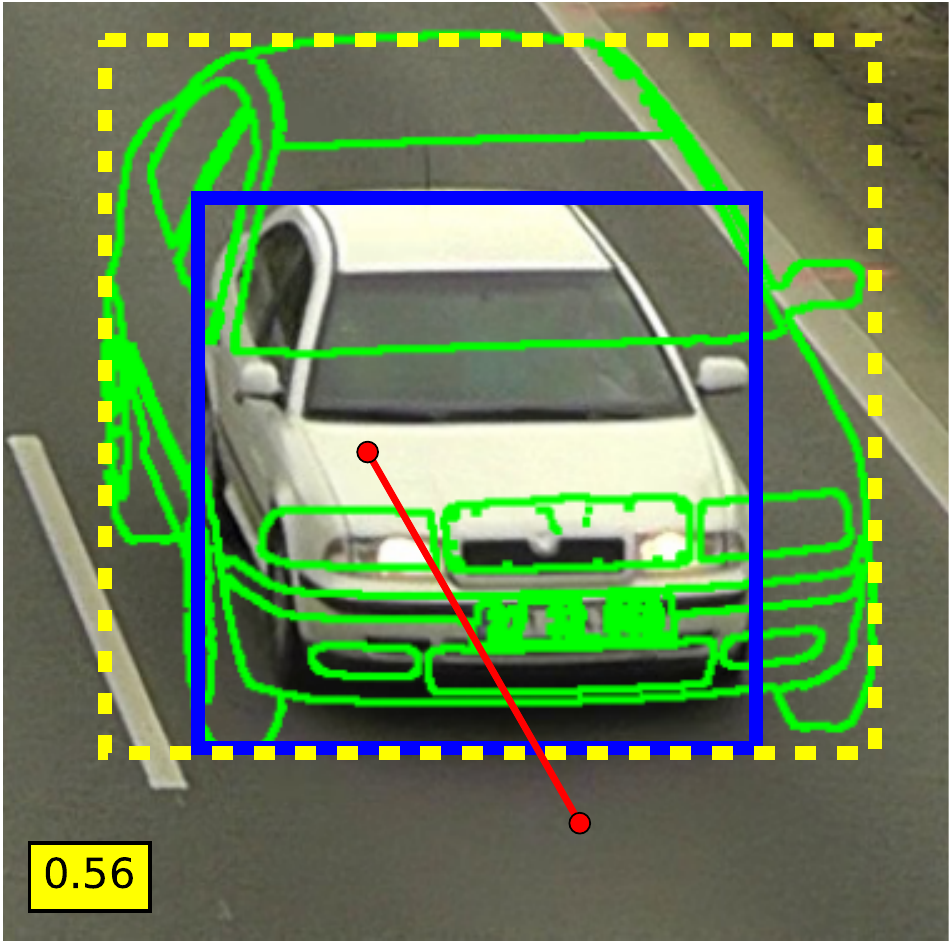}%
    \includegraphics[width=0.2\linewidth]{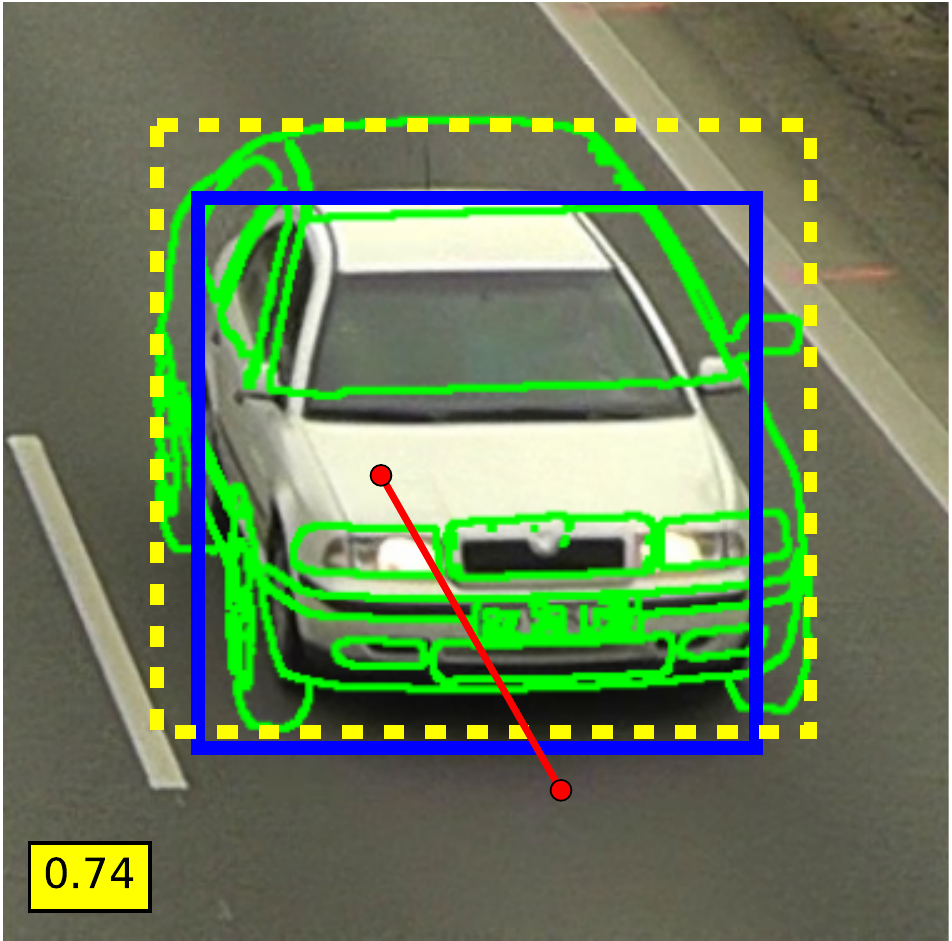}%
    \includegraphics[width=0.2\linewidth]{images_bb_iou_development_512_56134_031.pdf}%
    \includegraphics[width=0.2\linewidth]{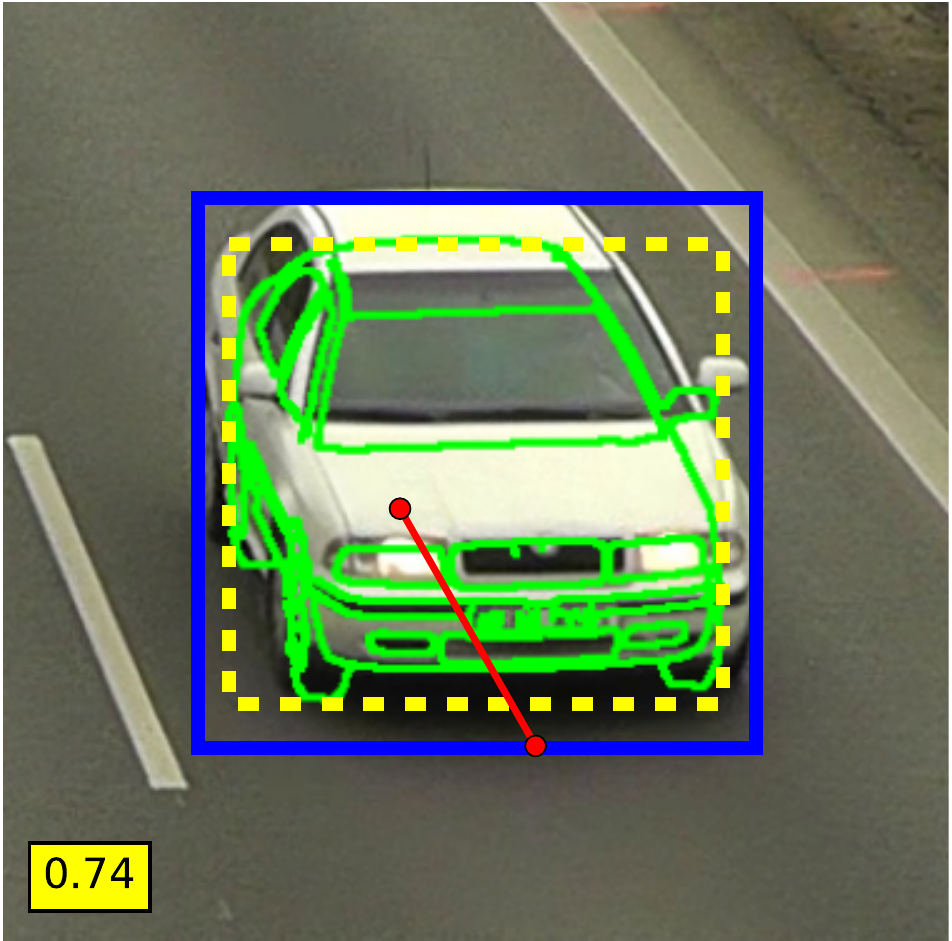}%
    \includegraphics[width=0.2\linewidth]{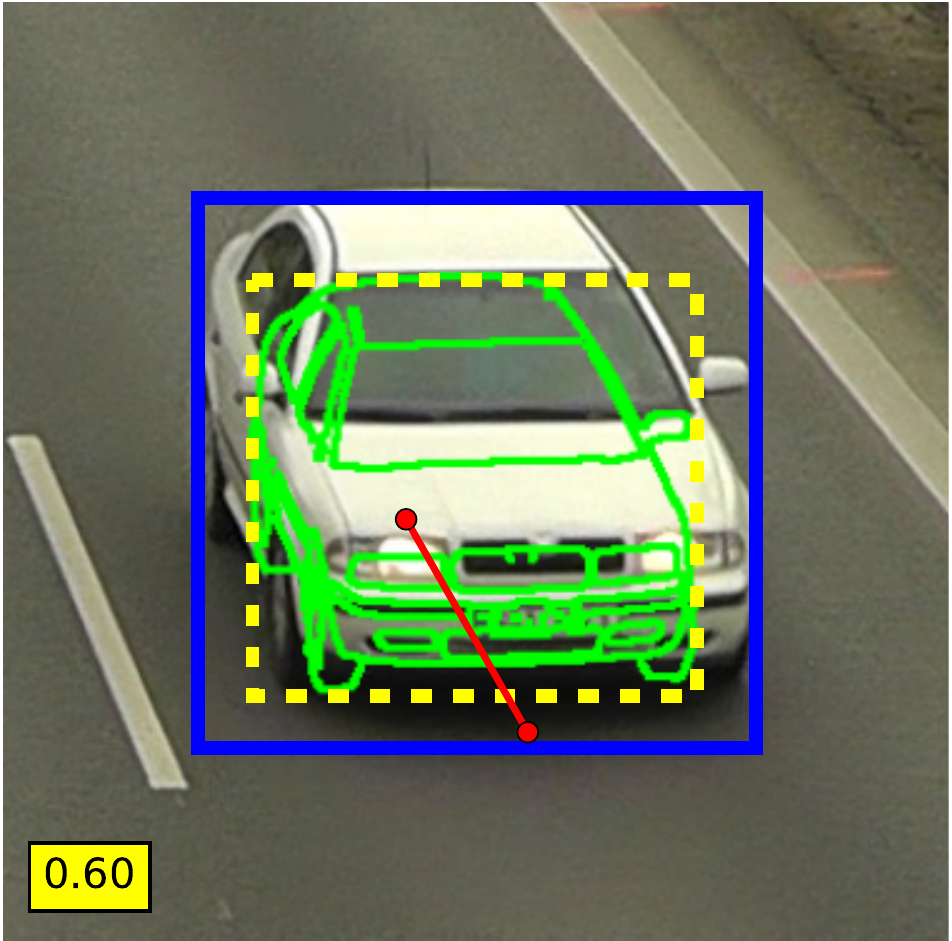}\\
    \includegraphics[width=0.2\linewidth]{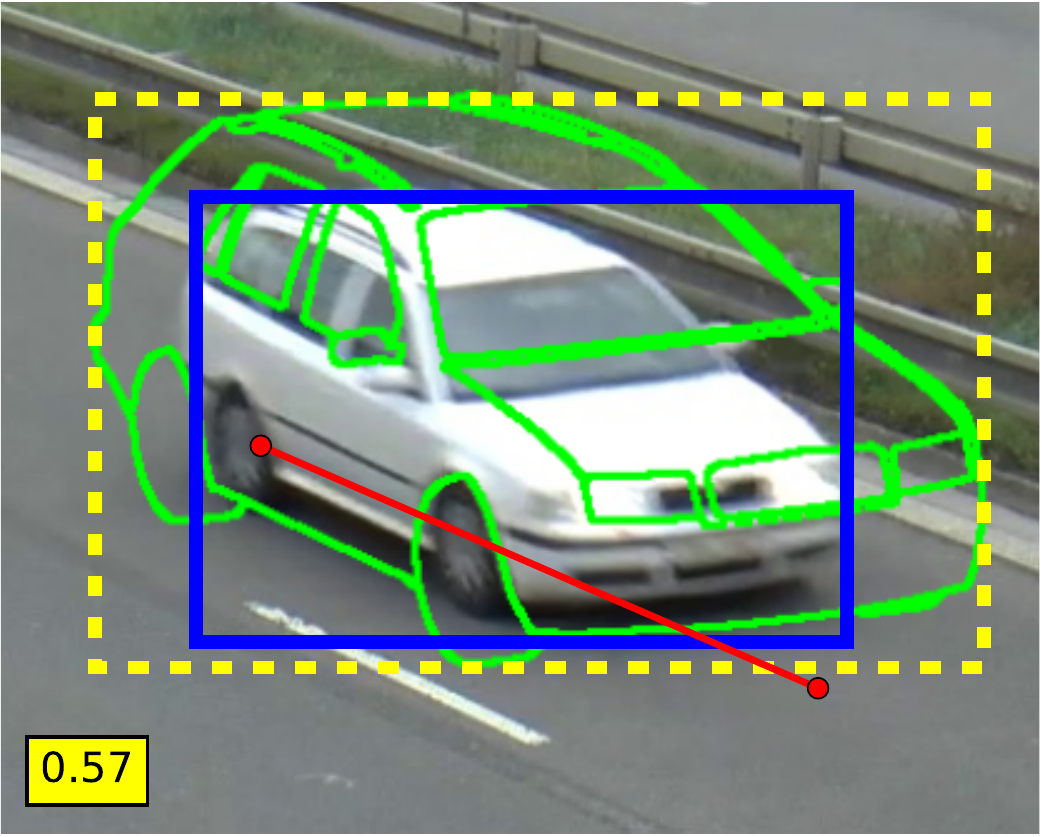}%
    \includegraphics[width=0.2\linewidth]{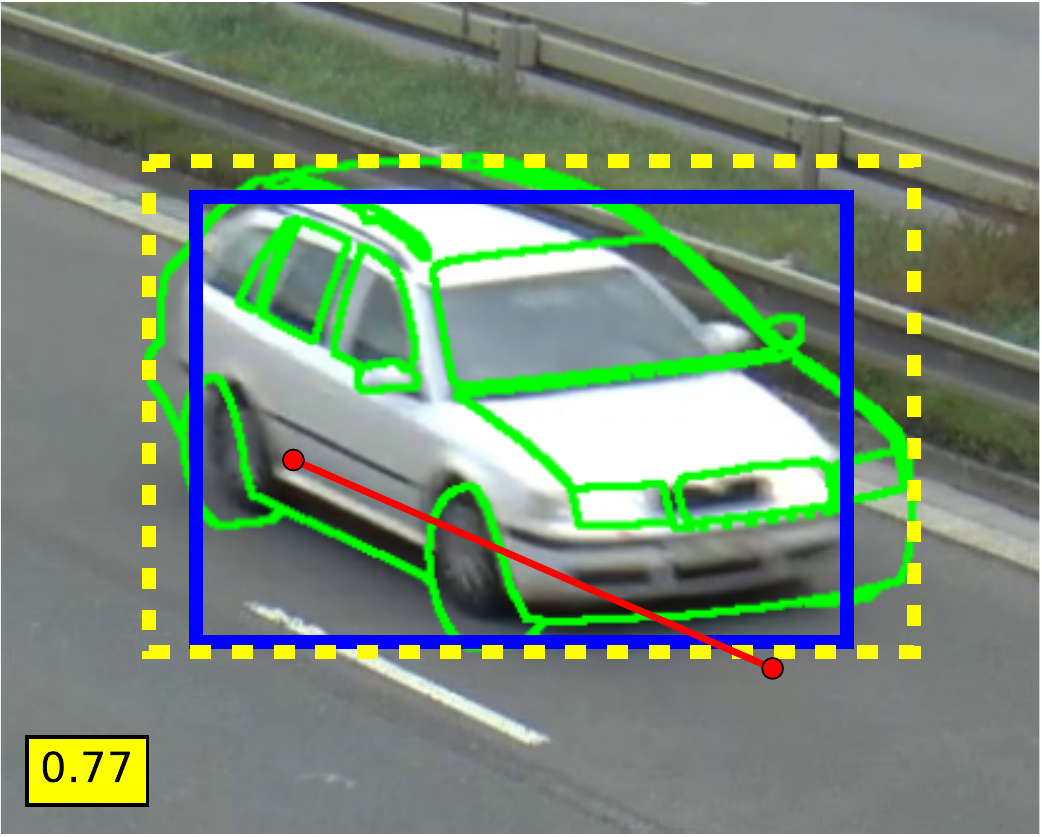}%
    \includegraphics[width=0.2\linewidth]{images_bb_iou_development_69_6284_025.pdf}%
    \includegraphics[width=0.2\linewidth]{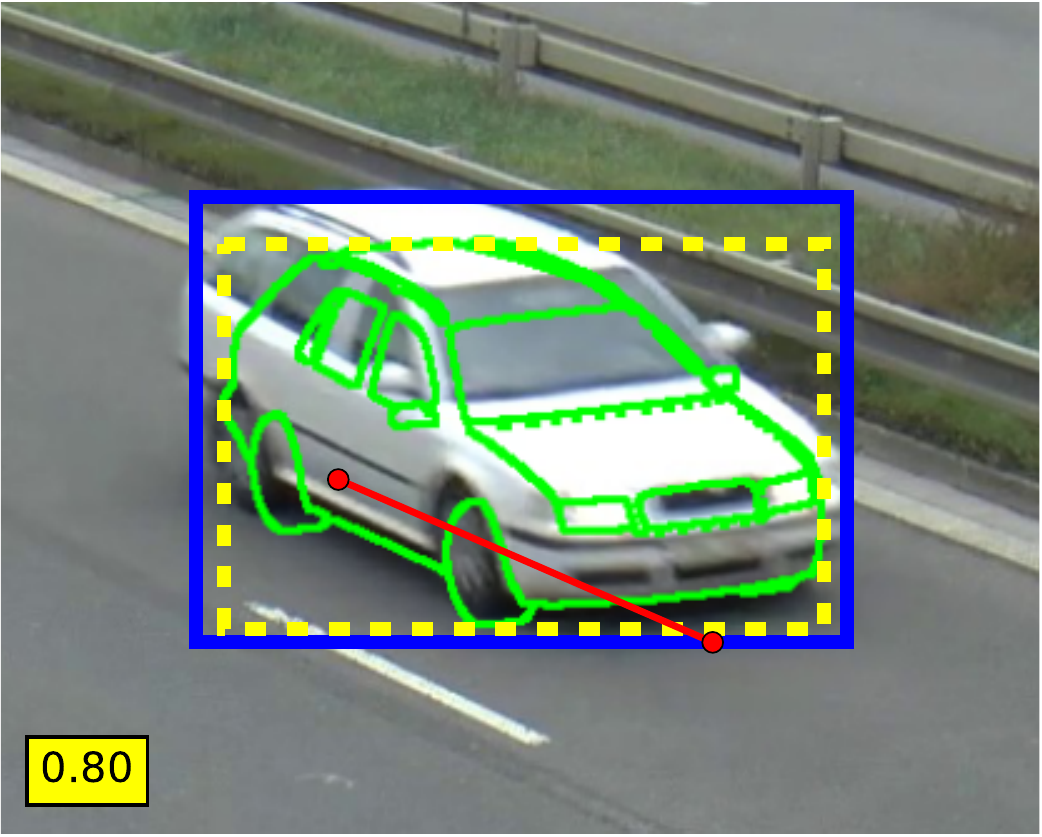}%
    \includegraphics[width=0.2\linewidth]{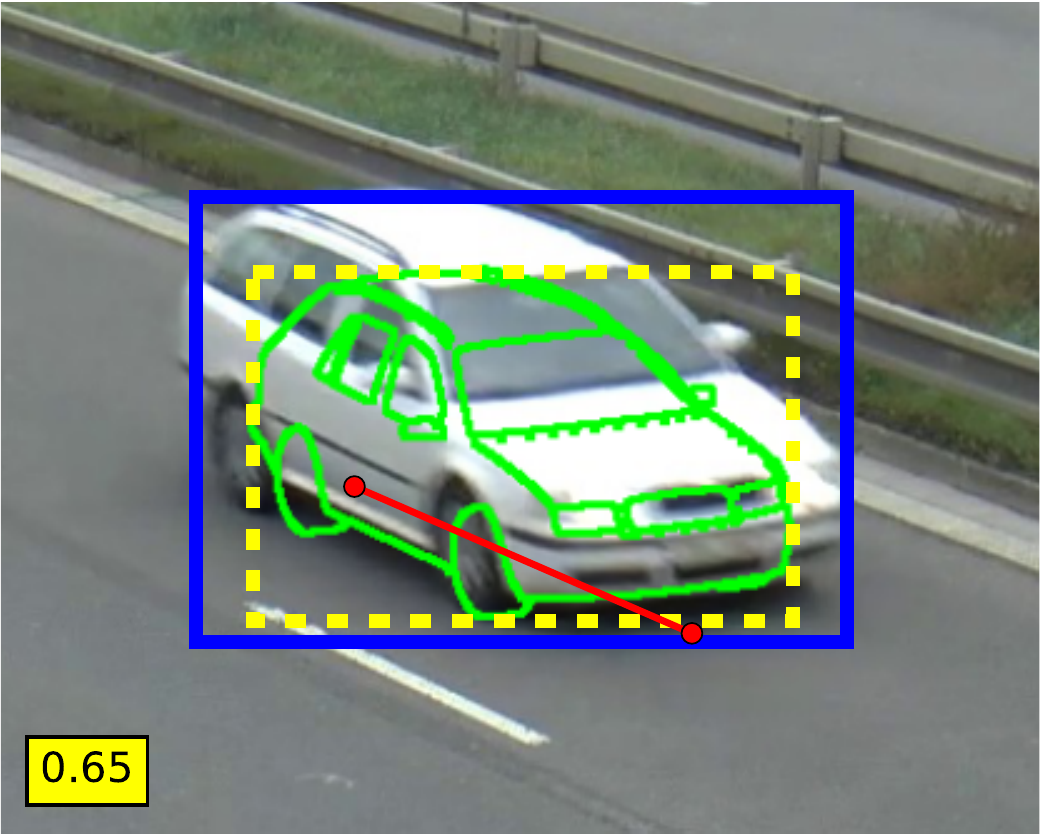}\\
    \includegraphics[width=0.2\linewidth]{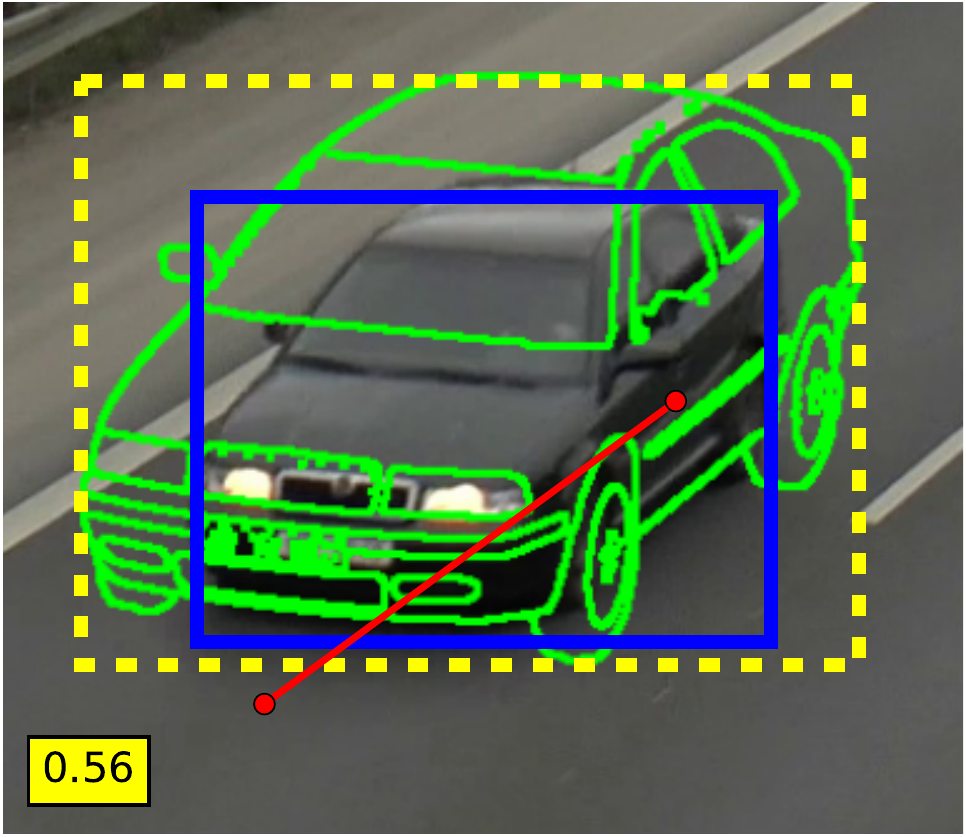}%
    \includegraphics[width=0.2\linewidth]{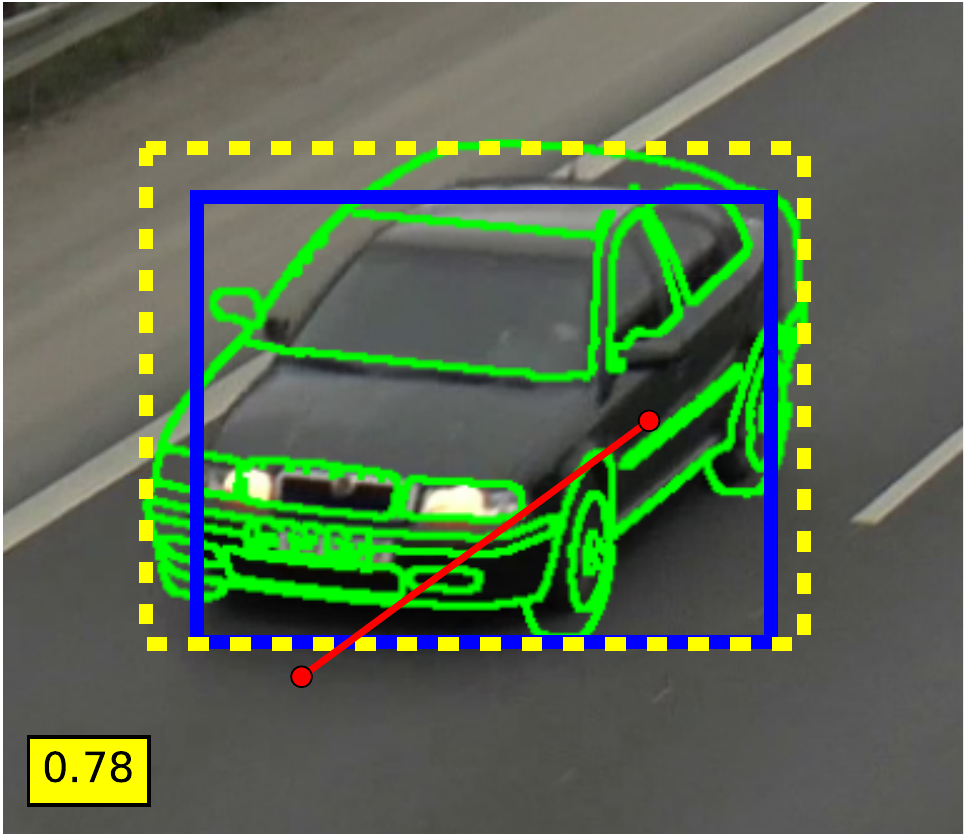}%
    \includegraphics[width=0.2\linewidth]{images_bb_iou_development_1640_185288_029.pdf}%
    \includegraphics[width=0.2\linewidth]{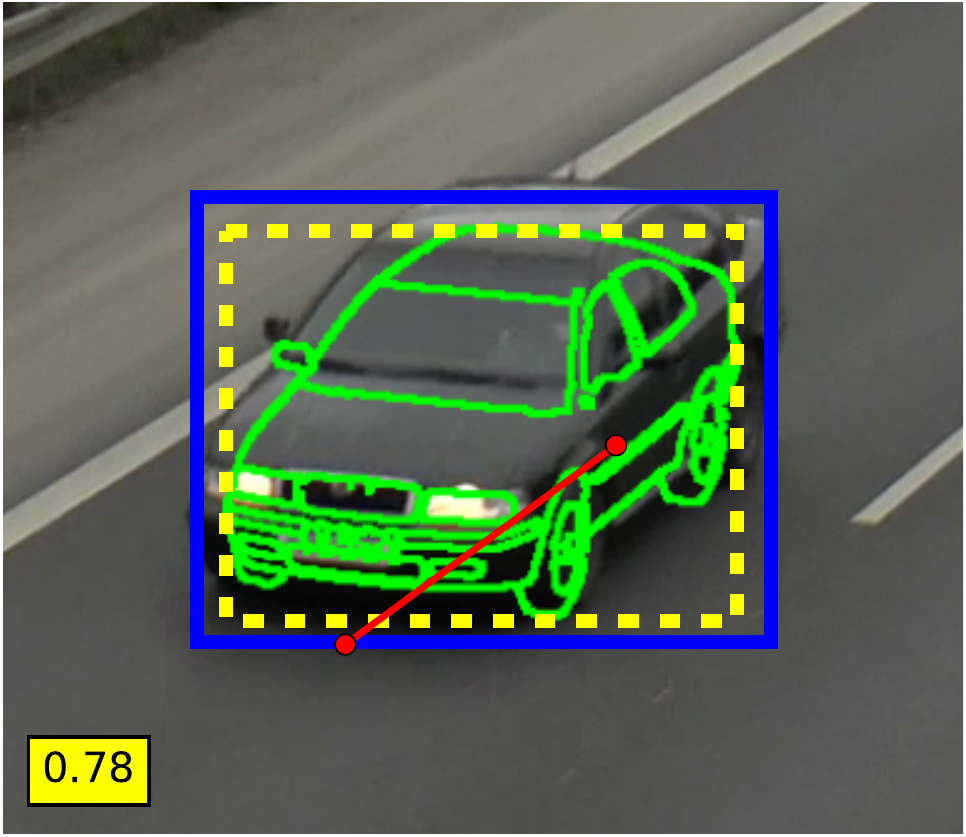}%
    \includegraphics[width=0.2\linewidth]{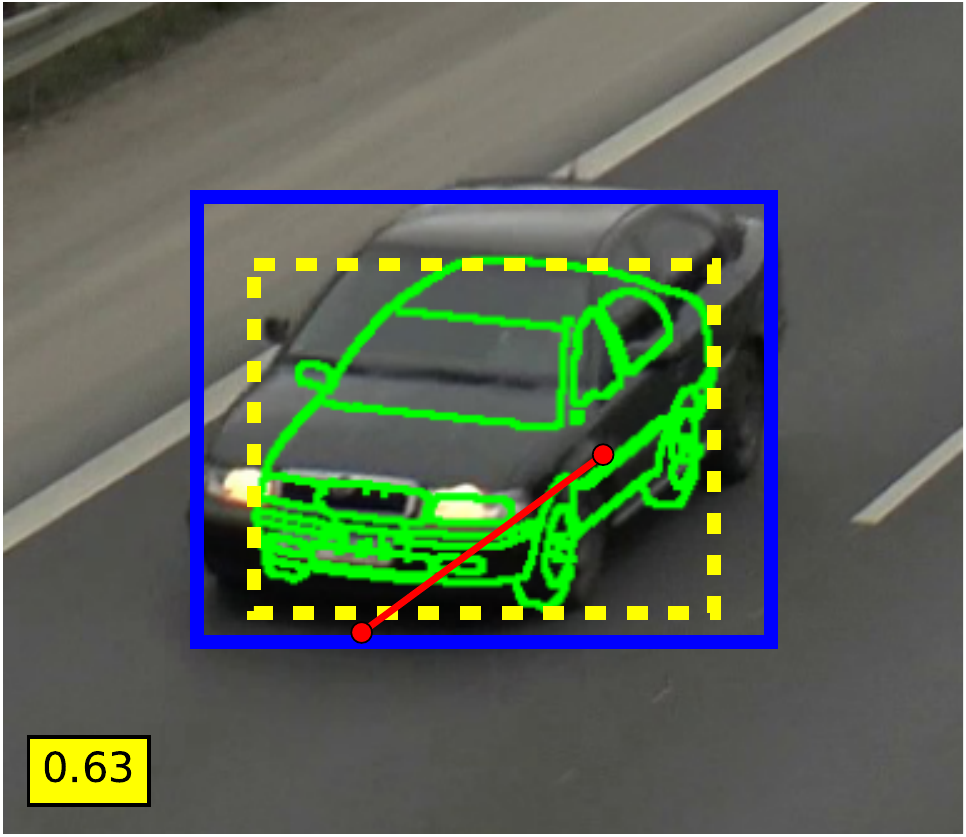}%
    \caption{Development of IoU (yellow boxes) metric for different scales (\textbf{left to right}), vehicle types and viewpoints (\textbf{top to bottom}). The left two images show larger rendered vehicles, the middle one shows the best match, and the right two images show smaller rendered vehicles. The rendered vehicle is shown only in a form of edges with the yellow rectangle bounding box of the rendered model and blue rectangle denoting the detected vehicle bounding box.}
    \label{fig:ScaleInference}
\end{figure*}

Once the viewpoint vector to the vehicle, the vehicle's class, and its position on the screen are determined, we render the appropriate 3D model given the parameters.  The only open variable is the scale of the vehicle to be rendered (i.e. the distance between the vehicle and the camera).
Examples of the two used 3D models are shown in Figure~\ref{fig:RenderVsReal}. Therefore, we render images of the vehicle in multiple different scales and match the bounding boxes of the rendered vehicles with the bounding box detected in the video by using the Intersection-over-Union (IoU) metric. Examples of such matches can be found in Figure~\ref{fig:ScaleInference}. 
The figure also shows two interesting points related to the vehicle in red: points on the base of the 3D models representing front $\mathbf{f}$ and rear $\mathbf{r}$ of the vehicle. Finally, for all vehicle instances $i$ and scales $j$, these points are projected on the road plane, yielding $\mathbf{F}_{ij}$ and $\mathbf{R}_{ij}$. They are used to compute the scale $\lambda_{ij}$ (Eq. \eqref{eq:BBScaleComputationOne}, where $l_{t_i}$ is the real world length of the type $t_i$). 
For all considered combinations of $i$ and $j$, the IoU matching metric $m_{ij}$ is computed.
\begin{eqnarray}
&\lambda_{ij} = \cfrac{l_{t_i}}{\|\mathbf{F}_{ij}-\mathbf{R}_{ij}\|} &\label{eq:BBScaleComputationOne}
\end{eqnarray}

To obtain the final camera's scale $\lambda^*$, all the scales $\lambda_{ij}$ are taken into account together with metrics $m_{ij}$.  We consider only cases with $m_{ij}$ larger then a predefined threshold (we used 0.85 in our experiments) to eliminate poor matches. Finally, we compute $\lambda^*$ according to Equation~\eqref{eq:BBScaleAll}. The probability $p\left(\lambda\,|\,(\lambda_{ij}, m_{ij}) \right)$ is computed by kernel density estimation with a discretized space:
\begin{eqnarray}
&\lambda^* = \arg \max_\lambda~p\left(\lambda\,|\,(\lambda_{ij}, m_{ij}) \right) &\label{eq:BBScaleAll}
\end{eqnarray}

In order to further improve the scale inference, we use several training videos from BrnoCompSpeed dataset \citep{BrnoCompSpeed}. We train the scale-correcting linear regression $\lambda^*_{reg} = \alpha\lambda^* + \beta$, using manually obtained scales as the ground truth. Even though this step is not necessary, it improves the scale acquisition further by correcting the imprecise geometry of the obtained 3D models. 

We also experimented with an alignment metric based on matching of edges on the rendered and detected vehicles (based on distance transform). However, the speed measurement did not improve further. The biggest problem with this method is that most of the edges on vehicles are blurry and therefore not detected at all. However, the vehicle detector \citep{Girshick2015} is able to detect the vehicles properly and in most cases accurately. Also, the proposed algorithm using just the bounding boxes is much more efficient in terms of storage (it is possible to store just the bounding boxes, not the images) and computation.

\subsection{Speed Measurement of Tracked Cars} \label{sec:SpeedMeasurement}
The speed measurement itself is done by following the methodology proposed by \cite{BrnoCompSpeed}.
Given a tracked car with reference points $\mathbf{p}_i$ and timestamps $t_i$ for each reference point, where $i=1 \dots N$, the speed $v$ is calculated from Equation \eqref{eq:SpeedComputation} by projecting the reference points $\mathbf{p}_i$ to the ground plane $\mathbf{P}_i$ (see Equation~\eqref{eq:3DCoords}).
\begin{eqnarray}
&v = \underset{i=1\dots N-\tau}{\mathrm{median}} \left( \frac{\displaystyle\lambda^*_{reg} \| \mathbf{P}_{i+\tau} - \mathbf{P}_i \|}{\displaystyle t_{i+\tau} - t_i} \right)&\label{eq:SpeedComputation}
\end{eqnarray}

The speed is computed as the median value of speeds between consecutive time positions. However, for stability of the measurement, it is better not to use the next frame, but the time position several video frames apart. This is controlled by the constant $\tau$, and for all our experiments, we use $\tau = 5$ (the time difference is usually $0.2\,\mathrm{s}$).

\section{Experiments and Results} \label{sec:Experiments}
\begin{figure}[t]
	\centering
	\includegraphics[width=\linewidth]{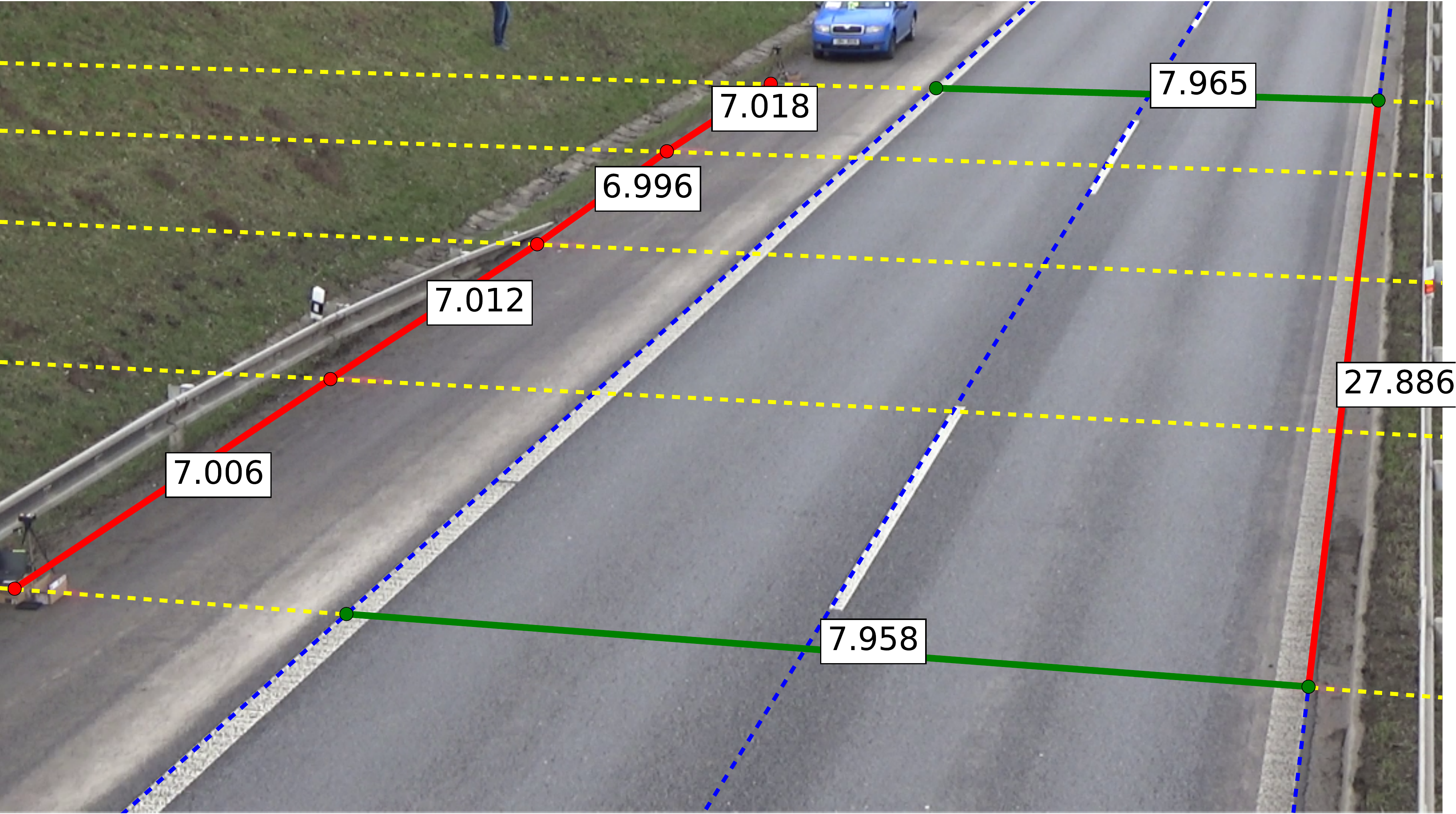}
	\caption{An example of manually measured distances between markers on the road plane. Other examples can be found in the original BrnoCompSpeed publication \citep{BrnoCompSpeed}. Blue lines denote the lane dividing lines, lines perpendicular to the vehicles direction are shown in yellow. Finally, measured distances between two points towards the first (second) vanishing point are shown by red (green) color.} 
	\label{fig:RoadMarkings}
\end{figure}

To evaluate our proposed methods for camera calibration and scene scale inference, we use the very recent BrnoCompSpeed dataset \citep{BrnoCompSpeed} which contains over 20\,k vehicles with precise ground truth speed from multiple locations. The dataset also contains markers on the road with known dimensions between them. For an example of such road markers, see Figure~\ref{fig:RoadMarkings}. The ground truth distances can be used for either calibration or evaluation of distance measurements on the road plane. It is also possible to evaluate the accuracy of vanishing point estimation by using the markings \citep{BrnoCompSpeed}. In the following text we will refer to various methods for camera calibration which are defined as: 
\begin{itemize}
	\item \textbf{\ITSCalib} -- Automatic camera calibration method as described by \cite{Dubska2015ITS}. Brief outline of the method is in Sections~\ref{sec:SOTASpeedMeasurement} and \ref{sec:VPEstimation}.
	\item \textbf{\EdgeLetsCalib} -- Camera calibration method proposed in this paper, Section~\ref{sec:VPEstimation}.
	\item \textbf{\ManualCalib} -- We use known distances (Figure~\ref{fig:RoadMarkings}) on the road for manual calibration of the camera. In agreement with the previous papers \citep{Cathey2005,Grammatikopoulos2005,He2007CalibMethod} we use intersection lanes dividing lines (blue dashed lines in Figure~\ref{fig:RoadMarkings}) for estimation of the first vanishing point $\mathbf{u}$. As there are usually more than just two lane dividing lines, we use least squares minimization to obtain the intersection of multiple lines. Formally, given lines $\mathbf{l}_i$ with normalized normal vectors, we compute the vanishing point $\mathbf{u}$ by solving $\mathbf{A} \mathbf{u} = -\mathbf{b}$ in a least squares manner, where rows of $\mathbf{A}$ contain transposed normal vectors of the lines, and rows of $\mathbf{b}$ contain constant terms of the lines. 
	
	The second vanishing point $\mathbf{v}$ can be obtained in the same manner (as the intersection of yellow dashed lines in Figure~\ref{fig:RoadMarkings}, since they are perpendicular to the vehicle flow on the road). However, we found out that it is more accurate and robust to use the intersection only as a first guess, and then use measured distances on the road to optimize the vanishing point position using Equation \eqref{eq:ManualCalib}.
	\begin{equation}
		\mathbf{v^*} = \arg \min_\mathbf{v} \left( \sum_{(\mathbf{p}_1, \mathbf{p}_2, d) \in \mathcal{D}_2} \left| \lambda \| \mathbf{P}_1 - \mathbf{P}_2 \| - d\right| \right) \label{eq:ManualCalib}, 
	\end{equation}
	where set $\mathcal{D}_2$ contains image endpoints and distances measured on the road towards the second vanishing point (green line segments in Figure~\ref{fig:RoadMarkings}) and scale $\lambda$ is computed for the given vanishing points $\mathbf{u}, \mathbf{v}$ by Equation~\eqref{eq:ManualScale}. It should be noted that the computation of 3D coordinates $\mathbf{P}_i$ of image point $\mathbf{p}_i$ depends on the vanishing points (see Equation~\eqref{eq:3DCoords} for details). The optimization itself is done by grid search (we loop over discretized feasible positions of $\mathbf{v}$ corresponding to reasonable focal lengths and evaluate the optimization objective \eqref{eq:ManualCalib}).

	The usage of standard manual methods based on calibration patterns (e.g checkerboards) proposed by \cite{Zhang2000} is impractical, as it would require a large checkerboard (more than $10\,\mathrm{m}^2$) placed on the road.
\end{itemize}

We also define method names for different approaches for scale inference:
\begin{itemize}
	\item \textbf{\BMVCScale} -- Scale inference method proposed by \cite{Dubska2014}. Brief outline of the method is in Section~\ref{sec:SOTASpeedMeasurement}.
	\item \textbf{\BBScaleReg} -- Our method for scale calibration using bounding box matching (Section~\ref{sec:ScaleInference}) with scale correction regression.
	\item \textbf{\ManualScale} -- Scale computed from manually measured distances between markers towards the first vanishing point on the road. The scale is computed as the mean value of Equation~\eqref{eq:ManualScale} from a set of endpoints and distances $(\mathbf{p}_{i,1}, \mathbf{p}_{i,2}, d_i)$ towards the first vanishing point (red line segments in Figure~\ref{fig:RoadMarkings}). 
	\begin{equation}
		\lambda = \mathbb{E}\left[ \frac{d_i}{\| \mathbf{P}_{i,1} - \mathbf{P}_{i,2} \|} \right] \label{eq:ManualScale}
	\end{equation}

	\item \textbf{\SpeedScale} -- Scale is computed from ground truth speed measurements and minimizes the speed measurement error for given camera calibration. It can be understood as the lower error bound for the given camera calibration method. The scale is computed as the mean value of Equation~\eqref{eq:SpeedScale} where, the set $\mathcal{M}$ contains pairs of ground truth speed $\gt{v}_i$ and measured speed $v_i$. It is assumed that scale $\lambda = 1$ was used for computation of speeds $v_i$.
	\begin{equation}
	\lambda =  \mathbb{E} \left[ \frac{\gt{v}_i}{v_i} \right] \label{eq:SpeedScale}
	\end{equation}
\end{itemize}

If not stated otherwise, the evaluation was done on BrnoCompSpeed -- Split C (contains more than 10\,k of vehicle tracks for evaluation), because our method requires parameter tuning for the scale correction regression and split C provides a sufficient amount of data for training and testing. For each metric, we report mean, median, and 99 percentile error for both absolute units ($err = |\gt{r} - r|$) and relative units ($err = |\gt{r} - r|/\gt{r} \cdot 100\%$), where $\gt{r}$ denotes the ground truth measurement, and $r$ represents the measured value.

\subsection{Evaluation of Vanishing Point Estimation -- Camera Calibration Error} \label{sec:CameraCalibEvaluation}
To evaluate the camera calibration itself (the obtained vanishing points), we follow the evaluation metric proposed with the BrnoCompSpeed dataset \citep{BrnoCompSpeed}. The evaluation measures the difference between ratios of distances between markings towards the first vanishing point (red lines in Figure~\ref{fig:RoadMarkings}) and the distances between markers towards the second vanishing point (green lines in Figure~\ref{fig:RoadMarkings}). As the ratio does not depend on scale, this metric considers only the camera calibration in the form of two detected vanishing points. 

Since we do not require any parameter tuning for the camera calibration method, we report the results on all videos in the BrnoCompSpeed dataset (including the extra session0).
The results (reported in Table~\ref{tab:CalibrationError}) show that our automatic calibration method \EdgeLetsCalib\ outperforms calibration method \ITSCalib\ almost twice on mean error. 
It should be noted that the same distances that were used to obtain the manual calibration were evaluated by the calibration error metric based on distance ratios; this gives the manual calibration an unfair advantage in the comparison.

The significant improvement of our method is caused by more precise acquisition of $\mathbf{v}$; position of $\mathbf{u}$ stays the same for our method as for the \ITSCalib\ calibration method. 
There are two reasons why vanishing points play an important role. The first one is that the vanishing points are directly used for estimating the focal length; the second one is that they are used for computation of the viewpoint on the vehicle for scale estimation. Therefore, if the viewpoint is computed imprecisely, the alignment of the rendered 3D model is also imprecise.

\begin{table}[t!]
	\centering
	\caption{Errors of distance measurement ratios (see Section~\ref{sec:CameraCalibEvaluation} for details). The first row for each calibration method contains \abso{absolute errors}; the \proc{relative errors in percents} are in the second row.}  \label{tab:CalibrationError}
	\vspace{2mm}
	\setlength{\tabwidth}{0.5\linewidth}
	\small
	\begin{tabular}{l r r r }
		\toprule
		\textbf{system} & \textbf{mean} & \textbf{median} & \textbf{99\,\%} \\
		\midrule
		
		\multirow{2}{*}{\EdgeLetsCalib\ (ours)} & \abso{0.09} & \abso{0.04} & \abso{0.49}\\
		& \proc{6.45} & \proc{3.38} & \proc{39.08}\\[-\jot]			    \multicolumn{4}{@{}c@{}}{\makebox[\tabwidth]{\dashrule[lightgray]}} \\[-\jot]
		
		\multirow{2}{*}{\ITSCalib} & \abso{0.18} & \abso{0.05} & \abso{1.36}\\
		& \proc{11.74} & \proc{5.25} & \proc{61.03}\\
		\midrule
		
		\multirow{2}{*}{\ManualCalib} & \abso{0.02} & \abso{0.01} & \abso{0.15}\\
		& \proc{1.80} & \proc{1.26} & \proc{10.98}\\
		
		\bottomrule
	\end{tabular}
\end{table}

\subsection{Evaluation of Distance Measurement in the Road Plane} \label{sec:DistanceMeasurementEval}

\begin{table}[t!]
	\centering
	\caption{Distance measurement errors on the road plane for different calibrations. Only distances towards the first vanishing point (red in Figure~\ref{fig:RoadMarkings}) were used for this evaluation. The first row for each calibration method contains \abso{absolute errors in meters}; the \proc{relative errors in percents} are in the second row.}  \label{tab:DistanceErrorVP1}
	\vspace{2mm}
	\setlength{\tabwidth}{0.65\linewidth}
	\small
	\begin{tabular}{l r r r }
		\toprule
		\textbf{system} & \textbf{mean} & \textbf{median} & \textbf{99\,\%} \\
		\midrule
		
		\multirow{2}{*}{\EdgeLetsCalib\ + \BBScaleReg\ (ours)} & \abso{0.26} & \abso{0.17} & \abso{1.08}\\
		& \proc{2.33} & \proc{2.06} & \proc{5.49}\\[-\jot]			    \multicolumn{4}{@{}c@{}}{\makebox[\tabwidth]{\dashrule[lightgray]}} \\[-\jot]
		
		\multirow{2}{*}{\ITSCalib\ + \BMVCScale } & \abso{1.23} & \abso{0.81} & \abso{5.40}\\
		& \proc{9.62} & \proc{10.65} & \proc{21.07}\\
		
		\midrule
		
		\multirow{2}{*}{\EdgeLetsCalib\ + \ManualScale\ (ours) } & \abso{0.10} & \abso{0.06} & \abso{0.57}\\
		& \proc{0.98} & \proc{0.62} & \proc{4.46}\\[-\jot]			    \multicolumn{4}{@{}c@{}}{\makebox[\tabwidth]{\dashrule[lightgray]}} \\[-\jot]
		
		\multirow{2}{*}{\ITSCalib\ + \ManualScale} & \abso{0.25} & \abso{0.14} & \abso{1.54}\\
		& \proc{2.11} & \proc{1.66} & \proc{8.07}\\[-\jot]			    \multicolumn{4}{@{}c@{}}{\makebox[\tabwidth]{\dashrule[lightgray]}} \\[-\jot]
		
		\multirow{2}{*}{\ManualCalib\ + \ManualScale} &\abso{0.10} & \abso{0.08} & \abso{0.32}\\
		& \proc{1.08} & \proc{0.65} & \proc{3.59}\\
		
		\bottomrule
	\end{tabular}
\end{table}
\begin{table}[t!]
	\centering
	\caption{Distance measurement errors on the road plane for different calibrations. Each segment of the table represents a different level of supervision in the calibration. The first row for each calibration method contains \abso{absolute errors in meters} and the \proc{relative errors in percents} are in the second row.}  \label{tab:DistanceErrorAll}
	\vspace{2mm}
	\setlength{\tabwidth}{0.65\linewidth}
	\small
	\begin{tabular}{l r r r }
		\toprule
		\textbf{system} & \textbf{mean} & \textbf{median} & \textbf{99\,\%} \\
		\midrule
		
		\multirow{2}{*}{\EdgeLetsCalib\ + \BBScaleReg\ (ours)} & \abso{0.34} & \abso{0.18} & \abso{2.29}\\
		& \proc{3.47} & \proc{2.28} & \proc{30.49}\\[-\jot]			    \multicolumn{4}{@{}c@{}}{\makebox[\tabwidth]{\dashrule[lightgray]}} \\[-\jot]
		
		\multirow{2}{*}{\ITSCalib\ + \BMVCScale } & \abso{1.17} & \abso{0.72} & \abso{5.82}\\
		& \proc{9.79} & \proc{9.00} & \proc{55.89}\\
		
		\midrule
		
		\multirow{2}{*}{\EdgeLetsCalib\ + \ManualScale\ (ours) } & \abso{0.24} & \abso{0.10} & \abso{2.60}\\
		& \proc{2.66} & \proc{1.00} & \proc{34.75}\\[-\jot]			    \multicolumn{4}{@{}c@{}}{\makebox[\tabwidth]{\dashrule[lightgray]}} \\[-\jot]
		
		\multirow{2}{*}{\ITSCalib\ + \ManualScale} & \abso{0.57} & \abso{0.20} & \abso{5.43}\\
		& \proc{5.84} & \proc{2.07} & \proc{52.19}\\[-\jot]			    \multicolumn{4}{@{}c@{}}{\makebox[\tabwidth]{\dashrule[lightgray]}} \\[-\jot]
		
		\multirow{2}{*}{\ManualCalib\ + \ManualScale} &\abso{0.07} & \abso{0.04} & \abso{0.30}\\
		& \proc{0.84} & \proc{0.50} & \proc{3.47}\\
		
		\bottomrule
	\end{tabular}
\end{table}
The next step is to evaluate the camera calibration together with the obtained scale. 
We use manual annotations of distances on the road plane which are directed towards the first or the second vanishing point, respectively (red and green in Figure~\ref{fig:RoadMarkings}). 

First, we evaluated the distance measurement only towards the first vanishing point as it is the direction in which the vehicles are going and it is more important for speed measurement. 
The results are shown in Table~\ref{tab:DistanceErrorVP1} for different combinations of calibrations and scale estimations. First, our fully automatic method for camera calibration (\EdgeLetsCalib) and scale inference (\BBScaleReg) significantly outperforms the previous automatic method \ITSCalib\ + \BMVCScale.
Second, when we use our automatically computed calibration and scale obtained with manual annotations, we achieve almost the same results as \ManualCalib\ + \ManualScale, which required much more manual effort than our automatic system.

When we evaluated the same metric with all the distances, the results are similar (see Table~\ref{tab:DistanceErrorAll}). Again, our method significantly outperforms the previous automatic method. Considering the calibrations with manually obtained scale, our system has a slightly higher error then the manual calibration. However, this is caused by the fact that the manual calibration is optimized directly to the evaluation metric by Equation~\eqref{eq:ManualCalib} and thus gets an unfair and unrealistic advantage.

To summarize the distance measurement results: our method significantly outperforms previous automatic state-of-the-art for speed measurement -- the mean error for distance measurement in the direction of vehicles' flow (which is important for speed measurement) was reduced by $79\,\%$ (1.23\,m to 0.26\,m).

\subsection{Evaluation of Speed Measurement} \label{sec:SpeedMeasurementEval}
The most important part of the evaluation is the speed measurement itself. We used the same vehicle detection and tracking system (see Section~\ref{fig:DetectionTracking}) in all experiments so that the results for different calibrations and scales are directly comparable.

\begin{table}[t!]
	\centering
	\caption{Evaluation of speed measurement errors; all systems differ only in the calibration and scale inference, with the same tracking of vehicles. Each segment represents one level of supervision in the calibration (automatic, known ground truth distances on road, known ground truth speeds). The first row for each calibration method contains \abso{absolute errors in km/h}; the \proc{relative errors in percents} are in the second row.}  \label{tab:SpeedMeasurementEval}
	\vspace{2mm}
	\setlength{\tabwidth}{0.65\linewidth}
	\small
	\begin{tabular}{l r r r }
		\toprule
		\textbf{system} & \textbf{mean} & \textbf{median} & \textbf{99\,\%} \\
		\midrule
		
		\multirow{2}{*}{\EdgeLetsCalib\ + \BBScaleReg\ (ours)} & \abso{1.10} & \abso{0.97} & \abso{3.05}\\
		& \proc{1.39} & \proc{1.22} & \proc{4.13}\\[-\jot]			    
		\multicolumn{4}{@{}c@{}}{\makebox[\tabwidth]{\dashrule[lightgray]}} \\[-\jot]
		
		\multirow{2}{*}{\ITSCalib\ + \BMVCScale } & \abso{7.98} & \abso{8.18} & \abso{18.58}\\
		& \proc{10.15} & \proc{11.45} & \proc{19.22}\\
		
		\midrule
		
		\multirow{2}{*}{\EdgeLetsCalib\ + \ManualScale\ (ours) } & \abso{1.04} & \abso{0.83} & \abso{3.48}\\
		& \proc{1.31} & \proc{1.04} & \proc{4.61}\\[-\jot]			    \multicolumn{4}{@{}c@{}}{\makebox[\tabwidth]{\dashrule[lightgray]}} \\[-\jot]
		
		\multirow{2}{*}{\ITSCalib\ + \ManualScale} & \abso{1.44} & \abso{1.17} & \abso{5.43}\\
		& \proc{1.76} & \proc{1.50} & \proc{6.16}\\[-\jot]			    \multicolumn{4}{@{}c@{}}{\makebox[\tabwidth]{\dashrule[lightgray]}} \\[-\jot]

		\multirow{2}{*}{\ManualCalib\ + \ManualScale} &\abso{1.35} & \abso{0.95} & \abso{4.84}\\
		& \proc{1.64} & \proc{1.18} & \proc{5.40}\\
		\midrule
		
		\multirow{2}{*}{\EdgeLetsCalib\ + \SpeedScale\ (ours)} & \abso{0.52} & \abso{0.35} & \abso{2.57}\\
		& \proc{0.66} & \proc{0.44} & \proc{3.71}\\[-\jot]			    \multicolumn{4}{@{}c@{}}{\makebox[\tabwidth]{\dashrule[lightgray]}} \\[-\jot]
		
		\multirow{2}{*}{\ITSCalib\ + \SpeedScale} & \abso{0.80} & \abso{0.57} & \abso{3.70}\\
		& \proc{0.99} & \proc{0.72} & \proc{4.68}\\[-\jot]			    \multicolumn{4}{@{}c@{}}{\makebox[\tabwidth]{\dashrule[lightgray]}} \\[-\jot]

		\multirow{2}{*}{\ManualCalib\ + \SpeedScale} &\abso{0.56} & \abso{0.38} & \abso{2.73}\\
		& \proc{0.71} & \proc{0.48} & \proc{3.63}\\
		
		\bottomrule
	\end{tabular}
\end{table}

\begin{figure}[t!]
	\centering
	\makebox[\textwidth][c]{%
		\includegraphics[width=0.7\linewidth]{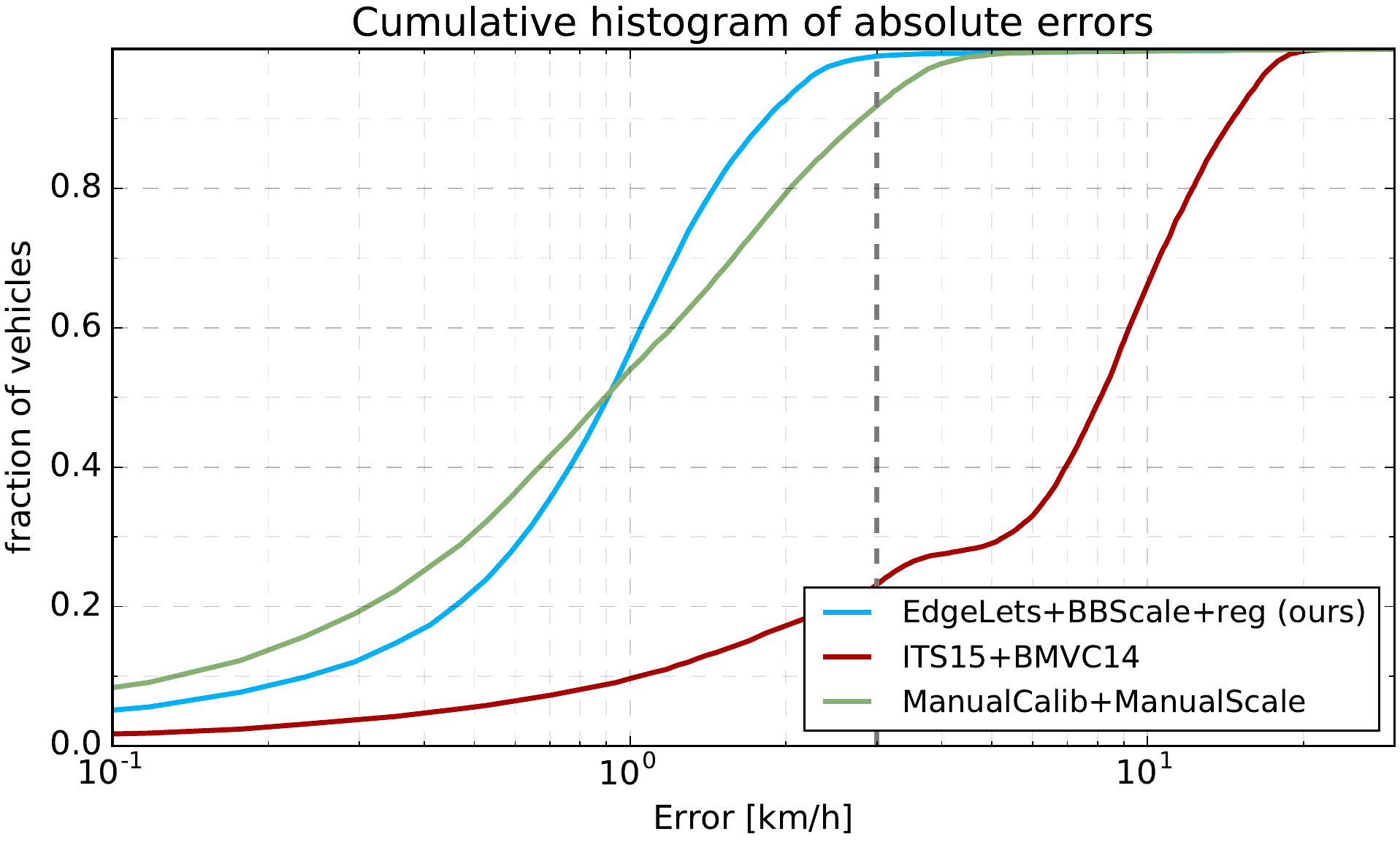}%
		\includegraphics[width=0.7\linewidth]{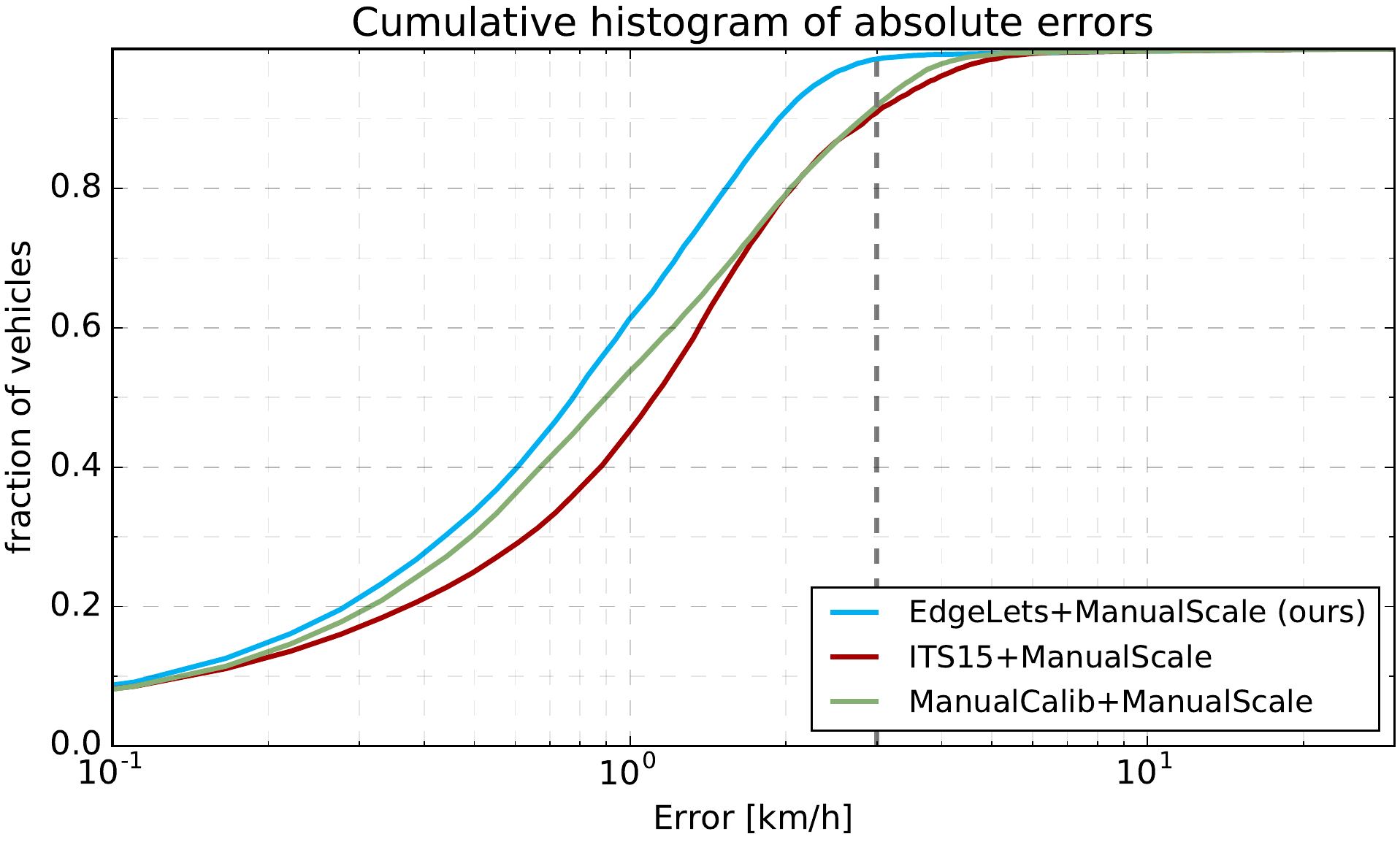}%
	}
	\makebox[\textwidth][c]{%
			\includegraphics[width=0.7\linewidth]{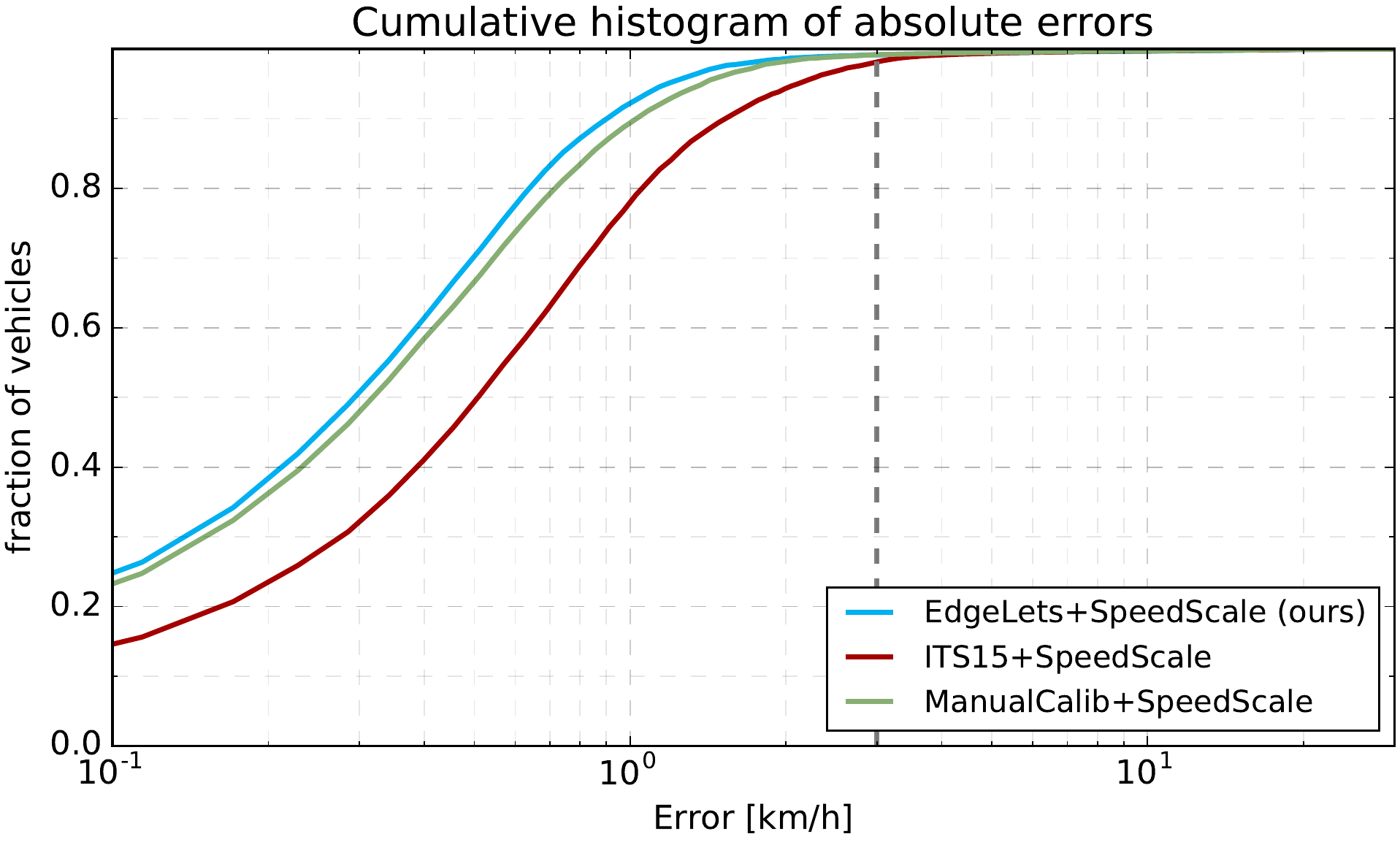}%
		\includegraphics[width=0.7\linewidth]{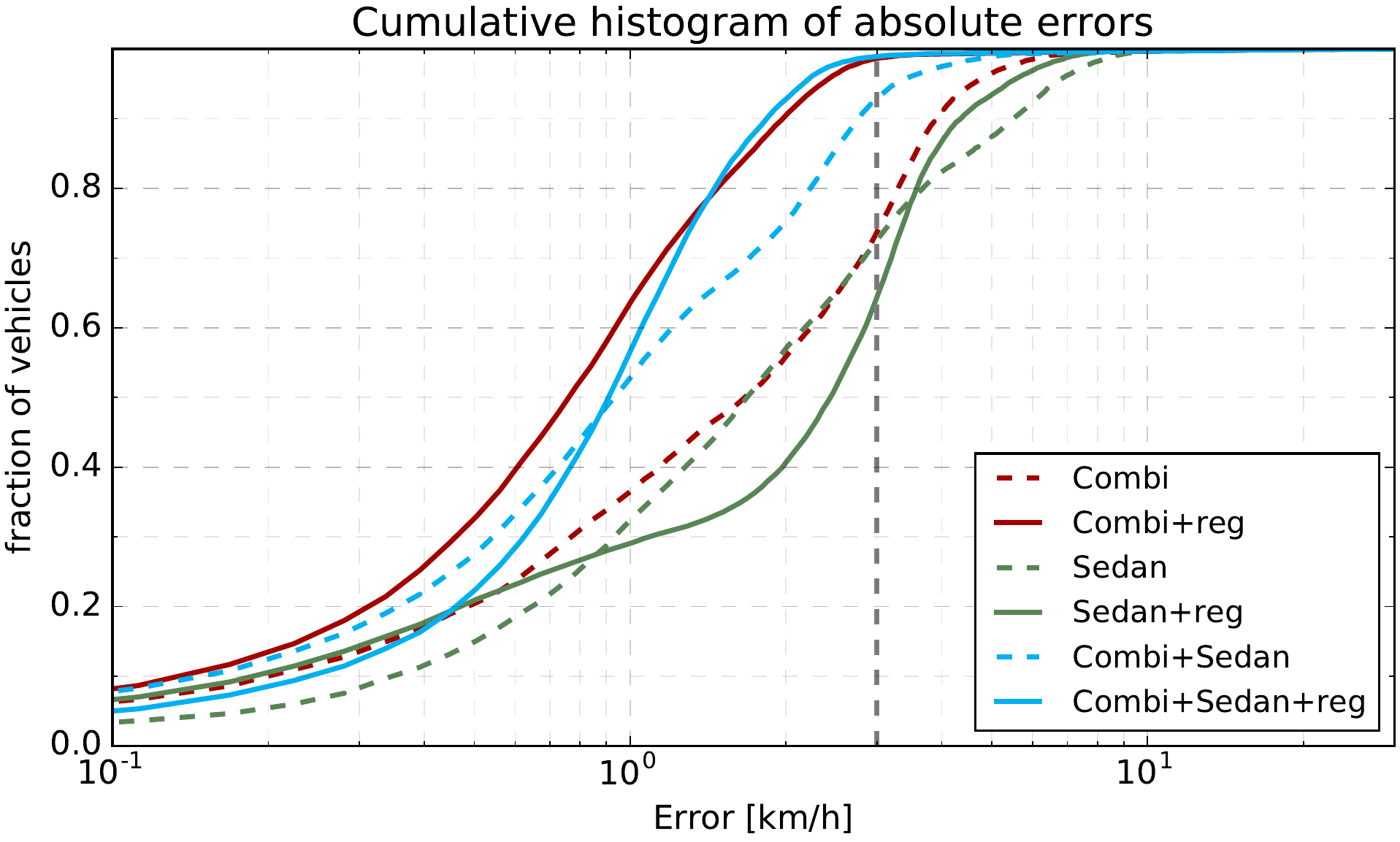}%
	}
	\caption{Evaluation of speed measurement -- cumulative histograms of errors. The gray dashed vertical lines represent 3\,km/h error. \textbf{top left:} comparison of automatic methods and a manual method for camera calibration, \textbf{top right:} calibrations obtained with known ground truth distances on the road plane, \textbf{bottom left:} calibrations with scale obtained by minimizing the speed measurement error, thus forming a lower bound error for speed measurement with given camera calibration and tracking algorithm, \textbf{bottom right:} analysis of influence of different aspects of used 3D car models evaluated on speed measurement, see Section~\ref{sec:SensitivityAnalysis}. 
		The cumulative histogram is suitable for directly obtaining the ``success rate'' for a given error tolerance.} 
	\label{fig:SpeedMeasurementEval}
\end{figure}

We show both quantitative results in the form of Table \ref{tab:SpeedMeasurementEval} and plots with cumulative error histograms in Figure~\ref{fig:SpeedMeasurementEval}. The table and the figures are divided into several parts where we compare similar levels of supervision. 

The first level of supervision is fully automatic; in the second level, known ground truth dimensions on the road plane are used. In the third and final level of supervision, we use known ground truth speeds to form the lower error bound for different calibration methods. 

Regarding the first level of supervision, our system \EdgeLetsCalib\ + \BBScaleReg\ significantly outperforms the previous automatic method \ITSCalib\ + \BMVCScale\ and we reduce the mean speed measurement error by $86\,\%$ (7.98\,km/h to 1.10\,km/h) . Another important fact is that our fully automatic method for camera calibration and scale inference also outperforms manual calibration and scale inference (1.35\,km/h mean error) while the error is reduced by $19\,\%$ (1.35\,km/h to 1.10\,km/h). 
This improvement is important because in previous approaches, the automation always compromised  accuracy, forcing the system developer to trade off between them.  Our work shows that fully automatic calibration methods may produce better results than manual calibration.

When it comes to the second and third level of supervision, the results follow the same trend with our calibration outperforming all of them (manual and automatic). The fact that manual calibration is better on the calibration metric (Section~\ref{sec:CameraCalibEvaluation}) and distance measurement (Section~\ref{sec:DistanceMeasurementEval}), while our method outperforms the manual calibration at the speed measurement task, is caused by the fact that manual calibration uses the same data which are then used for the evaluation of the calibration metric and distance measurement. 
The achieved accuracy is very close to meeting the standards for speed measurements accuracy required for enforcement (typically $3\,\%$ in many European countries).  The accuracy is definitely comparable to measurements achievable by radars \citep{BrnoCompSpeed}, while being considerably cheaper, more flexible, and passive.

\subsection{Sensitivity to Selection of the 3D Model} \label{sec:SensitivityAnalysis}

We also evaluated how using different 3D models of vehicles influences the speed measurement results. The results are shown in Table~\ref{tab:BBScaleAnalysis} and Figure~\ref{fig:SpeedMeasurementEval} (bottom right).
We tested several combinations of used vehicles: use of only one of the models (Combi, Sedan) or both of them together (Combi + Sedan), forming the first segment of the table. It shows that using both models significantly improves the results, as the errors in geometry of the 3D models cancel out. 
We consider that using only a few (as few as two) fine-grained models is beneficial because it is not necessary to obtain more 3D models and training data for fine-grained recognition.  
The experiments show that having two models is sufficient for obtaining usable results; using more than two models in practice would follow the same principles and could increase the robustness further.

\begin{table}[t]
	\centering
	\caption{Analysis of influence of different aspects of used 3D car models. It shows that it is best to use both models. The second segment of the table also shows that it is useful to use scale correction regression as described in Section~\ref{sec:ScaleInference}. The first row for each 3D model combination method contains \abso{absolute errors in km/h}; the \proc{relative errors in percents} are in the second row.}  \label{tab:BBScaleAnalysis}
	\vspace{2mm}
	\setlength{\tabwidth}{0.6\linewidth}
	\small
	\begin{tabular}{l r r r }
		\toprule
		\textbf{system} & \textbf{mean} & \textbf{median} & \textbf{99\,\%} \\
		\midrule
		
		\multirow{2}{*}{Sedan} & \abso{2.39} & \abso{1.74} & \abso{8.67}\\
		& \proc{2.82} & \proc{2.14} & \proc{7.74}\\[-\jot]			    \multicolumn{4}{@{}c@{}}{\makebox[\tabwidth]{\dashrule[lightgray]}} \\[-\jot]
		
		\multirow{2}{*}{Combi} & \abso{2.03} & \abso{1.72} & \abso{6.51}\\
		& \proc{2.48} & \proc{2.14} & \proc{5.94}\\[-\jot]			    \multicolumn{4}{@{}c@{}}{\makebox[\tabwidth]{\dashrule[lightgray]}} \\[-\jot]
		
		\multirow{2}{*}{Combi + Sedan} & \abso{1.38} & \abso{0.99} & \abso{5.18}\\
		& \proc{1.70} & \proc{1.23} & \proc{4.94}\\
		
		\midrule
		
		\multirow{2}{*}{Sedan + reg} & \abso{2.43} & \abso{2.49} & \abso{7.26}\\
		& \proc{2.97} & \proc{3.17} & \proc{6.56}\\[-\jot]			    \multicolumn{4}{@{}c@{}}{\makebox[\tabwidth]{\dashrule[lightgray]}} \\[-\jot]
		
		\multirow{2}{*}{Combi + reg} & \abso{1.03} & \abso{0.82} & \abso{3.29}\\
		& \proc{1.33} & \proc{1.04} & \proc{4.49}\\[-\jot]			    \multicolumn{4}{@{}c@{}}{\makebox[\tabwidth]{\dashrule[lightgray]}} \\[-\jot]
		
		\multirow{2}{*}{Combi + Sedan + reg} & \abso{1.10} & \abso{0.97} & \abso{3.05}\\
		& \proc{1.39} & \proc{1.22} & \proc{4.13}\\	
		
		\bottomrule
	\end{tabular}
\end{table}

The second segment of the table shows the performance of the system with scale correction regression to overcome the inaccuracies of the 3D models. The results show that for model Combi, the error significantly decreases. However, for the Sedan model, the results stay more or less the same. This paradox is caused by the smaller number of training data for Sedan version as for some training videos, no Sedan vehicle was detected.
The results also show that if we use both models, the performance drop is not that significant (1.10\,km/h to 1.38\,km/h) and therefore, it is possible to use the scale inference without the scale correction regression.

\subsection{Vehicle Detection and Tracking Evaluation} \label{sec:TrackingEval}

\begin{table}[t]
	\centering
	\caption{Evaluation of differences between vehicle detection and tracking proposed by  \cite{Dubska2014} and our detection and tracking method. FPPM denotes the number of False Positives Per Minute, recall was computed as mean recall across all videos and speed error denotes mean speed measurement error.}  \label{tab:CountingResults}
	\vspace{2mm}
	\setlength{\tabwidth}{0.90\linewidth}
	\small
	\begin{tabular}{l r r r}
		\toprule
		\textbf{method} & \textbf{FPPM} & \textbf{recall} & \textbf{speed error [km/h]}\\
		\midrule
		\cite{Dubska2014} & 9.77 & 0.885 & 1.46\\
		ours & 1.91 & 0.863 & 1.21\\	
		\bottomrule
	\end{tabular}
\end{table}

Since we use a different vehicle detection and tracking method from \cite{Dubska2014}, we also evaluate this part of the solution. We compare the methods on all videos of BrnoCompSpeed (including extra session0) with exactly the same calibration (\ManualCalib\ + \ManualScale) to isolate the influence of vehicle detection and tracking. 

We report the number of False Positives Per Minute and mean recall in vehicle counting. The results can be found in Table~\ref{tab:CountingResults}, and as the table shows, our method considerably reduces the number of false positives with essentially the same recall.

A tracked vehicle is matched to the ground truth if it passes through the correct lane and the time difference of pass through the measurement line (yellow line in Figure~\ref{fig:RoadMarkings} which is closest to the camera) compared to the ground truth is less than $0.2\,\mathrm{s}$. This threshold is used by \cite{BrnoCompSpeed} to correctly match the vehicles, as a higher threshold could lead to mismatches between the detected track and ground truth.

As we use the same calibration, we can also compare directly the speed measurement error which is influenced (with the same calibration) only by the tracking. As the table shows, our tracking method yields slightly reduced speed measurement error for the same scale and camera calibration.

% For geometrically correct speed measurement, it is necessary to measure the speed of a point on the road plane. We have also evaluated how the method is sensitive to whether we use the point on the road plane or a different one for the speed measurement. We have tried to evaluate the system when we were measuring speed of a point in the middle of the bottom side of the 2D bounding box. The difference between geometrically correct speed measurement of points on the base of the 3D bounding box and these results was negligible.

For the tracking and speed measurement, we use the point at the front of the vehicle on the road plane (using the 3D bounding box), which is geometrically correct, as the point is on the road plane. We evaluated how the choice of the tracking point influences the measurement error, comparing to a naive solution which takes the center of the bottom edge of the 2D bounding box for the tracking, and we found out that the difference to the correct solution was negligible.

\subsection{Camera Calibration on Real Surveillance Cameras}
\label{sec:RealCameras}

\begin{figure*}[t]
	\centering
	\makebox[\textwidth][c]{%
		\includegraphics[width=0.45\linewidth]{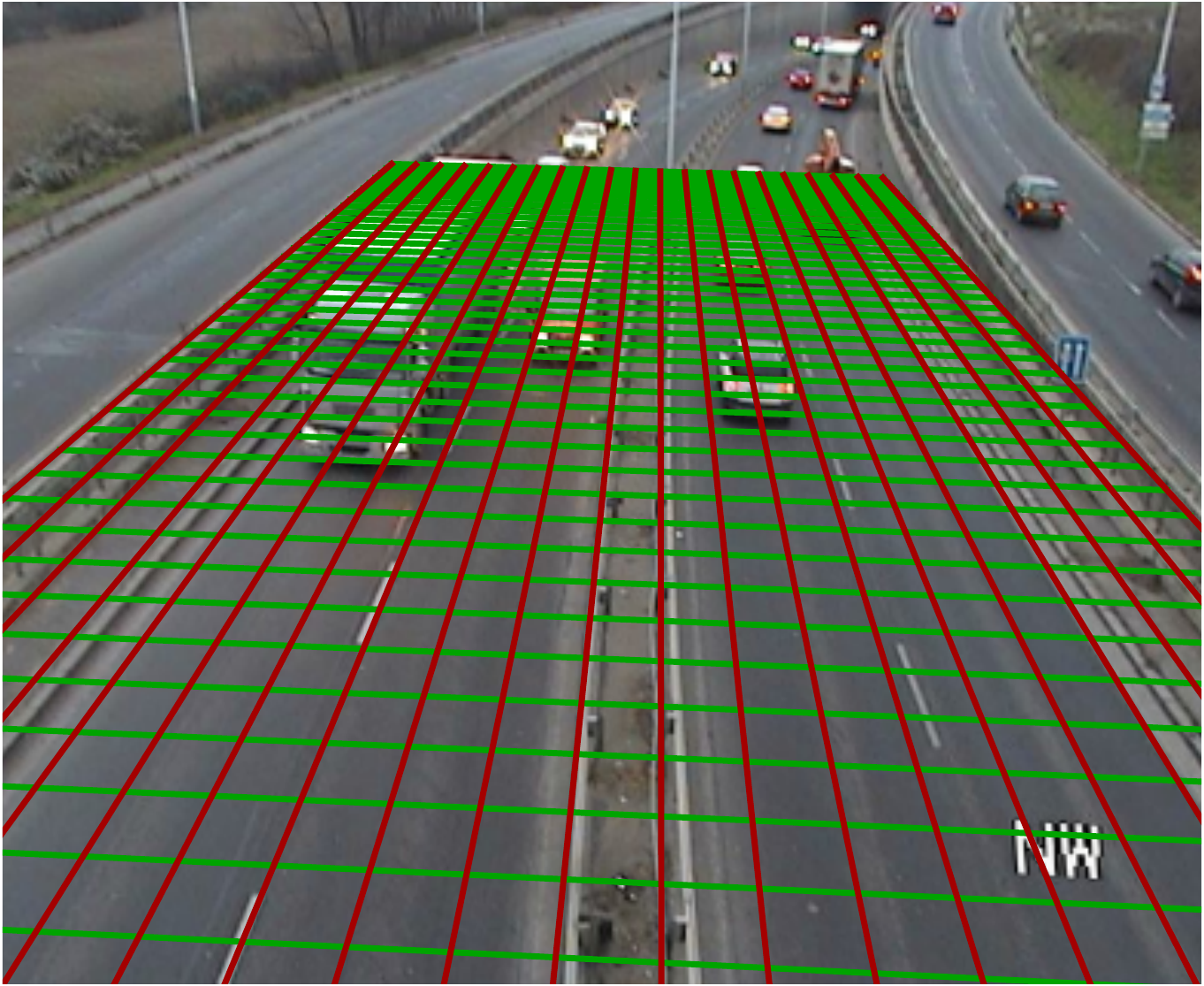}\hspace{2pt}%
		\includegraphics[width=0.45\linewidth]{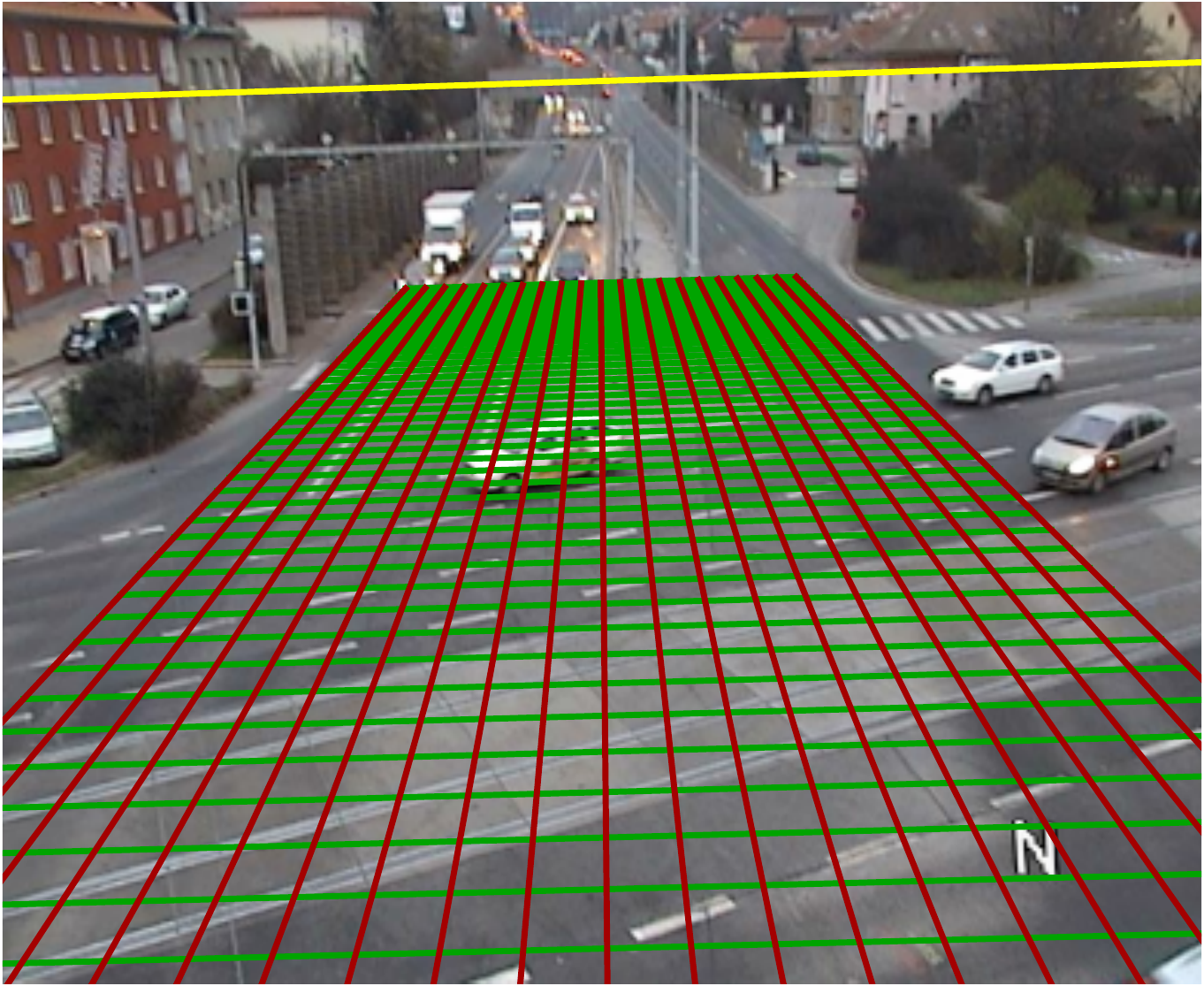}\hspace{2pt}%
		\includegraphics[width=0.45\linewidth]{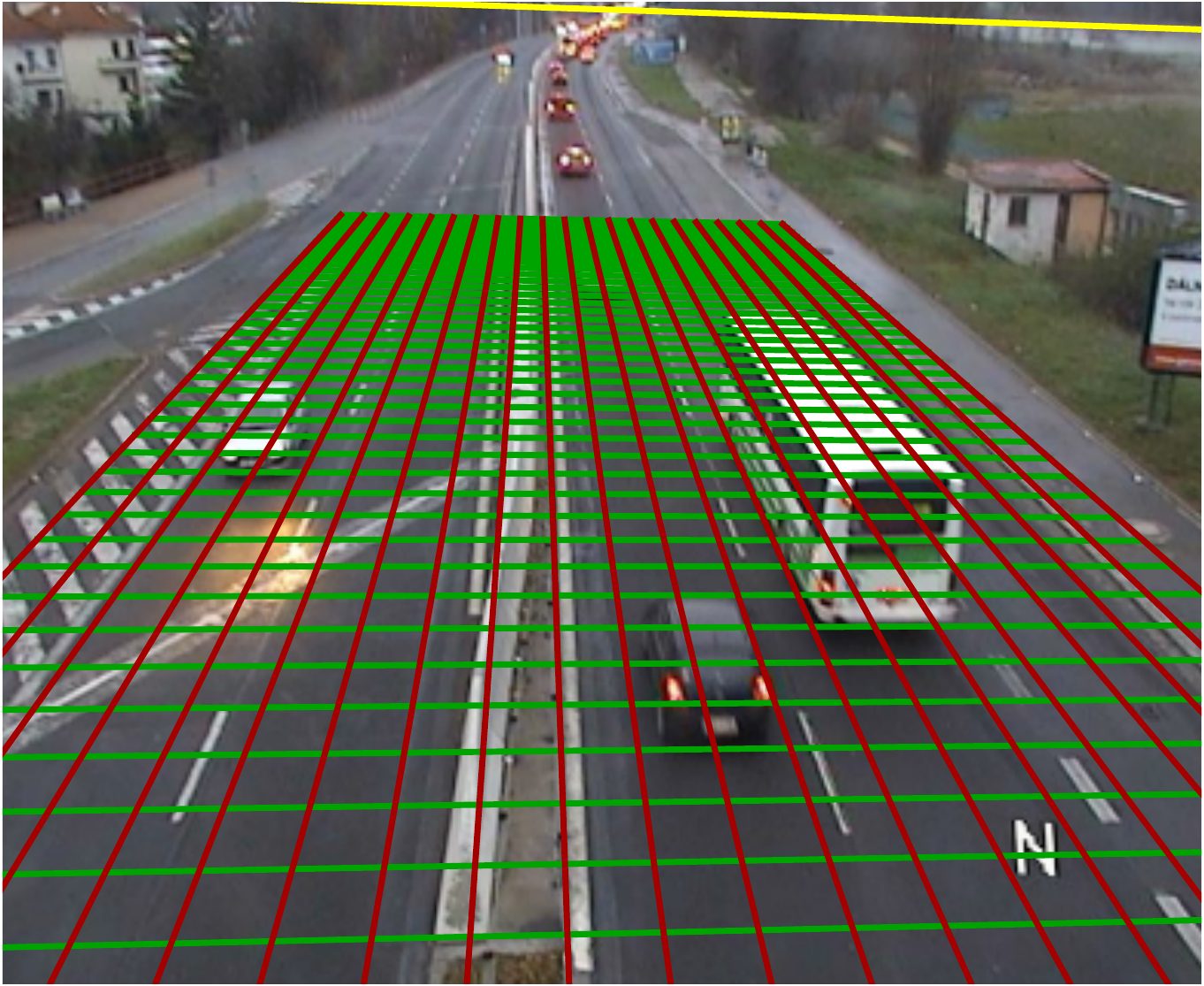}%
	}
	\makebox[\textwidth][c]{%
		\includegraphics[width=0.45\linewidth]{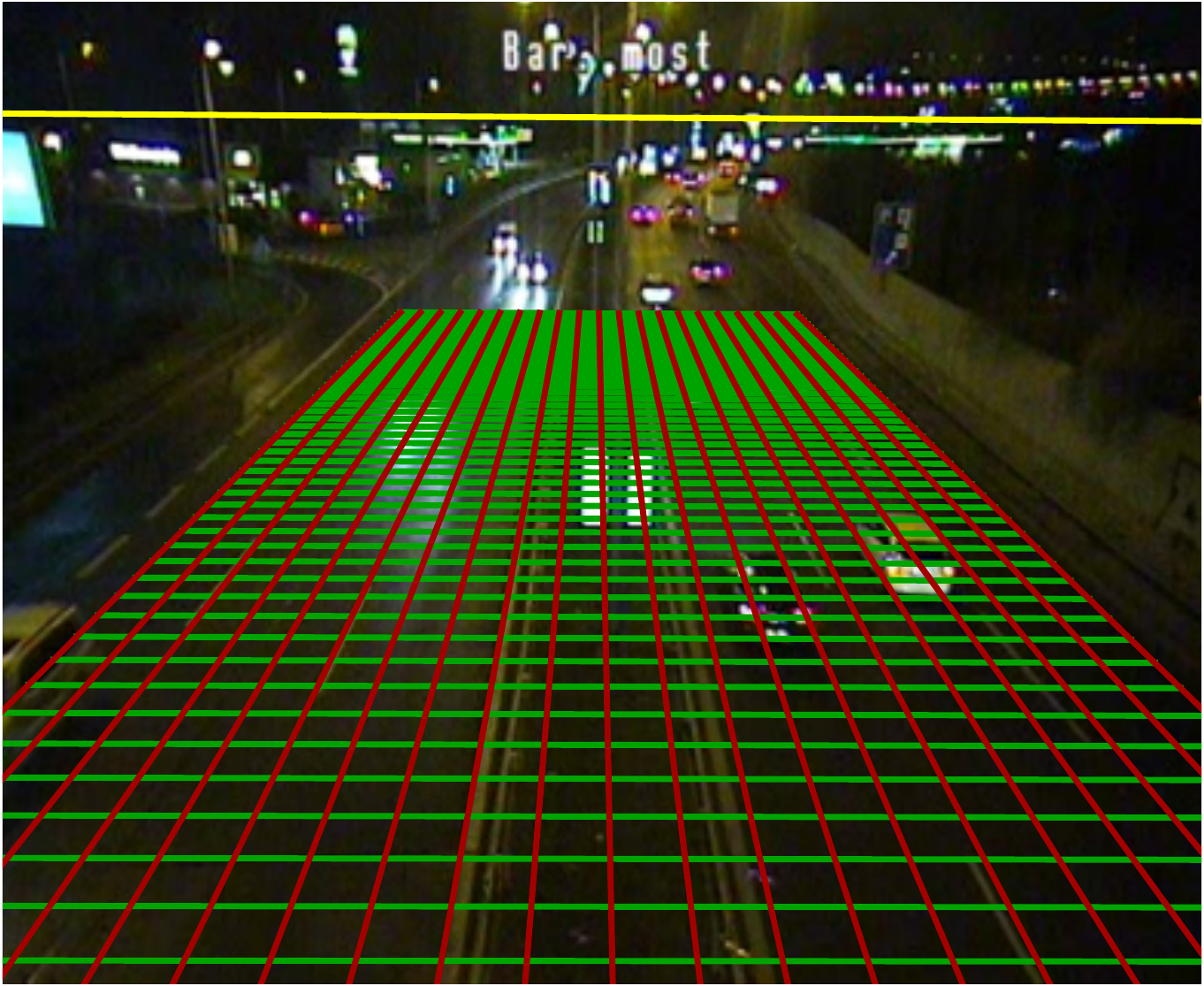}\hspace{2pt}%
		\includegraphics[width=0.45\linewidth]{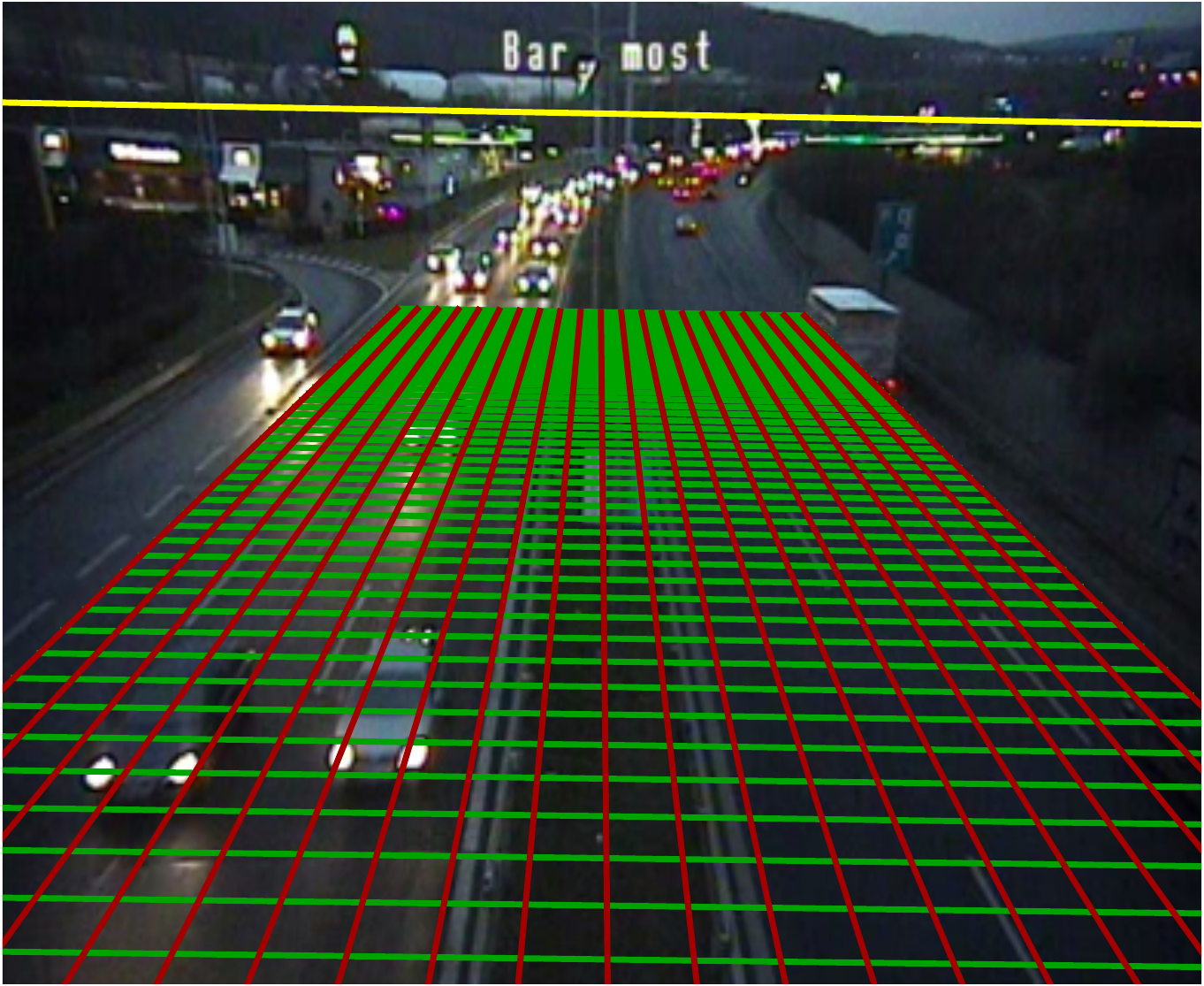}\hspace{2pt}%
		\includegraphics[width=0.45\linewidth]{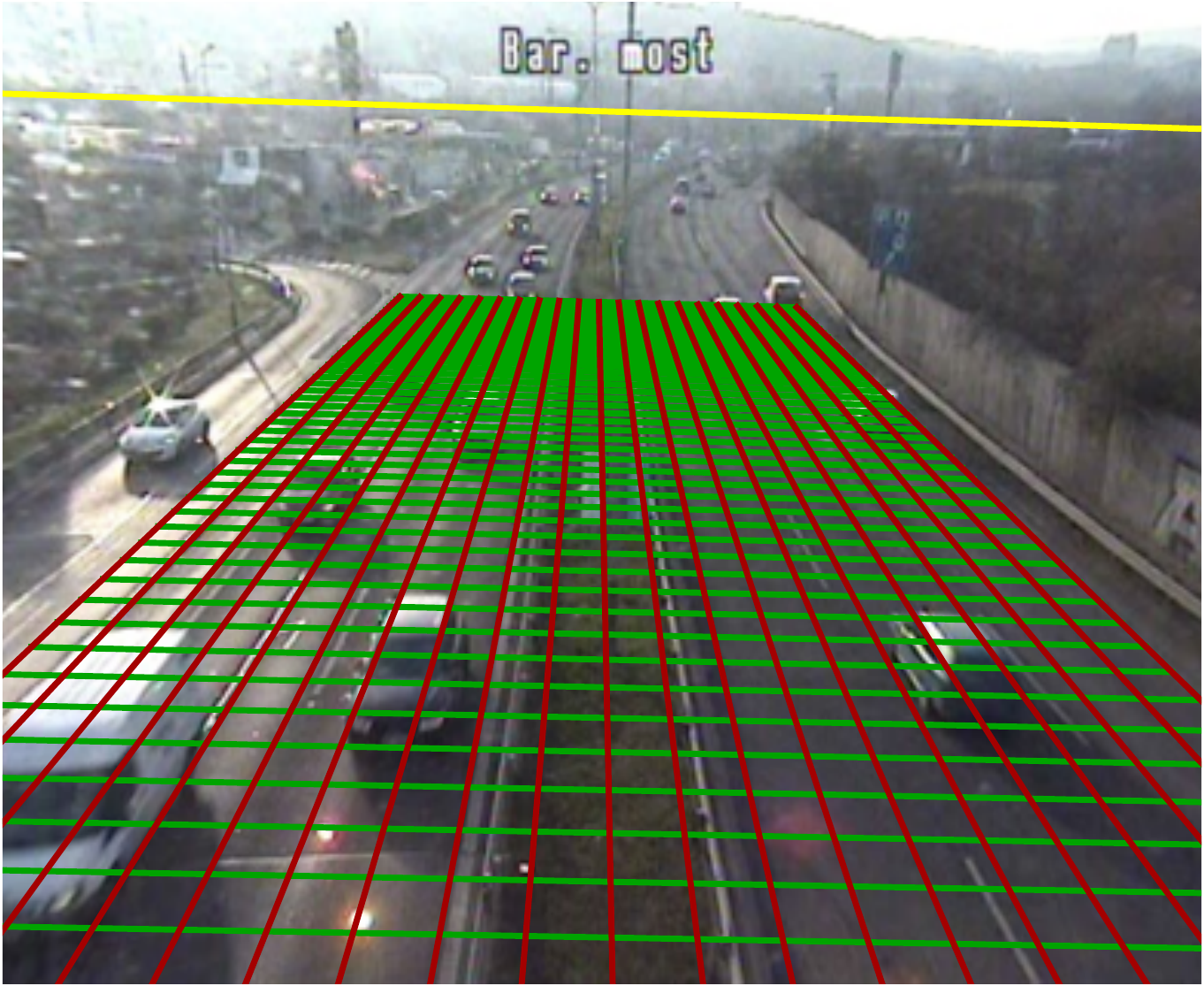}%
	}
	\caption{Example of camera calibration (two vanishing points) for real world surveillance cameras. The first row shows different locations, while the second one show the same locations at night, dawn, and during daylight. The yellow line denotes the detected horizon (if present inside the frames) and red-green grid is formed by lines going to the first vanishing point (red) and to the second one (green). In an ideal case the grid is perpendicular in the real world and the lines are parallel to the features which define the vanishing points on the ground (e.g. line marking). It should also be noted that the method is able to work even on an intersection (top center). }
	\label{fig:RealWorldCalib}
\end{figure*}

The automatic calibration from vehicle movement can be justifiably suspected of requiring idealized conditions and to be sensitive to bad lighting, etc.
In order to verify the usability of our camera calibration method in real-world conditions, we obtained data from surveillance cameras in production use at 9 different locations. The videos were captured both at day and night conditions. The data are of rather poor quality ($704 \times 576\,\mathrm{px}$ or $704 \times 288\,\mathrm{px}$) with 6 frames per second and a mean length of 40s.
As the ground truth calibration is not available for the data, we report only qualitative results in the form of equilateral grid projected on the road plane.
Despite the challenging character of the sequences (poor video quality and lighting conditions), we were able to correctly detect the vanishing points, as can be seen in Figure~\ref{fig:RealWorldCalib} on a few examples, and thus find the camera parameters and its orientation, which is important in many real-world surveillance applications (e.g estimation of vehicle viewpoints or image rectification).

\section{Conclusions}

We propose a fully automatic method for traffic surveillance camera calibration. It does not have any constraints on camera placement and does not require any manual input whatsoever. The results show that our system decreases the mean speed measurement error by $86\,\%$ (7.98\,km/h to 1.10\,km/h) compared to the previous automatic state-of-the-art method and by $19\,\%$ (1.35\,km/h to 1.10\,km/h) compared to the manual calibration method. This improvement is important, as in the previous approaches, automation always compromised accuracy, forcing the system developer to trade off between them.  Our work shows that fully automatic calibration methods may produce better results than manual calibration. This result can be important beyond the field of traffic surveillance, since different forms of manual camera calibration are often considered the ``ground truth'', but our work shows that automatic calibration from statistics of repeated inaccurate measurements can be more precise, despite requiring no user input.
Our method removes the necessity of per-camera setting or calibration, but it still requires some human annotations per coarse geographic region (e.g. European Union or the USA) and per time period when the car models get vastly replaced (e.g. per decade). 

In the experiments, we also showed that our method is able to calibrate real world traffic surveillance cameras and our proposed method for vehicle detection and tracking significantly reduces the number of false positives compared to the previous method.
In future work, we would like to simplify the system and remove the necessity to render the vehicles by approximation of the bounding box size with a function parametrized by viewpoint and image location.

\section*{Acknowledgments}
This work was supported by The Ministry of Education, Youth and Sports of the Czech Republic from the National Programme of Sustainability (NPU II); project IT4Innovations excellence in science -- LQ1602.

We would also like to thank to company CAMEA for providing us data from industrial surveillance cameras.

\section*{References}
\bibliography{2016-CVIU-RobustTraffic-bibliography}

\begin{thebibliography}{51}
\expandafter\ifx\csname natexlab\endcsname\relax\def\natexlab#1{#1}\fi
\providecommand{\url}[1]{\texttt{#1}}
\providecommand{\href}[2]{#2}
\providecommand{\path}[1]{#1}
\providecommand{\DOIprefix}{doi:}
\providecommand{\ArXivprefix}{arXiv:}
\providecommand{\URLprefix}{URL: }
\providecommand{\Pubmedprefix}{pmid:}
\providecommand{\doi}[1]{\href{http://dx.doi.org/#1}{\path{#1}}}
\providecommand{\Pubmed}[1]{\href{pmid:#1}{\path{#1}}}
\providecommand{\bibinfo}[2]{#2}
\ifx\xfnm\undefined \def\xfnm[#1]{\unskip,\space#1}\fi
%Type = Inproceedings
\bibitem[{Cathey and Dailey(2005)}]{Cathey2005}
\bibinfo{author}{Cathey\xfnm[ F.]}, \bibinfo{author}{Dailey\xfnm[ D.]}.
\newblock \bibinfo{title}{A novel technique to dynamically measure vehicle
  speed using uncalibrated roadway cameras}.
\newblock In: \bibinfo{booktitle}{Intelligent Vehicles Symposium}.
  \bibinfo{year}{2005}. p. \bibinfo{pages}{777--782}.
%Type = Article
\bibitem[{Chaperon et~al.(2011)Chaperon, Droulez and Thibault}]{Chaperon2011}
\bibinfo{author}{Chaperon\xfnm[ T.]}, \bibinfo{author}{Droulez\xfnm[ J.]},
  \bibinfo{author}{Thibault\xfnm[ G.]}.
\newblock \bibinfo{title}{Reliable camera pose and calibration from a small set
  of point and line correspondences: A probabilistic approach}.
\newblock \bibinfo{journal}{Computer Vision and Image Understanding}
  \bibinfo{year}{2011};\bibinfo{volume}{115}(\bibinfo{number}{5}):\bibinfo{pages}{576
  -- 585}.
\newblock \bibinfo{note}{Special issue on 3D Imaging and Modelling}.
%Type = Article
\bibitem[{Cootes et~al.(1995)Cootes, Taylor, Cooper and Graham}]{Cootes1995}
\bibinfo{author}{Cootes\xfnm[ T.F.]}, \bibinfo{author}{Taylor\xfnm[ C.J.]},
  \bibinfo{author}{Cooper\xfnm[ D.H.]}, \bibinfo{author}{Graham\xfnm[ J.]}.
\newblock \bibinfo{title}{Active shape models-their training and application}.
\newblock \bibinfo{journal}{Computer vision and image understanding}
  \bibinfo{year}{1995};\bibinfo{volume}{61}(\bibinfo{number}{1}):\bibinfo{pages}{38--59}.
%Type = Article
\bibitem[{Dailey et~al.(2000)Dailey, Cathey and Pumrin}]{Dailey2000}
\bibinfo{author}{Dailey\xfnm[ D.]}, \bibinfo{author}{Cathey\xfnm[ F.]},
  \bibinfo{author}{Pumrin\xfnm[ S.]}.
\newblock \bibinfo{title}{An algorithm to estimate mean traffic speed using
  uncalibrated cameras}.
\newblock \bibinfo{journal}{IEEE Transactions on Intelligent Transportation
  Systems}
  \bibinfo{year}{2000};\bibinfo{volume}{1}(\bibinfo{number}{2}):\bibinfo{pages}{98--107}.
%Type = Inproceedings
\bibitem[{Dalal and Triggs(2005)}]{Dalal2005}
\bibinfo{author}{Dalal\xfnm[ N.]}, \bibinfo{author}{Triggs\xfnm[ B.]}.
\newblock \bibinfo{title}{Histograms of oriented gradients for human
  detection}.
\newblock In: \bibinfo{booktitle}{Computer Vision and Pattern Recognition,
  2005. CVPR 2005. IEEE Computer Society Conference on}.
  \bibinfo{organization}{IEEE}; volume~\bibinfo{volume}{1};
  \bibinfo{year}{2005}. p. \bibinfo{pages}{886--893}.
%Type = Inproceedings
\bibitem[{Do et~al.(2015)Do, Nghiem, Thi and Ngoc}]{Do2015}
\bibinfo{author}{Do\xfnm[ V.H.]}, \bibinfo{author}{Nghiem\xfnm[ L.H.]},
  \bibinfo{author}{Thi\xfnm[ N.P.]}, \bibinfo{author}{Ngoc\xfnm[ N.P.]}.
\newblock \bibinfo{title}{A simple camera calibration method for vehicle
  velocity estimation}.
\newblock In: \bibinfo{booktitle}{Electrical Engineering/Electronics, Computer,
  Telecommunications and Information Technology (ECTI-CON), 2015 12th
  International Conference on}. \bibinfo{year}{2015}. p. \bibinfo{pages}{1--5}.
%Type = Article
\bibitem[{Dollár et~al.(2014)Dollár, Appel, Belongie and Perona}]{Dollar2014}
\bibinfo{author}{Dollár\xfnm[ P.]}, \bibinfo{author}{Appel\xfnm[ R.]},
  \bibinfo{author}{Belongie\xfnm[ S.]}, \bibinfo{author}{Perona\xfnm[ P.]}.
\newblock \bibinfo{title}{Fast feature pyramids for object detection}.
\newblock \bibinfo{journal}{IEEE Transactions on Pattern Analysis and Machine
  Intelligence}
  \bibinfo{year}{2014};\bibinfo{volume}{36}(\bibinfo{number}{8}):\bibinfo{pages}{1532--1545}.
%Type = Inproceedings
\bibitem[{Dubsk\'{a} et~al.(2014)Dubsk\'{a}, Sochor and Herout}]{Dubska2014}
\bibinfo{author}{Dubsk\'{a}\xfnm[ M.]}, \bibinfo{author}{Sochor\xfnm[ J.]},
  \bibinfo{author}{Herout\xfnm[ A.]}.
\newblock \bibinfo{title}{Automatic camera calibration for traffic
  understanding}.
\newblock In: \bibinfo{booktitle}{BMVC}. \bibinfo{year}{2014}. .
%Type = Inproceedings
\bibitem[{Dubská and Herout(2013)}]{Dubska2013}
\bibinfo{author}{Dubská\xfnm[ M.]}, \bibinfo{author}{Herout\xfnm[ A.]}.
\newblock \bibinfo{title}{Real projective plane mapping for detection of
  orthogonal vanishing points}.
\newblock In: \bibinfo{booktitle}{Proceedings of the British Machine Vision
  Conference}. \bibinfo{publisher}{BMVA Press}; \bibinfo{year}{2013}. .
%Type = Article
\bibitem[{Dubská et~al.(2015)Dubská, Herout, Juranek and
  Sochor}]{Dubska2015ITS}
\bibinfo{author}{Dubská\xfnm[ M.]}, \bibinfo{author}{Herout\xfnm[ A.]},
  \bibinfo{author}{Juranek\xfnm[ R.]}, \bibinfo{author}{Sochor\xfnm[ J.]}.
\newblock \bibinfo{title}{Fully automatic roadside camera calibration for
  traffic surveillance}.
\newblock \bibinfo{journal}{Intelligent Transportation Systems, IEEE
  Transactions on}
  \bibinfo{year}{2015};\bibinfo{volume}{16}(\bibinfo{number}{3}):\bibinfo{pages}{1162--1171}.
%Type = Article
\bibitem[{Fang et~al.(2016)Fang, Zhou, Yu and Du}]{Fang2016}
\bibinfo{author}{Fang\xfnm[ J.]}, \bibinfo{author}{Zhou\xfnm[ Y.]},
  \bibinfo{author}{Yu\xfnm[ Y.]}, \bibinfo{author}{Du\xfnm[ S.]}.
\newblock \bibinfo{title}{Fine-grained vehicle model recognition using a
  coarse-to-fine convolutional neural network architecture}.
\newblock \bibinfo{journal}{IEEE Transactions on Intelligent Transportation
  Systems}
  \bibinfo{year}{2016};\bibinfo{volume}{PP}(\bibinfo{number}{99}):\bibinfo{pages}{1--11}.
%Type = Article
\bibitem[{Felzenszwalb et~al.(2010)Felzenszwalb, Girshick, McAllester and
  Ramanan}]{Felzenszwalb2010}
\bibinfo{author}{Felzenszwalb\xfnm[ P.]}, \bibinfo{author}{Girshick\xfnm[ R.]},
  \bibinfo{author}{McAllester\xfnm[ D.]}, \bibinfo{author}{Ramanan\xfnm[ D.]}.
\newblock \bibinfo{title}{Object detection with discriminatively trained
  part-based models}.
\newblock \bibinfo{journal}{IEEE Transactions on Pattern Analysis and Machine
  Intelligence}
  \bibinfo{year}{2010};\bibinfo{volume}{32}(\bibinfo{number}{9}):\bibinfo{pages}{1627--1645}.
%Type = Article
\bibitem[{Fung et~al.(2003)Fung, Yung and Pang}]{Fung2003}
\bibinfo{author}{Fung\xfnm[ G.S.K.]}, \bibinfo{author}{Yung\xfnm[ N.H.C.]},
  \bibinfo{author}{Pang\xfnm[ G.K.H.]}.
\newblock \bibinfo{title}{Camera calibration from road lane markings}.
\newblock \bibinfo{journal}{Optical Engineering}
  \bibinfo{year}{2003};\bibinfo{volume}{42}(\bibinfo{number}{10}):\bibinfo{pages}{2967--2977}.
%Type = Inproceedings
\bibitem[{Gao et~al.(2016)Gao, Beijbom, Zhang and Darrell}]{Gao2016}
\bibinfo{author}{Gao\xfnm[ Y.]}, \bibinfo{author}{Beijbom\xfnm[ O.]},
  \bibinfo{author}{Zhang\xfnm[ N.]}, \bibinfo{author}{Darrell\xfnm[ T.]}.
\newblock \bibinfo{title}{Compact bilinear pooling}.
\newblock In: \bibinfo{booktitle}{The IEEE Conference on Computer Vision and
  Pattern Recognition (CVPR)}. \bibinfo{year}{2016}. .
%Type = Inproceedings
\bibitem[{Girshick et~al.(2014)Girshick, Donahue, Darrell and
  Malik}]{Girshick2014}
\bibinfo{author}{Girshick\xfnm[ R.]}, \bibinfo{author}{Donahue\xfnm[ J.]},
  \bibinfo{author}{Darrell\xfnm[ T.]}, \bibinfo{author}{Malik\xfnm[ J.]}.
\newblock \bibinfo{title}{Rich feature hierarchies for accurate object
  detection and semantic segmentation}.
\newblock In: \bibinfo{booktitle}{Computer Vision and Pattern Recognition}.
  \bibinfo{year}{2014}. .
%Type = Inproceedings
\bibitem[{Grammatikopoulos et~al.(2005)Grammatikopoulos, Karras and
  Petsa}]{Grammatikopoulos2005}
\bibinfo{author}{Grammatikopoulos\xfnm[ L.]}, \bibinfo{author}{Karras\xfnm[
  G.]}, \bibinfo{author}{Petsa\xfnm[ E.]}.
\newblock \bibinfo{title}{Automatic estimation of vehicle speed from
  uncalibrated video sequences}.
\newblock In: \bibinfo{booktitle}{Proceedings of International Symposium on
  Modern Technologies, Educationand Profeesional Practice in Geodesy and
  Related Fields}. \bibinfo{year}{2005}. p. \bibinfo{pages}{332--338}.
%Type = Article
\bibitem[{He and Yung(2007{\natexlab{a}})}]{He2007CalibMethod}
\bibinfo{author}{He\xfnm[ X.]}, \bibinfo{author}{Yung\xfnm[ N.]}.
\newblock \bibinfo{title}{New method for overcoming ill-conditioning in
  vanishing-point-based camera calibration}.
\newblock \bibinfo{journal}{Optical Engineering}
  \bibinfo{year}{2007}{\natexlab{a}};\bibinfo{volume}{46}(\bibinfo{number}{3}).
%Type = Inproceedings
\bibitem[{He and Yung(2007{\natexlab{b}})}]{He2007}
\bibinfo{author}{He\xfnm[ X.C.]}, \bibinfo{author}{Yung\xfnm[ N.H.C.]}.
\newblock \bibinfo{title}{A novel algorithm for estimating vehicle speed from
  two consecutive images}.
\newblock In: \bibinfo{booktitle}{IEEE Workshop on Applications of Computer
  Vision, WACV}. \bibinfo{year}{2007}{\natexlab{b}}. .
%Type = Inproceedings
\bibitem[{Hsiao et~al.(2014)Hsiao, Sinha, Ramnath, Baker, Zitnick and
  Szeliski}]{Hsiao2014}
\bibinfo{author}{Hsiao\xfnm[ E.]}, \bibinfo{author}{Sinha\xfnm[ S.]},
  \bibinfo{author}{Ramnath\xfnm[ K.]}, \bibinfo{author}{Baker\xfnm[ S.]},
  \bibinfo{author}{Zitnick\xfnm[ L.]}, \bibinfo{author}{Szeliski\xfnm[ R.]}.
\newblock \bibinfo{title}{Car make and model recognition using {3D} curve
  alignment}.
\newblock In: \bibinfo{booktitle}{IEEE WACV}. \bibinfo{year}{2014}. .
%Type = Inproceedings
\bibitem[{Juránek et~al.(2015)Juránek, Herout, Dubská and
  Zemčík}]{Juranek2015}
\bibinfo{author}{Juránek\xfnm[ R.]}, \bibinfo{author}{Herout\xfnm[ A.]},
  \bibinfo{author}{Dubská\xfnm[ M.]}, \bibinfo{author}{Zemčík\xfnm[ P.]}.
\newblock \bibinfo{title}{Real-time pose estimation piggybacked on object
  detection}.
\newblock In: \bibinfo{booktitle}{The IEEE International Conference on Computer
  Vision (ICCV)}. \bibinfo{year}{2015}. .
%Type = Article
\bibitem[{Kalman(1960)}]{Kalman1960}
\bibinfo{author}{Kalman\xfnm[ R.E.]}.
\newblock \bibinfo{title}{A new approach to linear filtering and prediction
  problems}.
\newblock \bibinfo{journal}{Transactions of the ASME – Journal of Basic
  Engineering} \bibinfo{year}{1960};(\bibinfo{number}{82 (Series
  D)}):\bibinfo{pages}{35--45}.
%Type = Inproceedings
\bibitem[{Krause et~al.(2015)Krause, Jin, Yang and Fei-Fei}]{Krause2015}
\bibinfo{author}{Krause\xfnm[ J.]}, \bibinfo{author}{Jin\xfnm[ H.]},
  \bibinfo{author}{Yang\xfnm[ J.]}, \bibinfo{author}{Fei-Fei\xfnm[ L.]}.
\newblock \bibinfo{title}{Fine-grained recognition without part annotations}.
\newblock In: \bibinfo{booktitle}{IEEE Conference on Computer Vision and
  Pattern Recognition (CVPR)}. \bibinfo{year}{2015}. .
%Type = Inproceedings
\bibitem[{Krause et~al.(2013)Krause, Stark, Deng and Fei-Fei}]{Krause2013}
\bibinfo{author}{Krause\xfnm[ J.]}, \bibinfo{author}{Stark\xfnm[ M.]},
  \bibinfo{author}{Deng\xfnm[ J.]}, \bibinfo{author}{Fei-Fei\xfnm[ L.]}.
\newblock \bibinfo{title}{{3D} object representations for fine-grained
  categorization}.
\newblock In: \bibinfo{booktitle}{ICCV Workshop 3dRR-13}. \bibinfo{year}{2013}.
  .
%Type = Incollection
\bibitem[{Krizhevsky et~al.(2012)Krizhevsky, Sutskever and
  Hinton}]{Krizhevsky2012}
\bibinfo{author}{Krizhevsky\xfnm[ A.]}, \bibinfo{author}{Sutskever\xfnm[ I.]},
  \bibinfo{author}{Hinton\xfnm[ G.E.]}.
\newblock \bibinfo{title}{Imagenet classification with deep convolutional
  neural networks}.
\newblock In: \bibinfo{editor}{Pereira\xfnm[ F.]},
  \bibinfo{editor}{Burges\xfnm[ C.]}, \bibinfo{editor}{Bottou\xfnm[ L.]},
  \bibinfo{editor}{Weinberger\xfnm[ K.]}, editors. \bibinfo{booktitle}{Advances
  in Neural Information Processing Systems 25}. \bibinfo{publisher}{Curran
  Associates, Inc.}; \bibinfo{year}{2012}. p. \bibinfo{pages}{1097--1105}.
%Type = Article
\bibitem[{Lan et~al.(2014)Lan, Li, Hu, Ran and Wang}]{Lan2014}
\bibinfo{author}{Lan\xfnm[ J.]}, \bibinfo{author}{Li\xfnm[ J.]},
  \bibinfo{author}{Hu\xfnm[ G.]}, \bibinfo{author}{Ran\xfnm[ B.]},
  \bibinfo{author}{Wang\xfnm[ L.]}.
\newblock \bibinfo{title}{Vehicle speed measurement based on gray constraint
  optical flow algorithm}.
\newblock \bibinfo{journal}{Optik - International Journal for Light and
  Electron Optics}
  \bibinfo{year}{2014};\bibinfo{volume}{125}(\bibinfo{number}{1}):\bibinfo{pages}{289
  -- 295}.
%Type = Inproceedings
\bibitem[{Li et~al.(2009)Li, Gatenby, Wang and Gore}]{Li2009}
\bibinfo{author}{Li\xfnm[ C.]}, \bibinfo{author}{Gatenby\xfnm[ C.]},
  \bibinfo{author}{Wang\xfnm[ L.]}, \bibinfo{author}{Gore\xfnm[ J.C.]}.
\newblock \bibinfo{title}{A robust parametric method for bias field estimation
  and segmentation of mr images}.
\newblock In: \bibinfo{booktitle}{Computer Vision and Pattern Recognition,
  2009. CVPR 2009. IEEE Conference on}. \bibinfo{organization}{IEEE};
  \bibinfo{year}{2009}. p. \bibinfo{pages}{218--223}.
%Type = Inproceedings
\bibitem[{Lin et~al.(2015)Lin, RoyChowdhury and Maji}]{Lin2015Bilinear}
\bibinfo{author}{Lin\xfnm[ T.Y.]}, \bibinfo{author}{RoyChowdhury\xfnm[ A.]},
  \bibinfo{author}{Maji\xfnm[ S.]}.
\newblock \bibinfo{title}{Bilinear cnn models for fine-grained visual
  recognition}.
\newblock In: \bibinfo{booktitle}{International Conference on Computer Vision
  (ICCV)}. \bibinfo{year}{2015}. .
%Type = Inproceedings
\bibitem[{Lin et~al.(2014)Lin, Morariu, Hsu and Davis}]{Lin2014}
\bibinfo{author}{Lin\xfnm[ Y.L.]}, \bibinfo{author}{Morariu\xfnm[ V.I.]},
  \bibinfo{author}{Hsu\xfnm[ W.]}, \bibinfo{author}{Davis\xfnm[ L.S.]}.
\newblock \bibinfo{title}{Jointly optimizing {3D} model fitting and
  fine-grained classification}.
\newblock In: \bibinfo{booktitle}{ECCV}. \bibinfo{year}{2014}. .
%Type = Incollection
\bibitem[{Liu et~al.(2012)Liu, Kanazawa, Jacobs and Belhumeur}]{Liu2012}
\bibinfo{author}{Liu\xfnm[ J.]}, \bibinfo{author}{Kanazawa\xfnm[ A.]},
  \bibinfo{author}{Jacobs\xfnm[ D.]}, \bibinfo{author}{Belhumeur\xfnm[ P.]}.
\newblock \bibinfo{title}{Dog breed classification using part localization}.
\newblock In: \bibinfo{editor}{Fitzgibbon\xfnm[ A.]},
  \bibinfo{editor}{Lazebnik\xfnm[ S.]}, \bibinfo{editor}{Perona\xfnm[ P.]},
  \bibinfo{editor}{Sato\xfnm[ Y.]}, \bibinfo{editor}{Schmid\xfnm[ C.]},
  editors. \bibinfo{booktitle}{ECCV 2012}. \bibinfo{publisher}{Springer Berlin
  Heidelberg}; volume \bibinfo{volume}{7572} of
  \textit{\bibinfo{series}{Lecture Notes in Computer Science}};
  \bibinfo{year}{2012}. p. \bibinfo{pages}{172--185}.
%Type = Inproceedings
\bibitem[{Liu et~al.(2016)Liu, Anguelov, Erhan, Szegedy, Reed, Fu and
  Berg}]{Liu2016SSD}
\bibinfo{author}{Liu\xfnm[ W.]}, \bibinfo{author}{Anguelov\xfnm[ D.]},
  \bibinfo{author}{Erhan\xfnm[ D.]}, \bibinfo{author}{Szegedy\xfnm[ C.]},
  \bibinfo{author}{Reed\xfnm[ S.]}, \bibinfo{author}{Fu\xfnm[ C.Y.]},
  \bibinfo{author}{Berg\xfnm[ A.C.]}.
\newblock \bibinfo{title}{{SSD}: Single shot multibox detector}.
\newblock \bibinfo{year}{2016}. \bibinfo{note}{To appear.}
%Type = Inproceedings
\bibitem[{Long et~al.(2015)Long, Shelhamer and Darrell}]{Long_2015_CVPR}
\bibinfo{author}{Long\xfnm[ J.]}, \bibinfo{author}{Shelhamer\xfnm[ E.]},
  \bibinfo{author}{Darrell\xfnm[ T.]}.
\newblock \bibinfo{title}{Fully convolutional networks for semantic
  segmentation}.
\newblock In: \bibinfo{booktitle}{The IEEE Conference on Computer Vision and
  Pattern Recognition (CVPR)}. \bibinfo{year}{2015}. .
%Type = Inproceedings
\bibitem[{Lowe(1999)}]{Lowe1999}
\bibinfo{author}{Lowe\xfnm[ D.G.]}.
\newblock \bibinfo{title}{Object recognition from local scale-invariant
  features}.
\newblock In: \bibinfo{booktitle}{Computer vision, 1999. The proceedings of the
  seventh IEEE international conference on}. \bibinfo{organization}{Ieee};
  volume~\bibinfo{volume}{2}; \bibinfo{year}{1999}. p.
  \bibinfo{pages}{1150--1157}.
%Type = Inproceedings
\bibitem[{Luvizon et~al.(2014)Luvizon, Nassu and Minetto}]{Luvizon2014}
\bibinfo{author}{Luvizon\xfnm[ D.]}, \bibinfo{author}{Nassu\xfnm[ B.]},
  \bibinfo{author}{Minetto\xfnm[ R.]}.
\newblock \bibinfo{title}{Vehicle speed estimation by license plate detection
  and tracking}.
\newblock In: \bibinfo{booktitle}{Acoustics, Speech and Signal Processing
  (ICASSP), 2014 IEEE International Conference on}. \bibinfo{year}{2014}. p.
  \bibinfo{pages}{6563--6567}.
%Type = Article
\bibitem[{Luvizon et~al.(2016)Luvizon, Nassu and Minetto}]{Luvizon2016}
\bibinfo{author}{Luvizon\xfnm[ D.C.]}, \bibinfo{author}{Nassu\xfnm[ B.T.]},
  \bibinfo{author}{Minetto\xfnm[ R.]}.
\newblock \bibinfo{title}{A video-based system for vehicle speed measurement in
  urban roadways}.
\newblock \bibinfo{journal}{IEEE Transactions on Intelligent Transportation
  Systems}
  \bibinfo{year}{2016};\bibinfo{volume}{PP}(\bibinfo{number}{99}):\bibinfo{pages}{1--12}.
%Type = Inproceedings
\bibitem[{Maduro et~al.(2008)Maduro, Batista, Peixoto and Batista}]{Maduro2008}
\bibinfo{author}{Maduro\xfnm[ C.]}, \bibinfo{author}{Batista\xfnm[ K.]},
  \bibinfo{author}{Peixoto\xfnm[ P.]}, \bibinfo{author}{Batista\xfnm[ J.]}.
\newblock \bibinfo{title}{Estimation of vehicle velocity and traffic intensity
  using rectified images}.
\newblock In: \bibinfo{booktitle}{Image Processing, 2008. ICIP 2008. 15th IEEE
  International Conference on}. \bibinfo{year}{2008}. p.
  \bibinfo{pages}{777--780}.
%Type = Inproceedings
\bibitem[{Nurhadiyatna et~al.(2013)Nurhadiyatna, Hardjono, Wibisono, Sina,
  Jatmiko, Ma'sum and Mursanto}]{Nurhadiyatna2013}
\bibinfo{author}{Nurhadiyatna\xfnm[ A.]}, \bibinfo{author}{Hardjono\xfnm[ B.]},
  \bibinfo{author}{Wibisono\xfnm[ A.]}, \bibinfo{author}{Sina\xfnm[ I.]},
  \bibinfo{author}{Jatmiko\xfnm[ W.]}, \bibinfo{author}{Ma'sum\xfnm[ M.]},
  \bibinfo{author}{Mursanto\xfnm[ P.]}.
\newblock \bibinfo{title}{Improved vehicle speed estimation using gaussian
  mixture model and hole filling algorithm}.
\newblock In: \bibinfo{booktitle}{Advanced Computer Science and Information
  Systems (ICACSIS), 2013 International Conference on}. \bibinfo{year}{2013}.
  p. \bibinfo{pages}{451--456}.
%Type = Inproceedings
\bibitem[{Prokaj and Medioni(2009)}]{Prokaj2009}
\bibinfo{author}{Prokaj\xfnm[ J.]}, \bibinfo{author}{Medioni\xfnm[ G.]}.
\newblock \bibinfo{title}{{3-D} model based vehicle recognition}.
\newblock In: \bibinfo{booktitle}{IEEE WACV}. \bibinfo{year}{2009}. .
%Type = Inproceedings
\bibitem[{Ren et~al.(2015)Ren, He, Girshick and Sun}]{Girshick2015}
\bibinfo{author}{Ren\xfnm[ S.]}, \bibinfo{author}{He\xfnm[ K.]},
  \bibinfo{author}{Girshick\xfnm[ R.]}, \bibinfo{author}{Sun\xfnm[ J.]}.
\newblock \bibinfo{title}{Faster {R-CNN}: Towards real-time object detection
  with region proposal networks}.
\newblock In: \bibinfo{booktitle}{Advances in Neural Information Processing
  Systems ({NIPS})}. \bibinfo{year}{2015}. .
%Type = Article
\bibitem[{Schoepflin and Dailey(2003)}]{Schoepflin2003}
\bibinfo{author}{Schoepflin\xfnm[ T.]}, \bibinfo{author}{Dailey\xfnm[ D.]}.
\newblock \bibinfo{title}{Dynamic camera calibration of roadside traffic
  management cameras for vehicle speed estimation}.
\newblock \bibinfo{journal}{Intelligent Transportation Systems, IEEE
  Transactions on}
  \bibinfo{year}{2003};\bibinfo{volume}{4}(\bibinfo{number}{2}):\bibinfo{pages}{90--98}.
%Type = Inproceedings
\bibitem[{Shi and Tomasi(1994)}]{Shi1994}
\bibinfo{author}{Shi\xfnm[ J.]}, \bibinfo{author}{Tomasi\xfnm[ C.]}.
\newblock \bibinfo{title}{Good features to track}.
\newblock In: \bibinfo{booktitle}{1994 Proceedings of IEEE Conference on
  Computer Vision and Pattern Recognition}. \bibinfo{year}{1994}. p.
  \bibinfo{pages}{593--600}.
%Type = Inproceedings
\bibitem[{Simon and Rodner(2015)}]{Simon2015}
\bibinfo{author}{Simon\xfnm[ M.]}, \bibinfo{author}{Rodner\xfnm[ E.]}.
\newblock \bibinfo{title}{Neural activation constellations: Unsupervised part
  model discovery with convolutional networks}.
\newblock In: \bibinfo{booktitle}{International Conference on Computer Vision
  (ICCV)}. \bibinfo{year}{2015}. .
%Type = Inproceedings
\bibitem[{Sina et~al.(2013)Sina, Wibisono, Nurhadiyatna, Hardjono, Jatmiko and
  Mursanto}]{Sina2013}
\bibinfo{author}{Sina\xfnm[ I.]}, \bibinfo{author}{Wibisono\xfnm[ A.]},
  \bibinfo{author}{Nurhadiyatna\xfnm[ A.]}, \bibinfo{author}{Hardjono\xfnm[
  B.]}, \bibinfo{author}{Jatmiko\xfnm[ W.]}, \bibinfo{author}{Mursanto\xfnm[
  P.]}.
\newblock \bibinfo{title}{Vehicle counting and speed measurement using
  headlight detection}.
\newblock In: \bibinfo{booktitle}{Advanced Computer Science and Information
  Systems (ICACSIS), 2013 International Conference on}. \bibinfo{year}{2013}.
  p. \bibinfo{pages}{149--154}.
%Type = Inproceedings
\bibitem[{Sochor et~al.(2016{\natexlab{a}})Sochor, Herout and
  Havel}]{Sochor2016}
\bibinfo{author}{Sochor\xfnm[ J.]}, \bibinfo{author}{Herout\xfnm[ A.]},
  \bibinfo{author}{Havel\xfnm[ J.]}.
\newblock \bibinfo{title}{{BoxCars}: {3D} boxes as {CNN} input for improved
  fine-grained vehicle recognition}.
\newblock In: \bibinfo{booktitle}{The IEEE Conference on Computer Vision and
  Pattern Recognition (CVPR)}. \bibinfo{year}{2016}{\natexlab{a}}. .
%Type = Article
\bibitem[{Sochor et~al.(2016{\natexlab{b}})Sochor, Juranek, Spanhel, Marsik,
  Siroky, Herout and Zemcik}]{BrnoCompSpeed}
\bibinfo{author}{Sochor\xfnm[ J.]}, \bibinfo{author}{Juranek\xfnm[ R.]},
  \bibinfo{author}{Spanhel\xfnm[ J.]}, \bibinfo{author}{Marsik\xfnm[ L.]},
  \bibinfo{author}{Siroky\xfnm[ A.]}, \bibinfo{author}{Herout\xfnm[ A.]},
  \bibinfo{author}{Zemcik\xfnm[ P.]}.
\newblock \bibinfo{title}{{BrnoCompSpeed}: Review of traffic camera calibration
  and a comprehensive dataset for monocular speed measurement}.
\newblock \bibinfo{journal}{Intelligent Transportation Systems (under review),
  IEEE Transactions on} \bibinfo{year}{2016}{\natexlab{b}};.
%Type = Techreport
\bibitem[{Tomasi and Kanade(1991)}]{Tomasi1991}
\bibinfo{author}{Tomasi\xfnm[ C.]}, \bibinfo{author}{Kanade\xfnm[ T.]}.
\newblock \bibinfo{title}{Detection and Tracking of Point Features}.
\newblock \bibinfo{type}{Technical Report}; International Journal of Computer
  Vision; \bibinfo{year}{1991}.
%Type = Inproceedings
\bibitem[{Yang et~al.(2016)Yang, Price, Cohen, Lee and Yang}]{Yang_2016_CVPR}
\bibinfo{author}{Yang\xfnm[ J.]}, \bibinfo{author}{Price\xfnm[ B.]},
  \bibinfo{author}{Cohen\xfnm[ S.]}, \bibinfo{author}{Lee\xfnm[ H.]},
  \bibinfo{author}{Yang\xfnm[ M.H.]}.
\newblock \bibinfo{title}{Object contour detection with a fully convolutional
  encoder-decoder network}.
\newblock In: \bibinfo{booktitle}{The IEEE Conference on Computer Vision and
  Pattern Recognition (CVPR)}. \bibinfo{year}{2016}. .
%Type = Article
\bibitem[{You and Zheng(2016)}]{You2016}
\bibinfo{author}{You\xfnm[ X.]}, \bibinfo{author}{Zheng\xfnm[ Y.]}.
\newblock \bibinfo{title}{An accurate and practical calibration method for
  roadside camera using two vanishing points}.
\newblock \bibinfo{journal}{Neurocomputing} \bibinfo{year}{2016};.
%Type = Article
\bibitem[{Yu et~al.(2009)Yu, Jiang, Cheong, Leong and Yan}]{Yu2009}
\bibinfo{author}{Yu\xfnm[ X.]}, \bibinfo{author}{Jiang\xfnm[ N.]},
  \bibinfo{author}{Cheong\xfnm[ L.F.]}, \bibinfo{author}{Leong\xfnm[ H.W.]},
  \bibinfo{author}{Yan\xfnm[ X.]}.
\newblock \bibinfo{title}{Automatic camera calibration of broadcast tennis
  video with applications to {3D} virtual content insertion and ball detection
  and tracking}.
\newblock \bibinfo{journal}{Computer Vision and Image Understanding}
  \bibinfo{year}{2009};\bibinfo{volume}{113}(\bibinfo{number}{5}):\bibinfo{pages}{643
  -- 652}.
\newblock \bibinfo{note}{Computer Vision Based Analysis in Sport Environments}.
%Type = Article
\bibitem[{Zhang(2000)}]{Zhang2000}
\bibinfo{author}{Zhang\xfnm[ Z.]}.
\newblock \bibinfo{title}{A flexible new technique for camera calibration}.
\newblock \bibinfo{journal}{IEEE Transactions on Pattern Analysis and Machine
  Intelligence}
  \bibinfo{year}{2000};\bibinfo{volume}{22}(\bibinfo{number}{11}):\bibinfo{pages}{1330--1334}.
%Type = Article
\bibitem[{Zhang et~al.(2013)Zhang, Tan, Huang and Wang}]{Zhang2013Calib}
\bibinfo{author}{Zhang\xfnm[ Z.]}, \bibinfo{author}{Tan\xfnm[ T.]},
  \bibinfo{author}{Huang\xfnm[ K.]}, \bibinfo{author}{Wang\xfnm[ Y.]}.
\newblock \bibinfo{title}{Practical camera calibration from moving objects for
  traffic scene surveillance}.
\newblock \bibinfo{journal}{IEEE Transactions on Circuits and Systems for Video
  Technology}
  \bibinfo{year}{2013};\bibinfo{volume}{23}(\bibinfo{number}{3}):\bibinfo{pages}{518--533}.
%Type = Article
\bibitem[{Zheng and Peng(2014)}]{Zheng2014}
\bibinfo{author}{Zheng\xfnm[ Y.]}, \bibinfo{author}{Peng\xfnm[ S.]}.
\newblock \bibinfo{title}{A practical roadside camera calibration method based
  on least squares optimization}.
\newblock \bibinfo{journal}{IEEE Transactions on Intelligent Transportation
  Systems} \bibinfo{year}{2014};\bibinfo{volume}{15}:\bibinfo{pages}{831--843}.

\end{thebibliography}

\end{document}